\documentclass[letterpaper, 10 pt, journal, twoside]{IEEEtran}  
\usepackage{booktabs} 
\usepackage{svg}
\usepackage{amssymb}
\usepackage{animate}
\usepackage{url}
\usepackage{booktabs}
\usepackage{multirow}
\usepackage{tikz}
\usepackage{graphicx}
\usepackage{xcolor}
\usepackage{tcolorbox}
\usepackage{amsmath}
\usepackage{bm} 
\usepackage{cancel}
\usepackage[hang,flushmargin,symbol]{footmisc}

\usepackage[resetlabels]{multibib}
\usepackage{subfiles}
\usepackage{wrapfig}
\usepackage{csquotes}
\usepackage{gensymb}
\usepackage{pifont}
\usepackage{textcomp}
\usepackage[table]{xcolor}
\usepackage{todonotes}
\usepackage{blindtext}
\usepackage{soul}
\usepackage{graphicx}
\usepackage{mathtools}
\usepackage{multirow}
\usepackage{booktabs}
\usepackage{array}
\usepackage{tabularx}
\usepackage[hang,flushmargin,symbol]{footmisc}
\usepackage{amsmath} 
\usepackage{amssymb} 
\usepackage{amsfonts}
\usepackage{bm} 
\usepackage{siunitx}
\usepackage{cancel}
\usepackage{microtype}
\usepackage{xcolor}
\usepackage[breaklinks,colorlinks]{hyperref}
\usepackage{caption}
\usepackage{subcaption}
\usepackage{cite}

\usepackage[inline]{enumitem}
\usepackage[export]{adjustbox}
\usepackage{cuted}
\usepackage{pgfplots}
\pgfplotsset{width=\linewidth,compat=1.9}
\definecolor{color_red}{RGB}{228,26,28}
\definecolor{color_blue}{RGB}{55,126,184}
\definecolor{color_green}{RGB}{77,175,74}
\definecolor{color_light_green}{rgb}{0.56, 0.93, 0.56}  
\definecolor{color_purple}{RGB}{152,78,163}
\definecolor{color_orange}{RGB}{255,127,0}
\definecolor{color_brown}{RGB}{166,86,40}
\definecolor{color_pink}{RGB}{247,129,191}
\newcites{S}{References}
\usepackage[capitalize]{cleveref}
\usepackage{hyperref}
\usepackage{svg}

\crefname{section}{Sec.}{Secs.}
\Crefname{section}{Section}{Sections}
\Crefname{table}{Table}{Tables}
\crefname{table}{Tab.}{Tabs.}

\definecolor{mygreen}{HTML}{00A64F}
\definecolor{myred}{HTML}{ED1B23}

\newcommand{\secref}[1]{Sec.~\ref{#1}}
\renewcommand{\eqref}[1]{Eq.~(\ref{#1})}
\newcommand{\figref}[1]{Fig.~\ref{#1}}
\newcommand{\tabref}[1]{Tab.~\ref{#1}}

\newcommand{\rebuttal}[1]{\textcolor{black}{#1}}

\renewcommand{\baselinestretch}{0.99}

\newcommand{\net}{COTRATE}

\title{
Self-Supervised Online Robot-Agnostic Traversability Estimation for Open-World Environments
}

\author{Julia Hindel, Simon Bultmann, Houman Masnavi, Daniele Cattaneo, and Abhinav Valada
\thanks{
This work was funded by the German Research Foundation (DFG) Emmy Noether Program grant number 468878300 and SFB 1597 – 499552394.} 
\thanks{All authors are with the Department of Computer Science, University of Freiburg, Germany,
        {\tt\footnotesize uni@julia-hindel.de}}%
}

\begin{document}

\maketitle

\begin{abstract}
Self-supervised online traversability estimation enables robots to continuously learn from \rebuttal{unlabeled experiences} and adapt their navigation behavior toward safe and efficient trajectories. Existing approaches either rely on handcrafted proprioceptive traversability scores, limiting robot-agnosticism, or cluster prior data, preventing online learning. Moreover, many continual learning methods incur substantial memory and computational costs, hindering onboard deployment. We introduce \net, an online learning framework for continuous traversability estimation from multimodal, unlabeled robot experience. Our method first infers robust traversability scores using a robot-agnostic, learning-based online terrain assessment module operating on proprioceptive and inertial signals. These scores then supervise a visual traversability network through a novel alignment loss that associates visual embeddings with online terrain assessments. To mitigate forgetting during continual learning with minimal overhead, we propose a diversity-aware feature selection strategy that preserves performance using a compact replay memory. We further show that the learned traversability representation supports knowledge transfer across different robot platforms with different locomotion kinematics. We evaluate \net~on a dataset of $\approx\,$50,000 images collected with two robotic platforms across $11$ outdoor terrains, and benchmark it on navigation tasks in \rebuttal{three outdoor} environments.
\end{abstract}


\begin{IEEEkeywords}
Deep Learning for Visual Perception, Semantic Scene Understanding, Continual Learning
\end{IEEEkeywords}


\section{Introduction}
\IEEEPARstart{M}{obile} robots are increasingly deployed in off-road environments for diverse applications, such as agriculture~\cite{hindel23} and forestry~\cite{valada2016towards}, where safe and energy-efficient navigation is essential. 
\rebuttal{Traversability estimation enables robots to assess terrain difficulty and update motion plans according to expected traversal cost. However, robots frequently encounter unseen terrains and changing surface conditions, an open-world problem that fixed models struggle to account for.}


Self-supervised learning from proprioceptive and inertial sensing \rebuttal{offers a promising solution}, as these signals directly reflect the \rebuttal{robot-terrain interaction}. Unlike pure visual supervision, proprioceptive feedback can capture quantities that are closely \rebuttal{tied} to physical traversal cost, including motion irregularity, slip, vibration, and energy expenditure. \rebuttal{Such signals can therefore generate traversability labels without manual annotation, supervising a visual model to predict terrain cost ahead of the robot.}
However, for fully autonomous deployment, this supervision should be generated online, generalize across robot platforms, and support continual visual learning without excessive memory or computation.

Existing self-supervised traversability methods only partially address these requirements. Several approaches derive traversability labels from handcrafted proprioceptive criteria, such as traction estimates~\cite{cai24evora, endo24deepprob} or \rebuttal{vertical }vibration intensity~\cite{howdoesitfeel}. While effective in specific settings, these scores are often tied to a particular \rebuttal{robot morphology or sensor configuration}, and they do not explicitly capture energy-efficient traversal \rebuttal{and hinder platform transfer}.
Other methods \rebuttal{instead} learn terrain representations from inertial, proprioceptive, or tactile sensors and cluster terrain types offline~\cite{karnan23sterling}. However, these approaches rely on pre-collected datasets and human ratings, whereas \rebuttal{open-world deployment requires adapting online as new terrains are encountered.}

\begin{figure}
    \centering
    \includegraphics[width=0.45\textwidth]{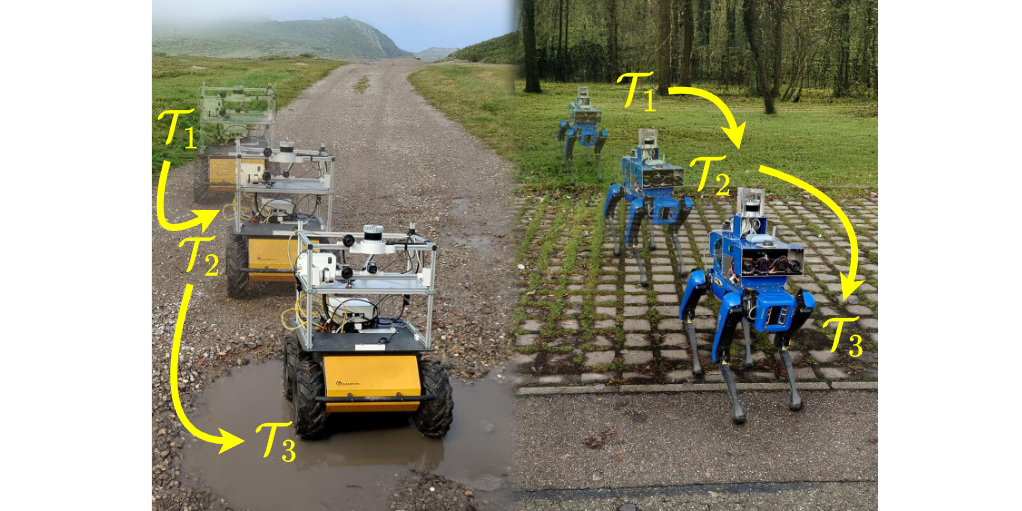}
    \caption{\net~learns self-supervised, robot-agnostic traversability scores from multimodal sensor data which supervise a visual traversability network during online learning.}
    \label{fig:teaser}
    \vspace{-0.5cm}
\end{figure}

Another critical challenge is the continual adaptation of the visual traversability model (\figref{fig:teaser}). As the robot encounters new terrain appearances, the visual model must incorporate new supervision while retaining prior knowledge. \rebuttal{Naive online updates lead to catastrophic forgetting~\cite{hindel24}. Existing continual learning strategies avoid this by storing raw data~\cite{ma2024imost, yoon24}, model ensembles~\cite{chen24alter}, or generative replay ~\cite{lee25cte}, often impractical under onboard compute and memory constraints.}

\rebuttal{In this work, we propose \net{} ({\textbf{C}}ontinual {\textbf{O}}nline {\textbf{T}}ransferable {\textbf{R}}obot-{\textbf{A}}gnostic {\textbf{T}}raversability {\textbf{E}}stimation), which unifies robot-agnostic terrain scoring, visual traversability learning, and continual adaptation into a single self-supervised pipeline.}
At the core of our method is a novel online traversability score generation module that predicts numerical traversability scores using an incrementally trained variational autoencoder \rebuttal{on multimodal robot experience}. These scores supervise a visual model that combines a pretrained visual foundation model with our novel feature-distance alignment loss\rebuttal{, correlating} visual feature similarities with learned continuous traversability scores. To enable continual learning, we introduce a diversity-aware feature-replay strategy that retains a compact set of representative features and mitigates forgetting under a minimal replay budget.
Finally, we study cross-platform transfer, \rebuttal{showing that the learned traversability representation transfers across robotic embodiments with different locomotion kinematics, given compatible terrain-dependent energy-efficiency ratings.
We evaluate \net{} on a new outdoor dataset spanning $11$ terrain types across two robotic platforms. Navigation experiments in three environments, with ablations, confirm the individual and combined contributions of terrain scoring, visual alignment, replay, and cross-platform transfer.}


In summary, our main contributions are:
\begin{itemize}[topsep=0pt,itemsep=0pt]
\item A novel robot-agnostic, online, self-supervised terrain assessment module that yields continuous traversability scores from proprioceptive and inertial data. 
\item A visual traversability \rebuttal{framework aligning monocular features with learned multimodal traversability scores.}
\item A \rebuttal{continual learning strategy based on diversity-aware} feature replay for low-memory onboard adaptation.
\item Extensive evaluations and ablation on two distinct robotic platforms, including navigation experiments in three representative outdoor environments.
\item We publicly release the code, models, and dataset at \\ \url{https://cotrate.cs.uni-freiburg.de}.
\end{itemize}
\section{Related Works}
\noindent\textit{Self-Supervised Traversability Estimation}: Self-supervised methods \rebuttal{estimate} traversability from on-board sensors without human labeling. \rebuttal{Common cues to distinguish traversable from non-traversable terrain include flat surfaces in 3D scene geometry~\cite{cho24, chen24alter, bu25otrav}, traction~\cite{frey2023fast}, and past terrain traversal~\cite{jung24}.} \rebuttal{Other work assesses collision risk via reinforcement learning~\cite{julian2020badgr}. These methods ensure safe but not energy-efficient navigation.}
Accordingly, other pertinent work focuses on computing continuous traversability costs from audio, inertial, and proprioceptive data. While vehicle-terrain acoustic interactions are clustered \rebuttal{for} weak terrain labels~\cite{Zurn2019SelfSupervisedVT}, others compute slip between the commanded and observed robot motion~\cite{endo24deepprob, gasparino22wayfast, gasparino2024wayfaster, cai24evora}. Thereby, terrains experiencing the highest deviation are regarded as least traversable. 
Another direction of work analyzes frequency patterns in IMU data to discriminate terrain difficulty~\cite{borges2022survey, sathy22terrapn, howdoesitfeel, seo24, sivaprakasam2024salon}. \rebuttal{High vibration along the robot's z-axis is} considered proportional to terrain difficulty.
Some prior works combine vibration with traction or joint measurements~\cite{sathy22terrapn, seneviratne25crossgait} to create robust metrics.
However, vibration and traction are not directly correlated with energy efficiency, and all available platform-dependent proprioceptive sensors should be automatically leveraged~\cite{karnan23sterling}. While AMCO~\cite{elnoor24AMCO} and ProNav~\cite{elnoor24pronav} cluster joint measurements to divide terrains, STERLING~\cite{karnan23sterling} proposes multimodal representation learning to correlate various tactile, inertial, and proprioceptive readings with image data. However, the latter requires offline clustering and human preferences to determine the final traversability scores~\cite{karnan23sterling}. \rebuttal{We therefore highlight the need for an online, robot-agnostic method that leverages all raw proprioceptive sensors for robust traversability estimation.}

\noindent\textit{Continual Traversability Estimation}:
While prior work commonly explores few- and zero-shot capabilities~\cite{cai24evora, jung24, endo24deepprob, sathy22terrapn, frey2023fast}, continual learning \rebuttal{instead avoids forgetting prior knowledge when learning new observations~\cite{zhou23deep}}. Traditional schemes include replay buffers, knowledge distillation, and network expansion~\cite{hindel24}, with replay methods outperforming the latter two on image classification and semantic segmentation benchmarks. Replay buffers save a small portion of previous data for periodic retraining.
In the context of traversability estimation, IMOST~\cite{ma2024imost} \rebuttal{selects suitable replay images via information-maximization, while ARTE~\cite{yoon24} targets LiDAR-based traversability, storing highest-loss samples.} \rebuttal{Lee~\textit{et~al.}~\cite{lee25cte} instead synthesize past data via a VAE-based replay model, avoiding raw image storage at higher computational cost. ALTER~\cite{chen24alter} employs model selection with ensemble models to alleviate forgetting.}
We use a frozen vision foundation backbone and store visual features selected via furthest-point sampling to span the full extent of the feature space, reducing computational and storage overhead while preventing forgetting.
\section{Technical Approach}
\begin{figure*}[h]
    \centering
    \includegraphics[width=0.75\textwidth]{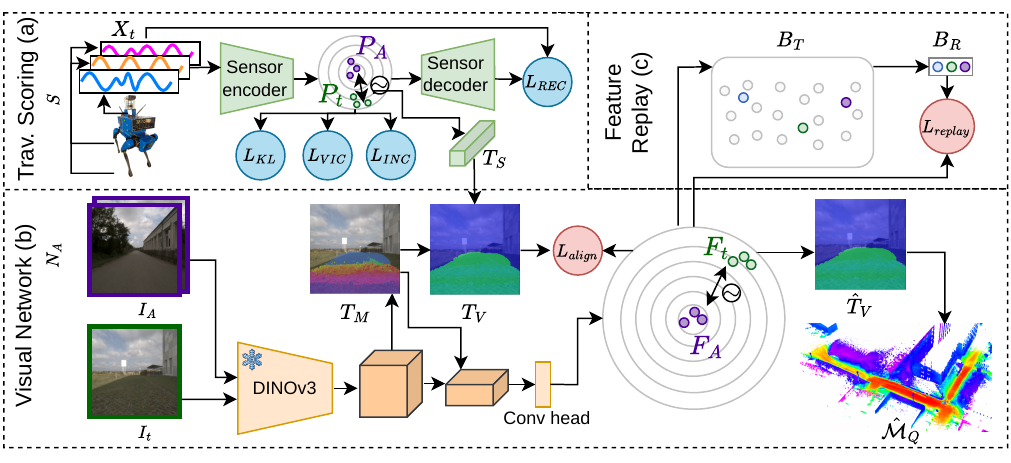} 
    \caption{\net~trains a Variational Auto-encoder (VAE) on different proprioceptive and inertial raw sensor data to obtain self-supervised traversability scores through learned embedding distances. Then, a DINOv3 backbone with a small trainable head extracts visual terrain features from a reference image and regularizes the observed feature differences to be equivalent to the embedding distances of the sensor model.}
    \label{fig:method}
    \vspace{-0.3cm}
\end{figure*}

In this section, we present the methodology of the \net~framework, which first learns traversability scores from inertial and proprioceptive signals and then uses them to incrementally train a visual network, as shown in \figref{fig:method}.

\subsection{Problem definition} \label{sec: def}
We address online traversability estimation for outdoor navigation with unknown terrain boundaries. We process \rebuttal{a continuous stream of }sensor sequences $\mathcal{D} = \{D_1, D_2, \ldots, D_o\}$, where each sequence $D_i$ contains RGB images $I$, point clouds \rebuttal{$Q$}, and \rebuttal{on-board} proprioceptive and inertial measurements. Our module learns normalized traversability scores $T_S \in [0,1]$ from multimodal sensor data. Scores are predicted every 5 minutes and converted to supervision signals for the vision model via odometry-based projection. Both visual and sensor models are updated online using a single training epoch. At inference, per-image traversability scores are back-projected onto the point cloud \rebuttal{$Q$} and accumulated into a dense 2.5D elevation map for downstream path planning and evaluation.

\subsection{Data Collection and Pre-Processing} \label{sec: data}

We collect a multimodal dataset \rebuttal{across} two platforms (Boston Dynamics Spot and Clearpath Husky) and $11$ terrain types, including walkways, open grounds, and playgrounds. Data is captured over four months with varying lighting and weather. We record three test environments at distant \rebuttal{sites}. In total, the dataset comprises $50{,}000$ images.
All sensor streams are recorded in ROS bag format with synchronized timestamps. Odometry is computed using Fast-LIO2 \cite{Xu20fastlio}. 

\noindent\textbf{Traversability score ($T_S$)}: 
To learn traversability scores in a self-supervised manner, we first temporally align sensor readings (IMU, joint motors, odometry, velocity commands) and linearly interpolate lower-frequency sensors. 
The synchronized data is partitioned into frames $X_t \in \mathbb{R}^{W \times S}$, where $W$ is the time window size and $S$ is the sensor dimensionality.

\rebuttal{We encode the data with a small Variational Auto-Encoder (VAE)~\cite{kingma13}, an established representation learning technique, consisting of an encoder with two CNN layers with ReLU, followed by adaptive average pooling across time and two parallel linear layers producing mean $P_t \in \mathbb{R}^{1 \times L}$ and variance $\sigma_t$. The decoder applies two CNN layers with ReLU to reconstruct $\mathbf{y}_t \in \mathbb{R}^{1 \times S}$.}
We employ four losses to separate terrains based on robot dynamics in the VAE latent space $\mathcal{P}$ as shown in \figref{fig:method}a. First, a Kullback-Leibler loss ($\mathcal{L}_{\text{KL}}$) \rebuttal{smooths the latent} feature distributions, and a \rebuttal{MSE-based reconstruction loss ($\mathcal{L}_{\text{REC}}$) supervises} the decoder output $\mathbf{y}_t$ against the last sensor reading of the input sequence $\mathbf{x}_{t}$ \rebuttal{with earlier timesteps providing temporal context to capture dynamics and periodicity rather than identity mapping~\cite{timeseriessurvey}.}
\rebuttal{We further enhance feature separation with VICReg~\cite{bardes22vicreg} ($\mathcal{L}_{\text{VIC}}$), a non-contrastive representation learning method.
VICReg combines an invariance term $s(\cdot,\cdot)$ (distance to a positive-pair target), a variance term $v(\cdot)$ (per-dimension hinge loss), and a covariance term $c(\cdot)$ (decorrelating feature dimensions), weighted by $\lambda, \mu, \nu$:
\small
\begin{equation}
\mathcal{L}_{\text{VIC}}(P, P^{+}) = \lambda s(P, P^{+}) + \mu[v(P){+}v(P^{+})] + \nu[c(P){+}c(P^{+})].
\end{equation}
\normalsize
Unlike STERLING~\cite{karnan23sterling}, which applies VICReg to image-viewpoint and image-sensor correlations, we learn time-series representations where the positive-pair target $P^{+}_{t_1}$ for $P_{t_1}$ is $P_{t_2}$ if $|t_1 - t_2| < H$, and $P_{t_1}$ itself otherwise; $t_1, t_2$ denote the timestamps of the last sensor readings of consecutive sequences, and $H$ is a predefined threshold.}
Consequently, $\mathcal{L}_{\text{VIC}}$ guides features from \rebuttal{temporally} adjacent samples to be similar. 
\rebuttal{Lastly, we ensure consistent features during online learning by replaying past sensor data at a 20:80 old-to-new data ratio, constraining the latent feature $P_t^{(i)}$ at its $i$-th visitation to stay close to its initial value $P_t^{(1)}$ (i.e., the latent feature the first time $t$ was seen) via $\mathcal{L}_{\text{INC}} = \| P_t^{(i)} - P_t^{(1)} \|_2^2$. Full data replay is feasible given the small memory overhead of sensor data ($<400\,\text{MB}$ on disk).
In summary, the total loss function for our traversability score creation module is: }
\begin{equation}
\mathcal{L} = \alpha \mathcal{L}_{\text{KL}} + \beta \mathcal{L}_{\text{REC}} + \gamma \mathcal{L}_{\text{VIC}} + \frac{\alpha + \beta + \gamma}{3} \mathcal{L}_{\text{INC}},
\end{equation}
where $\alpha$, $\beta$, $\gamma$ are tunable hyperparameters. \rebuttal{The weight for $\mathcal{L}_{INC}$ is averaged to remain independent of the incremental sequence or step count, fixed once after base training.}
We hypothesize that minimizing these losses organizes the embedding space such that \rebuttal{similarly-felt terrains cluster closely. We thus manually define a reference terrain $A$ from 5 minutes of navigation on one} smooth, well-traversable terrain (yielding $M_A$ timestamps), then compute traversability scores $T_{S,t}$ as the normalized cosine similarity \rebuttal{between $P_t$ and the mean reference feature $\overline{P_A}$}:
\begin{equation}
\overline{P_A} = \frac{1}{M_A}\sum_{m=1}^{M_A} P_{A,m}\,, \qquad T_{S,t} = \frac{\left<\overline{P_A}, P_t \right>}{\|\overline{P_A}\| \, \|P_t\|}.
\end{equation}
This brief calibration suffices for our self-supervised approach, and we observe the correlation between our computed scores and mean energy consumption per terrain in \secref{sec:ab}.

\begin{figure}
    \centering
    \includegraphics[width=0.42\textwidth]{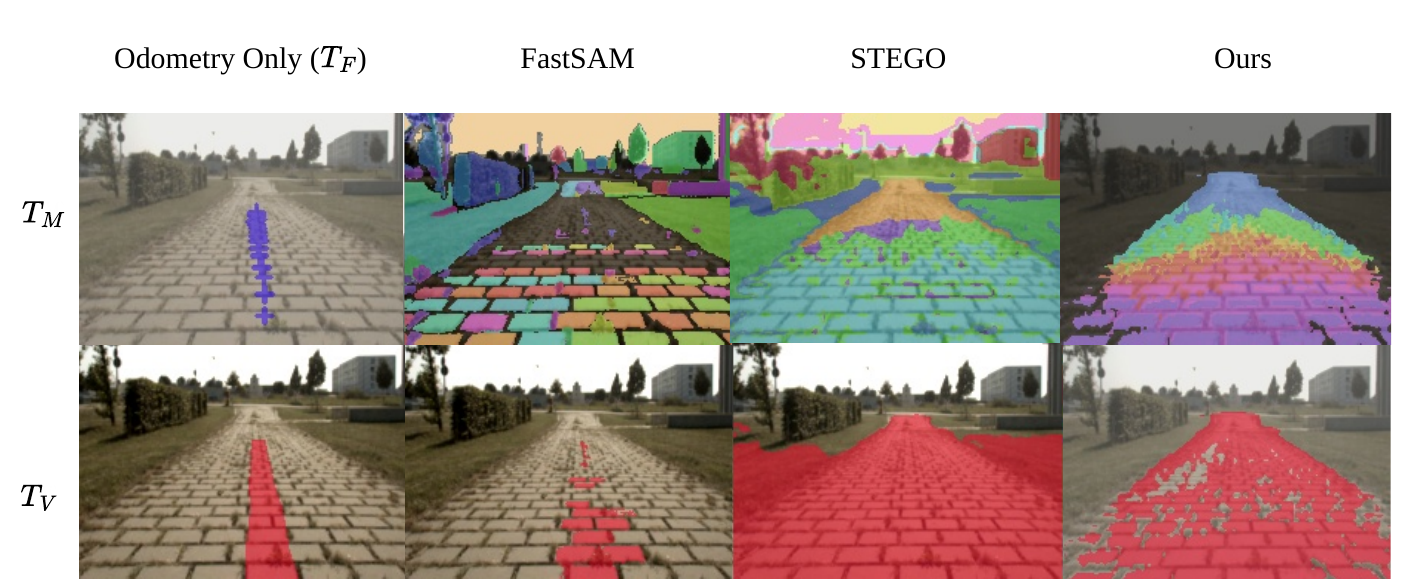} 
    \caption{Visual comparison of terrain masks predicted by STEGO~\cite{stego}, FastSAM~\cite{zhao2023fast} and DINOv3 (ours). $T_M$ shows segmentation masks in arbitrary colors, and Odometry Only indicates the robot's center as crosses. $T_V$ shows the combination of $T_M$ with visual footprints $T_F$.}
    \label{fig:TM}
    \vspace{-0.5cm}
\end{figure}

\noindent\textbf{Pixel-wise Visual Traversability ($T_V$)}:
To obtain sparse visual annotations, we \rebuttal{follow ~\cite {frey2023fast} and project $T_S$ onto image coordinates as visual footprints $T_F$, using odometry,} the center of mass, and camera calibration.
We then compute fine-grained terrain masks $T_M$ using a frozen DINOv3~\cite{dinov3} encoder, grouping footprint features with cosine similarity greater than threshold $c$ into segments (\figref{fig:TM}). \rebuttal{This grouping method outperforms the object-oriented FastSAM~\cite{zhao2023fast} and STEGO~\cite{stego} for ground segmentation (\figref{fig:TM}, \secref{sec:ab}).} We compute the final traversability mask $T_V$ by averaging $T_F$ within each segment of $T_M$.

\subsection{Visual Traversability Estimation Model}\label{sec: net_arch}
\rebuttal{We reuse the frozen DINOv3 ViT-backbone ~\cite{dinov3} introduced for generating $T_M$ as our visual encoder. Our decoder consists of two CNN layers with batch normalization and ReLU activations, yielding a $C \times H_d \times W_d$ feature tensor.} We interpolate to image dimensions $D=W\cdot H$, yielding $F\in\mathbb{R}^{D\times C}$.
 Then, we \rebuttal{select} $N_A$ reference images $\{I_A\}_{n=1}^{N_A}$ of the terrain $A$ and compute reference features $\{F_A\}$ in a forward pass.
 Next, we compute the mean $\overline{F_A}$ from $100$ randomly sub-sampled  \rebuttal{terrain features in $\{F_A\}$}.

\noindent\textbf{Multimodal Alignment Loss}: Our alignment loss enforces the cosine similarity of predictions $F_t$ for image $I_t$ versus $\overline{F_A}$ to match their traversability scores $T_{V,t}$:
\begin{equation}
\hat{T}_{V,t} = \frac{\left<\overline{F_A}, F_t\right>}{\|\overline{F_A}\| \, \|F_t\|}, ~
\mathcal{L}_\text{align} = \frac{1}{T} \sum_{t=1}^{T} \left(\hat{T}_{V,t} - T_{V,t} \right) ^2.
\label{eq:loss}
\end{equation}

This loss function ensures that distances between visual features reflect relative terrain difficulty. \rebuttal{We only backpropagate gradients through $F_t$ to update the decoder, and rescale the computed cosine similarities in $\hat{T}_{V,t}$ to a value range of $[0,1]$ for inference. We highlight that our traversability score prediction requires no clustering, unlike prior work which relies on k-means centroid distances~\cite{bu25otrav, karnan23sterling}.}

\noindent\textbf{Continual Terrain Learning}\label{sec: inc}:
\rebuttal{Continual learning enables adaptation to new terrains, but catastrophic forgetting is a major challenge. We mitigate this with selective feature replay, an established strategy that stores and replays selected features (prototypes) through our custom decoder, outperforming alternatives on classification tasks while minimizing storage~\cite{zhou23deep}. However, prototype selection relies on distributional heuristics designed for classification and does not apply to regression. We therefore select features via furthest-point sampling (FPS), adapting selective feature replay to our continuous traversability regression setting.}
Specifically, we maintain a replay feature buffer $B^R$ and a temporary feature buffer $B^T$. During online training, incoming visual terrain features are first stored in $B^T$. After a fixed time interval $t_b$, we update $B^R$ as \rebuttal{$B^R_ \text{new} = FPS(B^R_\text{old} \cup B^T)$}. 
\rebuttal{Then, we perform a forward pass through our decoder and save the resulting predictions as $F_{{B^R}_{\text{target}}}$. Our replay loss $\mathcal{L}_\text{replay} = \| F_{{B^R}_{\text{target}}} - F_{{B^R}_{\text{current}}} \|_2^2$ enforces consistency between the current decoder prediction $F_{{B^R}_{\text{current}}}$ and the saved target with an MSE loss every $t_r$ iterations.}
This online learning strategy results in stable training with minimal storage requirements. To reduce sparsity, we introduce feature cut-mix (FCM), which replaces features in unannotated image regions with features from $B^R$. FCM complements $\mathcal{L}_\text{replay}$ and stabilizes the semi-supervised training.\looseness=-1

\subsection{Traversability-Aware Navigation} \label{sec: nav} 
We build a 3D voxel map with a traversability channel using a semantic mapping approach~\cite{bultmann2021real}. 
Thus, we first project the time-synchronized LiDAR scans into the camera-based traversability prediction $\hat{T}_V$ using the known calibration and obtain point-wise traversability scores \rebuttal{$\hat{T}_Q$} via bilinear interpolation. Multiple \rebuttal{$\hat{T}_Q$} are then aggregated into a 3D voxel map \rebuttal{$\hat{\mathcal{M}}_Q$} using LiDAR odometry~\cite{Xu20fastlio}, where voxel-wise traversability scores are updated over time using a recursive moving average.
Subsequently, we reduce \rebuttal{$\hat{\mathcal{M}}_Q$} to a 2.5D elevation map $\hat{\mathcal{M}}_E$ by projecting voxels between the ground plane, estimated via RANSAC, and a maximum height $h_\text{max}$.

\noindent\textbf{Path Evaluation}: The planner \rebuttal{uses} A* search with a terrain-aware cost \rebuttal{integrating} geometric distance and traversability. For each edge traversal, the cost is defined as:
\begin{equation}
R_{\text{edge}} = \sqrt{\ell_{\text{horiz}}^2 + \Delta z^2} \cdot \left(1 + (1 - \hat{T}_G)\, w_{\text{trav}}\right),
\end{equation}
where $\ell_{\text{horiz}}$ is the horizontal distance, $\Delta z$ the elevation change for obstacle avoidance, $\hat{T}_G \in [0,1]$ the average traversability of adjacent cells, and $w_{\text{trav}}$ a tunable weight. This formulation penalizes low-traversability regions while accounting for 3D geometry. The heuristic uses Euclidean distance to ensure admissibility and \rebuttal{optimality with respect to this cost}.

\section{Experimental Results}
In this section, we quantitatively and qualitatively evaluate the performance of \net~in comparison to six traversability estimation baselines. 
With our experiments, we aim to answer the following questions:
\begin{itemize}
    \item Can our traversability estimation method choose the most energy-efficient path? 
    \item Can our introduced feature replay strategy prevent catastrophic forgetting more effectively than the baselines?
    \item Can our introduced traversability estimation model generalize zero-shot to a different robot embodiment?
\end{itemize}

\subsection{Datasets}\label{sec:data}
Each platform is equipped with a frontal RGB camera, an Ouster OS0-128 3D LiDAR, an IMU, and robot-internal proprioceptive sensors (e.g., joint encoders and motor currents). \rebuttal{Images were recorded at Full-HD resolution with Spot and $1024\times1024$ resolution with Husky, with camera and LiDAR data captured at $10\,\text{Hz}$ and IMU measurements at $100\,\text{Hz}$.}

\begin{figure}
    \centering
    \includegraphics[width=0.4\textwidth]{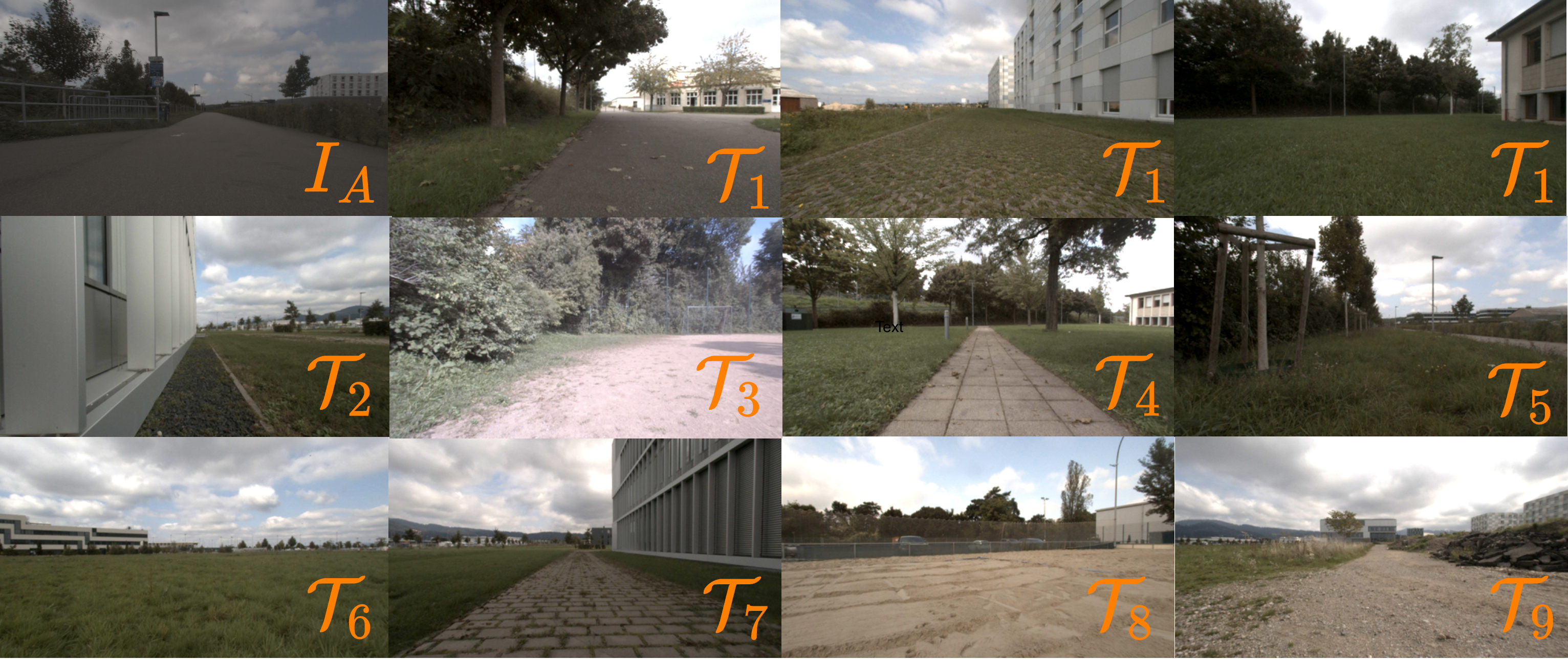} 
    \caption{Example images of our dataset recorded on Spot. $I_A$: example of reference terrain (asphalt), $\mathcal{T}_i$: terrains in continual learning order: asphalt, spaced cobblestone w. grass, grass, coarse gravel, bare dirt, cobblestone, tall grass, medium-high grass, spaced cobblestone, sand, and fine gravel.}
    \label{fig:data}
    \vspace{-.3cm}
\end{figure}

\begin{table*}
\centering
\caption{\rebuttal{Comparison of Effort and Effort per Path Length (EPL) across three environments ($E1-E3$) and two robots, with mIoU [\%] for 2.5D and 2D segmentation. Training settings: $S_i$/$H_i$: continual learning on Spot/Husky, $S$/$H$: joint training on Spot/Husky, ZS: zero-shot transfer from the other platform. Bold: best; underline: second best.}}

\setlength\tabcolsep{2pt}
\begin{tabular}{@{}llc|ccc|ccc|ccc|c}
 \toprule
 & & \textbf{Train.} & \multicolumn{3}{c|}{\textbf{Effort} $\downarrow$} & \multicolumn{3}{c|}{\textbf{EPL} $\downarrow$} & \multicolumn{3}{c|}{\textbf{2.5D segm.} $\uparrow$} & \textbf{2D segm.} $\uparrow$ \\
 \cmidrule{4-13}
& \textbf{Method} & \textbf{Data} & E1 & E2 & E3 & E1 & E2 & E3 & E1 & E2 & E3 & all E\\
\midrule
\multirow{9}{*}{\rotatebox{90}{Spot}} & Euclidean Dist. & & \rebuttal{10.02 $\pm$ 0.57} & \rebuttal{15.95 $\pm$ 0.91} & \rebuttal{11.13 $\pm$ 1.03} & \rebuttal{0.427 $\pm$ 0.02} & \rebuttal{0.513 $\pm$ 0.02} & \rebuttal{0.644 $\pm$ 0.01}& - & - & - & - \\
& Joint & $S$ & \rebuttal{7.09 $\pm$ 0.56} & \rebuttal{2.80 $\pm$ 1.06} & \rebuttal{7.10 $\pm$ 3.22} & \rebuttal{0.247 $\pm$ 0.01} & \rebuttal{0.063 $\pm$ 0.02} &\rebuttal{0.391 $\pm$ 0.23} & \rebuttal{58.21} & \rebuttal{42.14} & \rebuttal{33.57} & \rebuttal{41.56} \\
 \cmidrule{2-13}
 \cmidrule{2-13}
& LangSAM & & \rebuttal{7.21 $\pm$ 0.57} & \rebuttal{4.16 $\pm$ 1.79} & \rebuttal{11.09 $\pm$ 4.11} & \rebuttal{0.265 $\pm$ 0.01} & \rebuttal{0.106 $\pm$ 0.05} & \rebuttal{0.695 $\pm$ 0.15} &29.09 & 7.90 & 6.71 & 23.87 \\
& WVN \cite{frey2023fast} & $S_{i}$ & \rebuttal{9.57 $\pm$ 1.03} & \rebuttal{\underline{3.87} $\pm$ 1.93} & \rebuttal{17.79 $\pm$ 0.51} & \rebuttal{0.395 $\pm$ 0.04} & \rebuttal{\underline{0.104} $\pm$ 0.06} & \rebuttal{0.605 $\pm$ 0.01} & - & - & - & - \\
& IMOST \cite{ma2024imost} &$S_{i}$ & \rebuttal{9.78 $\pm$ 0.61} & \rebuttal{9.93 $\pm$ 1.26} & \rebuttal{\underline{6.65} $\pm$ 0.62} & \rebuttal{0.347 $\pm$ 0.01} & \rebuttal{0.290 $\pm$ 0.04} & \rebuttal{\underline{0.320} $\pm$ 0.05} & 26.27 & 9.22 & 2.46 & 10.92 \\
& Image replay &$S_{i}$ & \rebuttal{7.66 $\pm$ 0.56} & \rebuttal{5.08 $\pm$ 1.67} & \rebuttal{10.16 $\pm$ 1.62} & \rebuttal{0.296 $\pm$ 0.01} & \rebuttal{0.143 $\pm$ 0.06} & \rebuttal{0.658 $\pm$ 0.02} & \underline{40.18} & 28.20 & \underline{18.15} & \underline{30.73} \\
& VAE-replay \cite{lee25cte} &$S_{i}$ & \rebuttal{8.42 $\pm$ 0.57} & \rebuttal{9.41 $\pm$ 1.18} & \rebuttal{10.83 $\pm$ 1.68} & \rebuttal{0.327 $\pm$ 0.01} & \rebuttal{0.260 $\pm$ 0.04} & \rebuttal{0.703 $\pm$ 0.01} & 26.17 & 16.06 & 8.62 & 13.17   \\
& \rebuttal{Ensemble \cite{chen24alter}} &$S_{i}$  & \rebuttal{\underline{7.12} $\pm$ 0.56} & \rebuttal{4.51 $\pm$ 2.41} & \rebuttal{9.18 $\pm$ 1.43} & \rebuttal{\underline{0.245} $\pm$ 0.01}& \rebuttal{0.126 $\pm$ 0.07}& \rebuttal{0.597 $\pm$ 0.01}& \rebuttal{27.41}& \rebuttal{18.54}& \rebuttal{12.26}&\rebuttal{17.25} \\ 
 \cmidrule{2-13}
& COTRATE (Ours) & $S_{i}$ & \rebuttal{\textbf{7.02} $\pm$ 0.56} & \rebuttal{\textbf{3.28} $\pm$ 0.87} & \rebuttal{\textbf{6.18} $\pm$ 1.07} & \rebuttal{\textbf{0.239} $\pm$ 0.01} & \rebuttal{\textbf{0.075} $\pm$ 0.01} & \rebuttal{\textbf{0.287} $\pm$ 0.07} &\textbf{53.57} & \textbf{39.40} & \textbf{25.88} & \textbf{33.23} \\
& COTRATE-ZS & $H_{i}$ & \rebuttal{8.42 $\pm$ 0.58} & \rebuttal{4.41 $\pm$ 1.81} & \rebuttal{{8.45} $\pm$ 3.18} & \rebuttal{0.305 $\pm$ 0.01} & \rebuttal{0.116 $\pm$ 0.06} & \rebuttal{0.465 $\pm$ 0.30} & 23.68 & \underline{37.50} & 13.00 & 21.15 \\
\midrule
\midrule 
\multirow{9}{*}{\rotatebox{90}{Husky}} & Euclidean Dist. & & \rebuttal{12.35 $\pm$ 0.66} & \rebuttal{5.19 $\pm$ 0.41} & \rebuttal{11.51 $\pm$ 2.53} & \rebuttal{0.394 $\pm$ 0.01} & \rebuttal{0.195 $\pm$ 0.02} & \rebuttal{0.541 $\pm$ 0.11}& - & - & - & - \\
& Joint & $H$ & \rebuttal{11.16 $\pm$ 1.45} & \rebuttal{1.14 $\pm$ 0.62} &  \rebuttal{9.55 $\pm$ 2.85} & \rebuttal{0.324 $\pm$ 0.09} & \rebuttal{0.057 $\pm$ 0.05} & \rebuttal{0.323 $\pm$ 0.18}& \rebuttal{43.07} & \rebuttal{28.30} & \rebuttal{9.26} & \rebuttal{24.53}\\
 \cmidrule{2-13}
 \cmidrule{2-13}
& LangSAM & & \rebuttal{10.76 $\pm$ 1.19} & \rebuttal{3.79 $\pm$ 0.07} & \rebuttal{9.94 $\pm$ 1.35} & \rebuttal{0.281 $\pm$ 0.05} & \rebuttal{0.120 $\pm$ 0.01} & \rebuttal{0.470 $\pm$ 0.04} & 33.79 & 12.04 & 2.50 & \underline{21.49}\\
& WVN \cite{frey2023fast} & $H_{i}$ & \rebuttal{12.75 $\pm$ 1.29} & \rebuttal{3.93 $\pm$ 0.18} & \rebuttal{15.79 $\pm$ 1.42} & \rebuttal{0.373 $\pm$ 0.07} & \rebuttal{0.134 $\pm$ 0.01} & \rebuttal{0.489 $\pm$ 0.03} & - & - & - & -\\
& IMOST \cite{ma2024imost} &$H_{i}$ & \rebuttal{11.23 $\pm$ 0.73} & \rebuttal{3.83 $\pm$ 0.13} & \rebuttal{9.81 $\pm$ 2.36} & \rebuttal{0.327 $\pm$ 0.06} & \rebuttal{0.127 $\pm$ 0.01} & \rebuttal{0.468 $\pm$ 0.11} & 8.68 & 9.61 & 3.53 & 10.63\\
& Image replay &$H_{i}$ & \rebuttal{10.66 $\pm$ 1.71} & \rebuttal{4.03 $\pm$ 0.23} & \rebuttal{14.92 $\pm$ 2.87} & \rebuttal{0.301 $\pm$ 0.09} & \rebuttal{0.139 $\pm$ 0.02} & \rebuttal{0.712 $\pm$ 0.17} & 26.07 & 20.71 & \underline{5.43} & 17.06 \\
& VAE-replay\cite{lee25cte}&$H_{i}$ & \rebuttal{10.54 $\pm$ 1.34} & \rebuttal{3.79 $\pm$ 0.06} & \rebuttal{10.62 $\pm$ 2.69} & \rebuttal{0.298 $\pm$ 0.08} & \rebuttal{0.122 $\pm$ 0.01} & \rebuttal{0.506 $\pm$ 0.15} & 12.87 & 19.84 & 4.27 & 14.66\\
& \rebuttal{Ensemble \cite{chen24alter}} &$H_{i}$ & \rebuttal{\rebuttal{11.10 $\pm$ 0.86}}& \rebuttal{\rebuttal{3.48 $\pm$ 0.85}} &\rebuttal{\rebuttal{11.45 $\pm$ 2.16}} & \rebuttal{\rebuttal{0.331 $\pm$ 0.06}} & \rebuttal{\rebuttal{0.113 $\pm$ 0.04}}& \rebuttal{\rebuttal{0.547 $\pm$ 0.08}}& \rebuttal{7.58}& \rebuttal{15.36}& \rebuttal{4.31}& \rebuttal{14.69} \\ 
 \cmidrule{2-13}
& COTRATE (Ours) & $H_{i}$ & \rebuttal{\textbf{9.57} $\pm$ 0.94} & \rebuttal{\underline{1.15} $\pm$ 0.04} & \rebuttal{\underline{9.11} $\pm$ 2.91} & \rebuttal{\textbf{0.214} $\pm$ 0.02} & \rebuttal{\textbf{0.0205} $\pm$ 0.01} & \rebuttal{\underline{0.309} $\pm$ 0.17} & \underline{35.87} & \underline{25.42} & \textbf{7.83} & 17.73\\
& COTRATE-ZS & $S_{i}$ & \rebuttal{\underline{10.46} $\pm$ 1.04} & \rebuttal{\textbf{1.11} $\pm$ 0.05} & \rebuttal{\textbf{9.10} $\pm$ 3.05} & \rebuttal{\underline{0.234} $\pm$ 0.03} & \rebuttal{\underline{0.0207} $\pm$ 0.01} & \rebuttal{\textbf{0.308} $\pm$ 0.18} & \textbf{44.52} & \textbf{25.78} & 5.38 & \textbf{21.80}\\
\bottomrule 
\end{tabular}
\label{tab:inc_comparison}
\vspace{-1em}
\end{table*}

\subsection{Experimental Settings} \label{sec:setting}
We incrementally learn the terrains in the order shown in \figref{fig:data}\rebuttal{, training the VAE on the first three terrains for $13$ epochs, then applying online learning ($1$ epoch) to the rest.
The latent variable $P$ has dimensionality $L=16$, with loss weights $\alpha = 16$, $\beta = 16$, $\gamma = 3$, and $H=1.6\,\mathrm{s}$.} We set threshold $c=0.95$ to produce terrain masks $T_M$ using ViT-B at full resolution\rebuttal{; lowering it merges terrains ($c=0.85$) or background structure ($c<0.7$)}.
We use $5$ asphalt images as reference $I^A$ and a feature dimension $C=64$. Replay buffers are set to $t_b=100$, $t_r=1$, and $B^R=200$. We weight $\mathcal{L}_\text{replay}$ $20\times$ higher than $\mathcal{L}_\text{align}$. Training uses $224\times224$ crops with augmentations (horizontal flip, color jitter, Gaussian blur, grayscale), SGD with momentum $0.9$ and weight decay $10^{-4}$, learning rate $10^{-5}$, batch size $2$, and $1$ epoch per terrain. \rebuttal{Higher input resolutions empirically improve terrain discrimination at boundaries, at the cost of increased compute.} Navigation sets $h_{\max}=1\,\text{m}$.\looseness=-1

We formulate traversability as a regression to capture within-terrain variations. For terrain change detection, we discretize predictions and compute IoU with ground truth, counting a prediction correct if within $\pm 2\times\text{std}$ of the per-terrain mean. We hand-label $2{,}450$ test images and grid maps for evaluation. Navigation efficiency is assessed via terrain effort signals, computed from average motor current during training sequences, which correlates with energy expenditure.
We compute path effort $U_{\text{effort}}$ as terrain effort weighted by distance traveled:
$ U_{\text{effort}} = \sum_{i=1}^{N} u_i \, \Delta s_i$ 
where $u_i \in [0,1]$ is normalized terrain effort, $\Delta s_i$ the segment distance, and $N$ the number of segments. Effort Weighted Path Length (EPL) normalizes $U_{\text{effort}}$ by total distance. \rebuttal{An EPL reduction of $0.1$ corresponds to an estimated power decrease of
$13\,\text{W}$ for Spot and $9\,\text{W}$ for Husky. We run $50$ trials per experiment with starting positions randomly sampled from a $3\times3\,\mathrm{m}$ region.
We train \net~on an NVIDIA RTX 4060, requiring $4.2\,\text{GB}$ for the visual model and $0.6\,\text{GB}$ for the score VAE.
On the robot's Zotac Z-Box (Intel i7-11800H, 64\,GB RAM, RTX 3070 mobile GPU with 8\,GB VRAM), inference takes 5\,ms (FP32) for $224\times224$ inputs, i.e., 37.46\,GFLOPs per forward pass.}

\subsection{Quantitative Results}\label{sec:results}
We evaluate \net~against \rebuttal{five} self-supervised baselines supporting continual learning or fast adaptation. For WVN~\cite{frey2023fast}, we use published code with equivalent training. For continual learning baselines, we integrate replay strategies from IMOST~\cite{ma2024imost}, \rebuttal{Ensemble~\cite{chen24alter}} and VAE-replay~\cite{lee25cte}. \rebuttal{Image replay uses a buffer of $5$ images, updated regularly.} LangSAM serves as a zero-shot semantic segmentation baseline, converting predicted masks into effort-based traversability scores. \rebuttal{Joint training on all data for $30$ epochs (Spot: bs$=2$, lr$=10^{-5}$; Husky: bs$=8$, lr$=10^{-4}$) provides a baseline.} \looseness=-1

Evaluation results for three environments with both robots are shown in \tabref{tab:inc_comparison}, with additional visuals in the multimedia material. \net~achieves the lowest mean Effort on all scenes on both robots.
\rebuttal{Effort decreases by an average of $0.4$ for Spot and $1.3$ for Husky compared to the best baseline.}
Our method also outperforms baselines in EPL, reflecting its ability to choose the most energy-efficient routes, which we attribute to coherent traversability scores and knowledge retention.
Further, \net~outperforms baselines by at least $2.5$pp on both mIoU metrics \rebuttal{for Spot, and by at least $2.1$pp on 2.5D mIoU for Husky, showing reliable prediction of energy-efficient preferences across all environments.
Further, \net~matches or beats joint training in Effort/EPL in E1 (Husky) and E3 (Spot), likely as sequential online learning with replay limits terrain interference, whereas joint training must fit all terrains jointly.}
Next, \rebuttal{LangSAM performs poorly on different grass types and cobblestone, evident from low 2.5D IoU scores in E2--E3. Finally, evaluating zero-shot capabilities, we find the zero-shot model performs on-par with the platform-specific one on E2--E3 (Husky), confirming that traversability scores are calibrated and transfer effectively across embodiments with similar energy-efficiency ratings (\secref{sec:ab}).}

\begin{figure}[t]
    \centering
    \begin{subfigure}[b]{0.49\linewidth}
        \centering
        \resizebox{\linewidth}{!}{\resizebox{\columnwidth}{!}{%
\begin{tikzpicture} [font=\small]
\begin{axis}[
    ylabel={mIoU [\%]},
    xlabel={Step $T_i$ for Spot},
    xlabel style={yshift=1.6mm}, 
    tick label style={font=\scriptsize, black},
    label style={font=\scriptsize},
    every mark/.append style={scale=0.6},
    xmin=0.5, xmax=9.5,
    ymin=0, ymax=38,
    ytick={0, 10, 20, 30},
    xtick={1,3,5,7,9},
    ymajorgrids=true,
    grid style=dashed,
    height=4cm,
    legend columns=5,
    legend to name={sharedlegend},
    legend style={
        font=\scriptsize,
        text=black,   
    },
]
\addplot[color=color_light_green, mark=square, line width=0.4mm, legend image code/.code={
        \draw[color_light_green, mark=square, mark options={fill=color_light_green}]
            plot coordinates {(0.3cm,0cm)};
    }]
    coordinates {(1,31.29)(2,28.36)(3,25.90)(4,27.46)(5,31.11)(6,30.57)(7,30.02)(8,31.55)(9,33.23)};
\addlegendentry{\net}
\addplot[color=color_brown, mark=triangle, line width=0.4mm, 
legend image code/.code={
        \draw[color_brown, mark=triangle, mark options={fill=color_brown}]
            plot coordinates {(0.3cm,0cm)};
    }]
    coordinates {(1,9.74)(2,10.51)(3,9.24)(4,6.42)(5,26.58)(6,32.93)(7,11.67)(8,21.05)(9,13.17)};
\addlegendentry{VAE-replay}
\addplot[color=color_purple, mark=star, line width=0.4mm, legend image code/.code={
        \draw[color_purple, mark=star, mark options={fill=color_purple}]
            plot coordinates {(0.3cm,0cm)};
    }]
    coordinates {(1,7.34)(2,8.84)(3,8.33)(4,6.30)(5,26.40)(6,27.93)(7,7.41)(8,8.92)(9,10.93)};
\addlegendentry{IMOST}
\addplot[color=color_blue, mark=triangle, line width=0.4mm, legend image code/.code={
        \draw[color_blue, mark=triangle, mark options={fill=color_blue}]
            plot coordinates {(0.3cm,0cm)};
    }]
    coordinates {(1,7.26)(2,18.41)(3,27.60)(4,29.48)(5,30.50)(6,31.66)(7,27.58)(8,29.83)(9,30.72)};
\addlegendentry{Image replay}
\addplot[color=color_red, mark=triangle, line width=0.4mm, 
legend image code/.code={
        \draw[color_red, mark=triangle, mark options={fill=color_red}]
            plot coordinates {(0.3cm,0cm)};
    }]
      coordinates {(1,11.96)(2,12.87)(3,10.04)(4,7.61)(5,9.93)(6,27.10)(7,14.44)(8,21.48)(9,17.74)};
\addlegendentry{Ensemble}
\end{axis}
\end{tikzpicture}}}
        \label{fig:forgetting_spot}
    \vspace{-1em}
    \end{subfigure}
    \hfill
    \begin{subfigure}[b]{0.49\linewidth}
        \centering
        \resizebox{\linewidth}{!}{\begin{tikzpicture}[font=\small]
\begin{axis}[
    ylabel={mIoU [\%]},
    xlabel={\rebuttal{Step $T_i$ for Spot Reversed}},
    xlabel style={yshift=1.5mm}, 
    tick label style={font=\scriptsize, black},
    label style={font=\scriptsize},
    every mark/.append style={scale=0.6},
    xmin=0.5, xmax=9.5,
    ymin=0, ymax=38,
    ytick={0, 10, 20, 30},
    xtick={1,3,5,7,9},
    ymajorgrids=true,
    grid style=dashed,
    height=4cm,
]
\addplot[color=color_light_green, mark=square, line width=0.4mm]
    coordinates {(1,33.44)(2,24.12)(3,19.48)(4,35.03)(5,33.73)(6,26.74)(7,17.85)(8,22.80)(9,22.82)};
\addplot[color=color_brown, mark=triangle, line width=0.4mm]
    coordinates {(1,21.34)(2,11.05)(3,13.63)(4,7.26)(5,20.40)(6,17.36)(7,19.25)(8,11.10)(9,13.87)};
\addplot[color=color_purple, mark=star, line width=0.4mm]
    coordinates {(1,20.28)(2,12.77)(3,12.35)(4,7.52)(5,22.10)(6,19.28)(7,6.69)(8,5.45)(9,8.85)};
\addplot[color=color_blue, mark=triangle, line width=0.4mm]
    coordinates {(1,19.76)(2,7.29)(3,20.70)(4,20.29)(5,32.37)(6,32.42)(7,21.53)(8,22.79)(9,24.08)};
\addplot[color=color_red, mark=triangle, line width=0.4mm]
      coordinates {(1,20.20)(2,13.71)(3,15.77)(4,5.13)(5,19.52)(6,17.65)(7,7.43)(8,5.00)(9,9.75)};
\end{axis}
\end{tikzpicture}}
        \label{fig:lineplot_spot}
        \vspace{-1em}
    \end{subfigure}
    \ref*{sharedlegend}
    \vspace{-.2em}
    \caption{2D segmentation mIoU recorded over all test terrains per increment.}
    \label{fig:forgetting_combined}
\vspace{-.5em}
\end{figure}
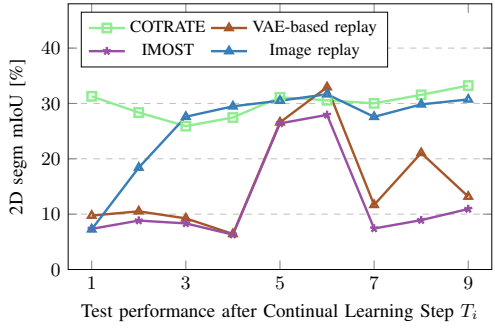

We evaluate catastrophic forgetting for the continual learning methods \rebuttal{in normal and reversed order on Spot in \figref{fig:forgetting_combined}.} Image replay and \net~exhibit strong continual learning capabilities, evidenced by consistent predictions, with \rebuttal{\net~achieving the best overall performance in the forward order at lower memory cost.}
IMOST and VAE-replay fluctuate strongly between steps 4 and 5 (3--4 in reverse order), likely from relearning grass when a new grass type appears.

\begin{figure}[t]
    \centering

    \begin{tikzpicture}


    \node[anchor=north west,inner sep=0] (r1c1) at (0,0)
    {\includegraphics[trim={0 0 0 0},clip,height=2.31cm]
    {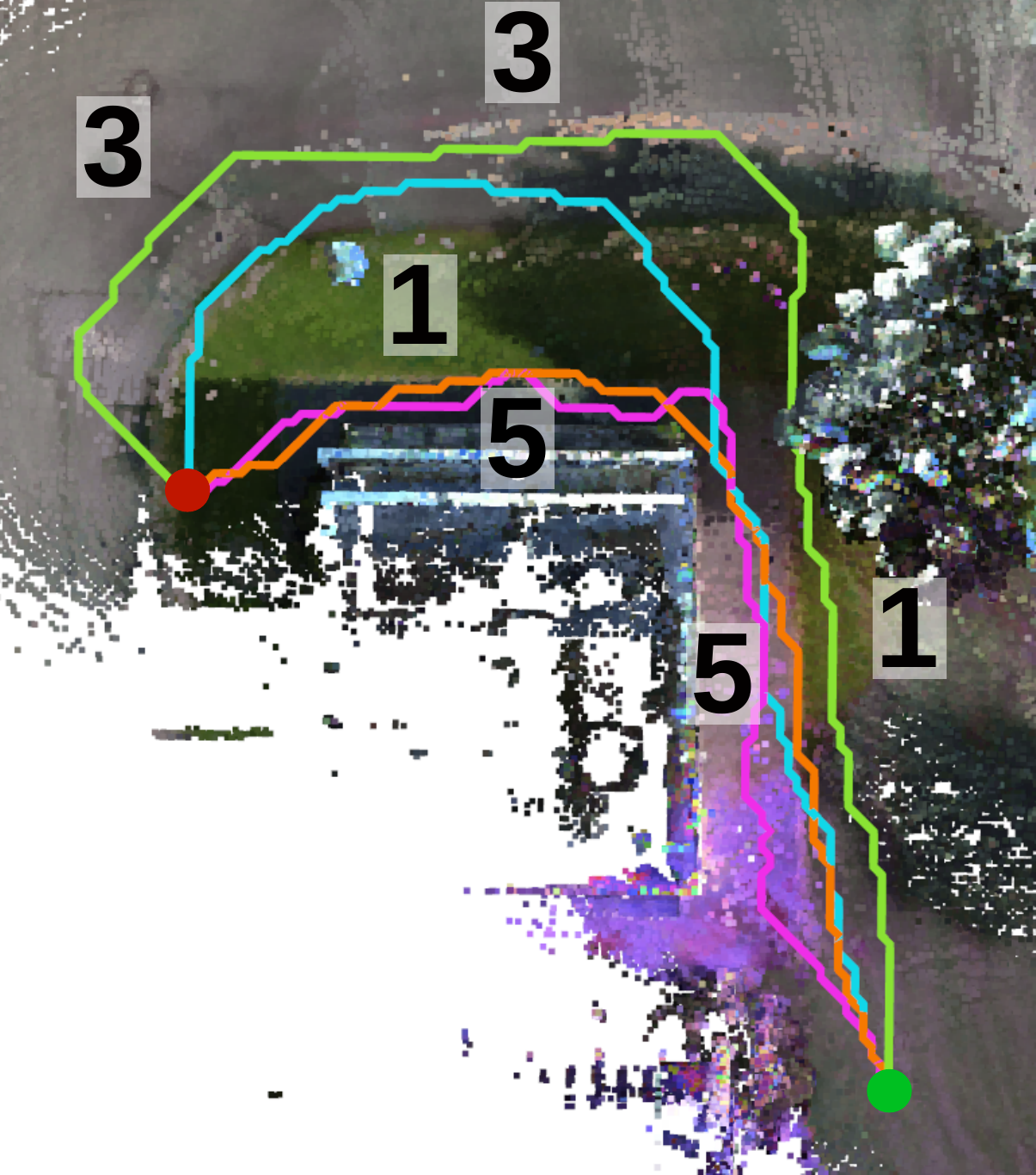}};

    \node[anchor=north west,inner sep=0,xshift=0.15em] (r1c2) at (r1c1.north east)
    {\includegraphics[trim={0 0 0 0},clip,height=2.31cm]
    {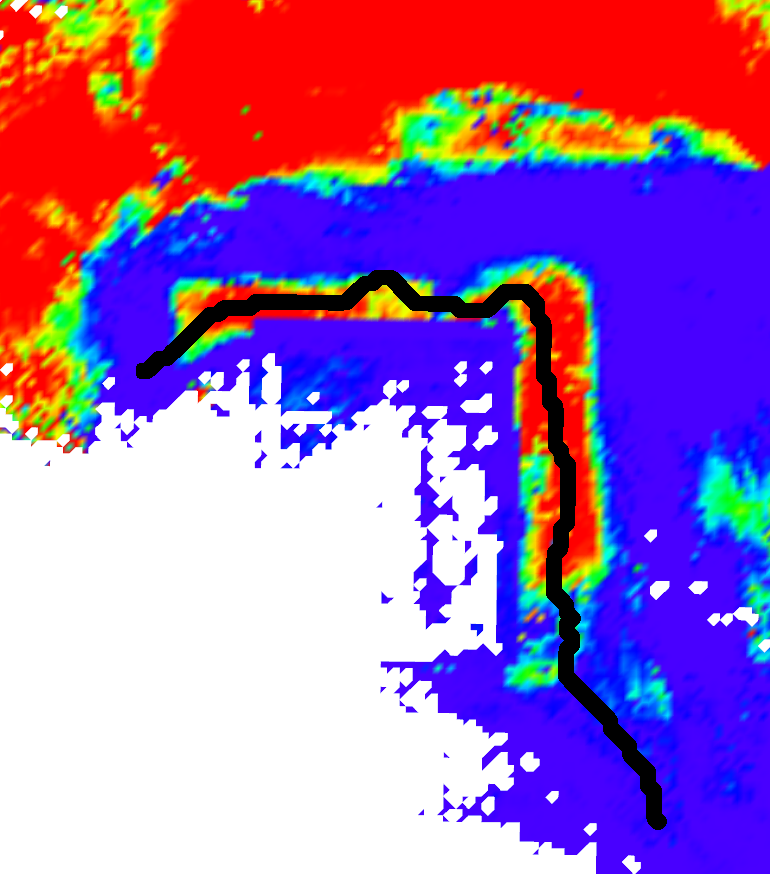}};

    \node[anchor=north west,inner sep=0,xshift=0.15em] (r1c3) at (r1c2.north east)
    {\includegraphics[trim={0 0 0 0},clip,height=2.31cm]
    {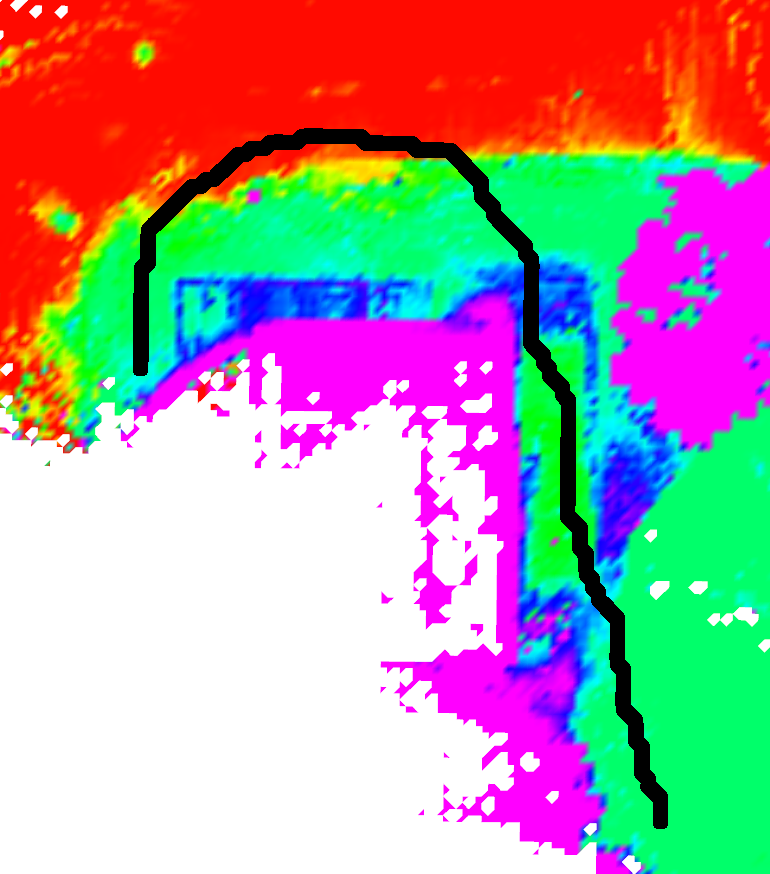}};

    \node[anchor=north west,inner sep=0,xshift=0.15em] (r1c4) at (r1c3.north east)
    {\includegraphics[trim={0 0 0 0},clip,height=2.31cm]
    {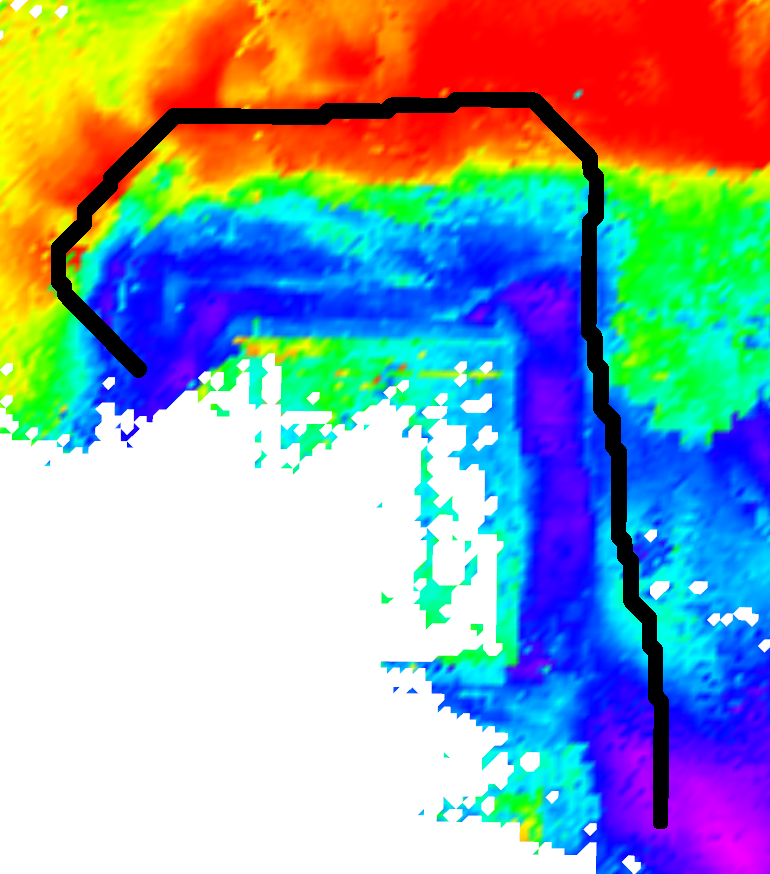}};

    \node[anchor=north west,inner sep=0,yshift=-0.35em] (r2c1) at (r1c1.south west)
    {\includegraphics[trim={0 0 0 0},clip,height=4.74cm]
    {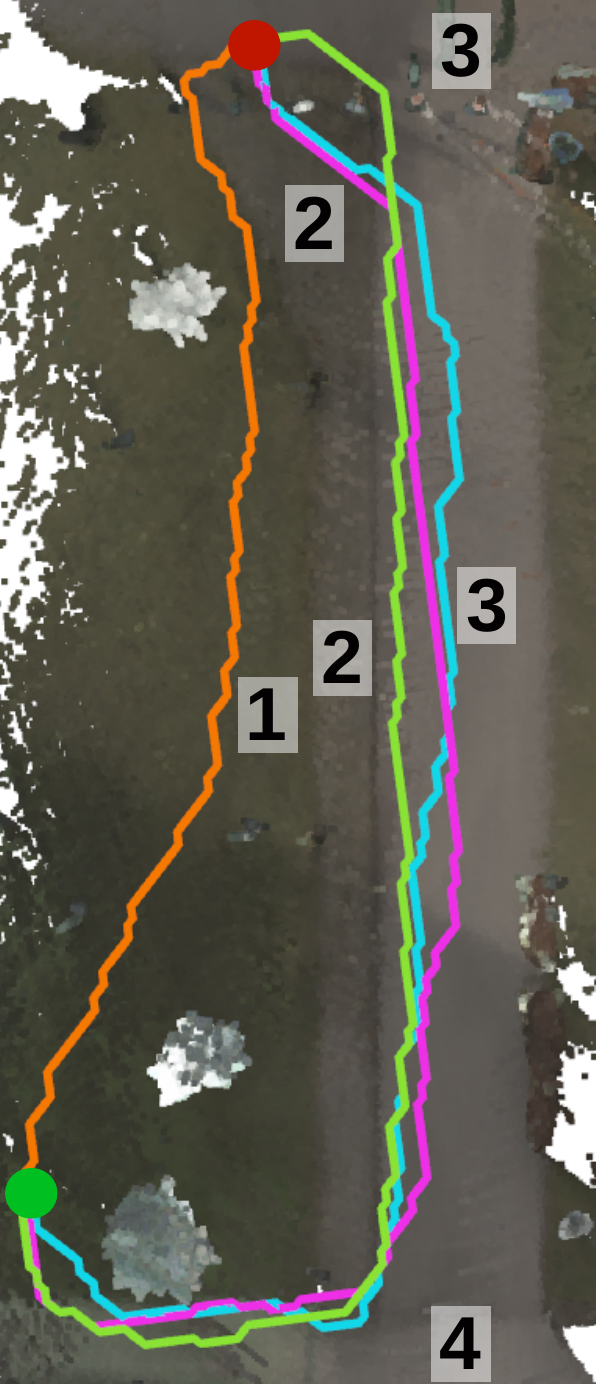}};

    \node[anchor=north west,inner sep=0,xshift=0.15em] (r2c2) at (r2c1.north east)
    {\includegraphics[trim={0 0 0 0},clip,height=4.74cm]
    {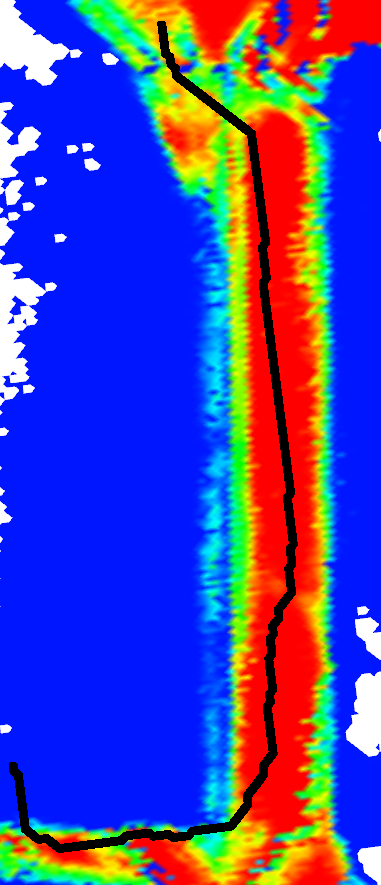}};

    \node[anchor=north west,inner sep=0,xshift=0.15em] (r2c3) at (r2c2.north east)
    {\includegraphics[trim={0 0 0 0},clip,height=4.74cm]
    {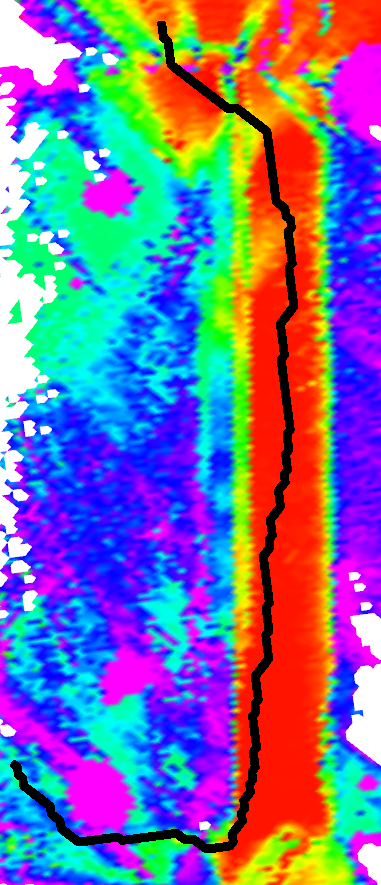}};

    \node[anchor=north west,inner sep=0,xshift=0.15em] (r2c4) at (r2c3.north east)
    {\includegraphics[trim={0 0 0 0},clip,height=4.74cm]
    {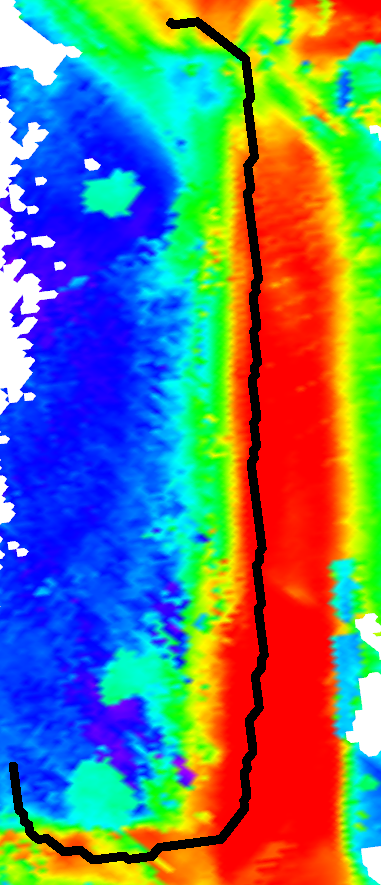}};

    \node[anchor=north west,inner sep=0,yshift=-0.35em] (r3c1) at (r2c1.south west)
    {\includegraphics[trim={0 0 0 0},clip,height=2.31cm]
    {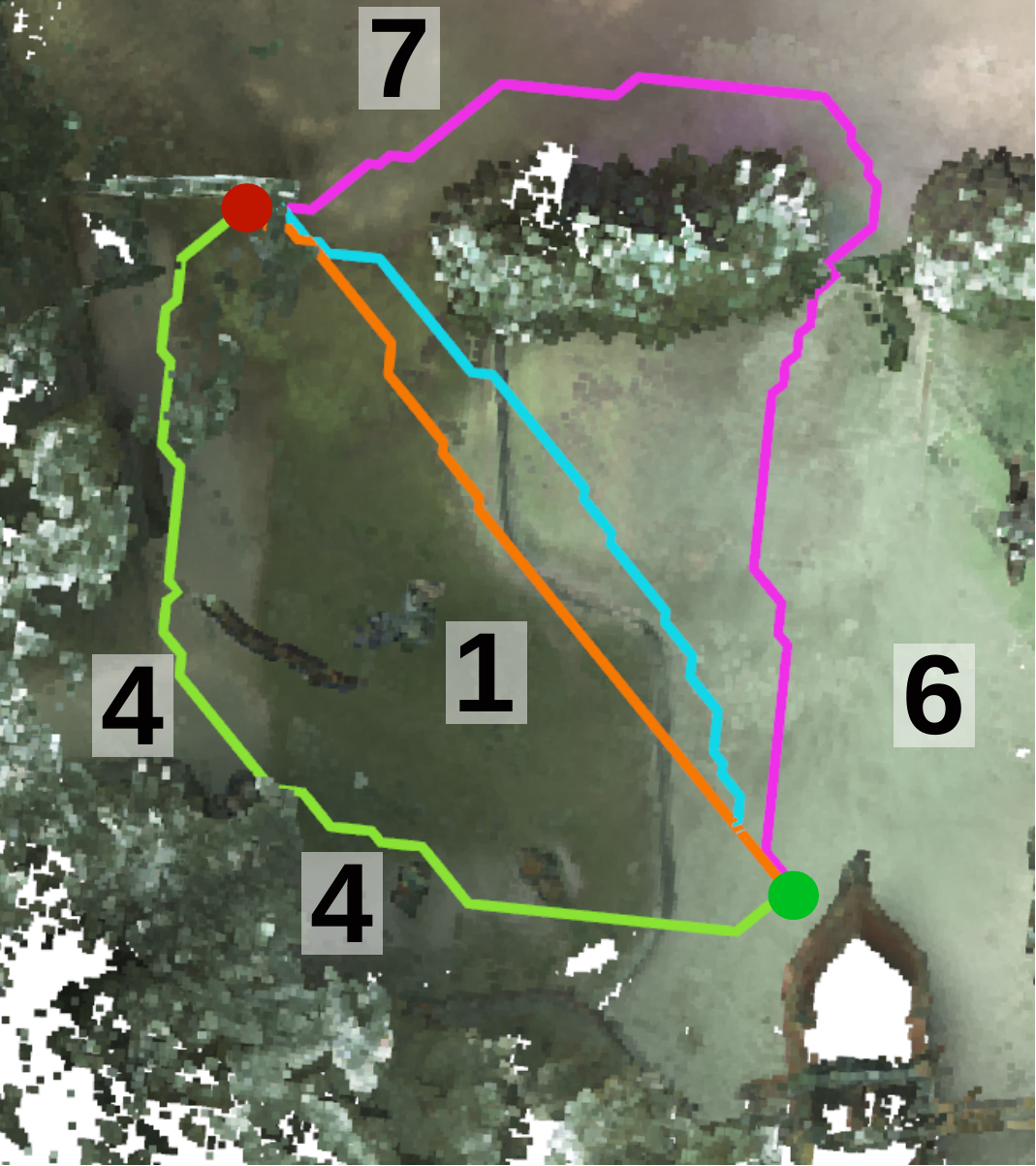}};

    \node[anchor=north west,inner sep=0,xshift=0.15em] (r3c2) at (r3c1.north east)
    {\includegraphics[trim={0 0 0 0},clip,height=2.31cm]
    {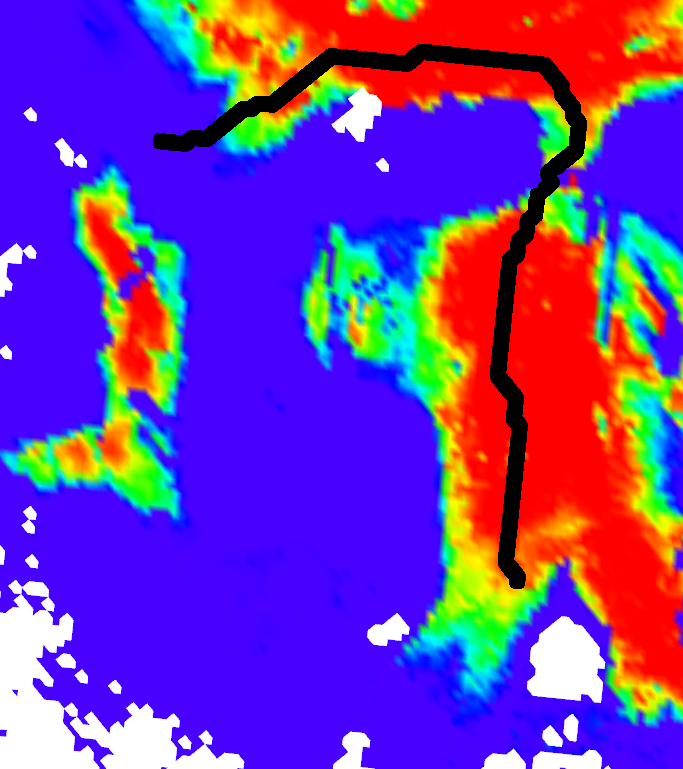}};

    \node[anchor=north west,inner sep=0,xshift=0.15em] (r3c3) at (r3c2.north east)
    {\includegraphics[trim={0 0 0 0},clip,height=2.31cm]
    {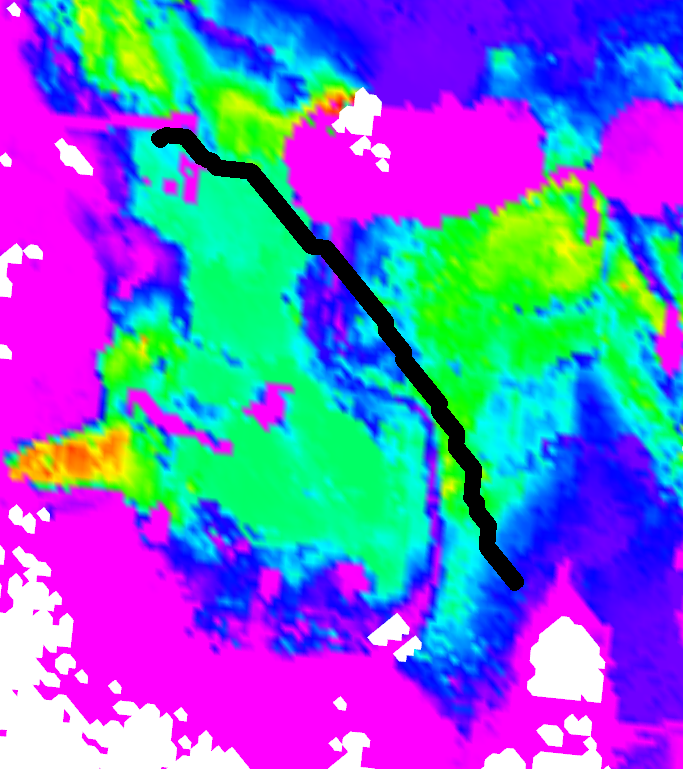}};

    \node[anchor=north west,inner sep=0,xshift=0.15em] (r3c4) at (r3c3.north east)
    {\includegraphics[trim={0 0 0 0},clip,height=2.31cm]
    {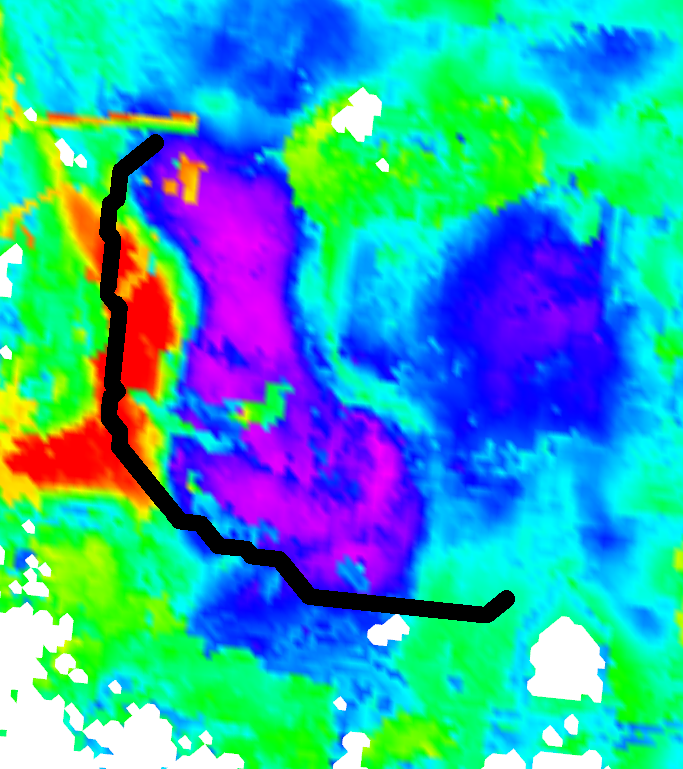}};

    \node[anchor=north,font=\scriptsize\sffamily,yshift=-0.01em] at (r3c1.south) {RGB};
    \node[anchor=north,font=\scriptsize\sffamily,yshift=-0.01em] at (r3c2.south) {WVN};
    \node[anchor=north,font=\scriptsize\sffamily,yshift=-0.01em] at (r3c3.south) {LangSAM};
    \node[anchor=north,font=\scriptsize\sffamily,yshift=-0.01em] at (r3c4.south) {Ours};

    \node[inner sep=0.15em,scale=.7, anchor=south west,yshift=0.em,xshift=0.em, rectangle, align=left, fill=lightgray, font=\sffamily] (n_0) at (r1c1.south west) {E1, Husky};
   \node[inner sep=0.15em,scale=.7, anchor=south west,yshift=-0.5em,xshift=0.em, rectangle, align=left, fill=lightgray, font=\sffamily] (n_0) at (r2c1.south west) {E2, Spot};
   \node[inner sep=0.15em,scale=.7, anchor=south west,yshift=0.em,xshift=0.em, rectangle, align=left, fill=lightgray, font=\sffamily] (n_0) at (r3c1.south west) {E3, Spot};

    \end{tikzpicture}

    \caption{
    Selected paths for Husky-E1 and Spot-E2, E3:
    Euclidean (orange), WVN (pink), LangSAM (light blue), and Ours (light green);
    start (red) to goal (green). 1--grass, 2--cobble w. grass, 3--asphalt, 4--cobblestone, 5--gravel, 6--dirt, 7--sand.
    \textit{E1, Husky}: \rebuttal{WVN rates high-effort gravel (5) like asphalt/cobble, while LangSAM overestimates grass and gravel (1,5), increasing path effort. \net~avoids gravel, yielding the lowest-effort paths.
    \textit{E2, Spot}: Euclidean crosses high-effort grass (1); WVN and LangSAM cross cobble (2) instead of asphalt (3) near the start. \net~avoids cobble and minimizes grass traversal.}
    \textit{E3, Spot}: WVN rates sand (7) similarly to cobble and dirt (4,6), yielding high-effort paths. LangSAM poorly distinguishes terrains. \net~favors cobble and minimizes traversal on sand and grass.}
    \label{fig:maps_combined}
    \vspace{-1em}

\end{figure}

\subsection{Ablation study}\label{sec:ab}

We analyze the component influence on navigation and traversability prediction. Navigation and 2.5D grid map evaluations are on E2 (Spot); 2D results are averaged across E1--E3.\looseness=-1

\subsubsection{Effect of each Loss function for Score Creation}
\rebuttal{\tabref{tab:ab_clustering} shows the impact of each loss function.} We evaluate traversability scores via pairwise terrain overlap, mean score range per location, absolute change in the mean, standard deviation of terrain features during continual learning (stability), Pearson correlation with computed energy-efficiency ranking \rebuttal{and Adjusted Rand Index (ARI)}. $\mathcal{L}_{REC}$ alone can achieve terrain separation, evident in the pairwise overlap, \rebuttal{ARI} and average range scores. 
On both platforms, $\mathcal{L}_{KL}$ reduces the average range while hindering the stability of the feature space; \rebuttal{we retain it as it proves beneficial for avoiding overfitting during VAE training.}
Next, $\mathcal{L}_{VIC}$ improves correlation by $0.46$ on Spot and $0.57$ on Husky by better ordering terrains in feature space, and boosts stability on both platforms, indicating that this loss already divides the feature space favorably during base training ($\mathcal{T}_1$). Finally, $\mathcal{L}_{INC}$ improves stability by $0.06$ on Husky and reduces pairwise terrain overlap by $0.12$ on this platform, though it is less crucial for the stable feature space on Spot. \rebuttal{The moderate-to-good ARI scores show that terrain identity is meaningfully recovered from the latent space.} 

\begin{table}
\centering
\caption{Ablation study on efficacy of \rebuttal{each loss in \net~score clustering. We report pairwise terrain} overlap, average score range per location, feature space stability, Pearson correlation with energy efficiency, and Adjusted Rand Index (ARI) between clustering and terrain labels.}
\label{tab:ab_clustering}
\setlength\tabcolsep{5pt}
\begin{tabular}
{p{0.05cm}p{0.49cm}p{0.43cm}p{0.45cm}p{0.65cm}|p{0.63cm}p{0.60cm}p{0.51cm}p{0.68cm}p{0.42cm}}
 \toprule
&\textbf{$\mathcal{L}_{REC}$}& \textbf{$\mathcal{L}_{KL}$}& \textbf{$\mathcal{L}_{VIC}$}& \textbf{$\mathcal{L}_{INC}$} & Pairwise overlap $\downarrow$& Average Range $\downarrow$& Stab- ility $\uparrow$& Cor- relation $\uparrow$& \rebuttal{ARI $\uparrow$}\\
\midrule
\multirow{4}{*}{$S_i$} & \checkmark & & & & 0.37 & 0.15 & 0.85 & 0.62 & \rebuttal{0.32}\\
& \checkmark & \checkmark & & & 0.35 & \textbf{0.06} & 0.73 & 0.44 & \rebuttal{0.44}\\
& \checkmark & \checkmark & \checkmark & & 0.21 & 0.13 & 0.85 & 0.90 & \rebuttal{0.43}\\
& \checkmark & \checkmark &\checkmark & \checkmark & \textbf{0.21} & 0.13 & \textbf{0.87} & \textbf{0.90} & \rebuttal{\textbf{0.44}}\\
\midrule
\multirow{4}{*}{$H_i$} & \checkmark & & & & 0.25 & 0.04 & \textbf{0.75} & 0.14 & \rebuttal{\textbf{0.58}}\\
& \checkmark & \checkmark & & & 0.25 & \textbf{0.03} & 0.41 & 0.14 & \rebuttal{0.43}\\
& \checkmark & \checkmark & \checkmark & & 0.37 & 0.13 & 0.65 & \textbf{0.71} & \rebuttal{0.47}\\
& \checkmark & \checkmark &\checkmark & \checkmark & \textbf{0.25} & 0.17 & 0.71 & 0.64 & \rebuttal{0.44}\\
\hline
\end{tabular}
\vspace{-0.6cm}
\end{table}

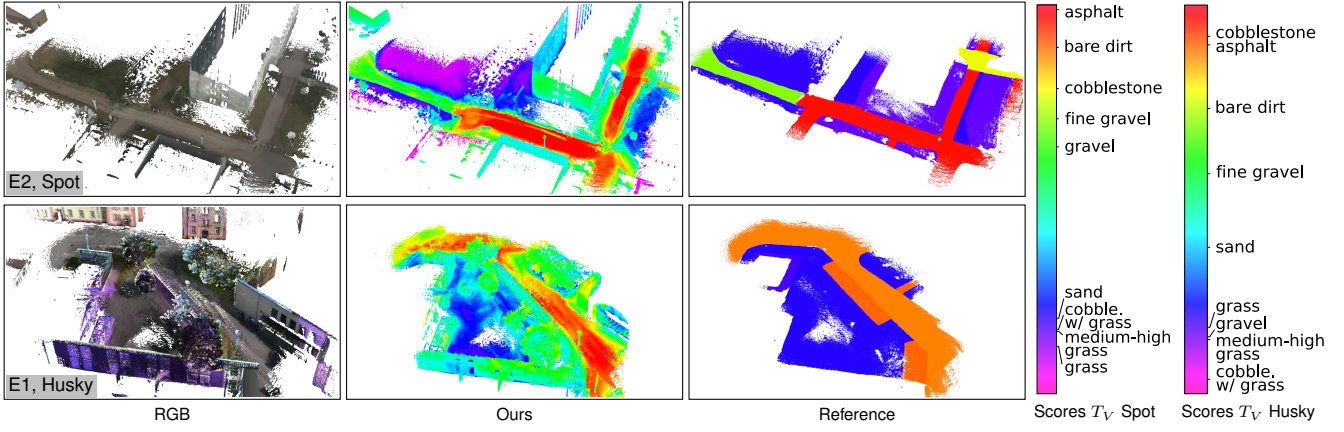
\begin{figure*}[t]
  \centering
  \begin{tikzpicture}
  \node[anchor=north west,inner sep=1, draw] (image) at (0,0){\includegraphics[trim={0 0 0 0},clip,height=2.4cm]{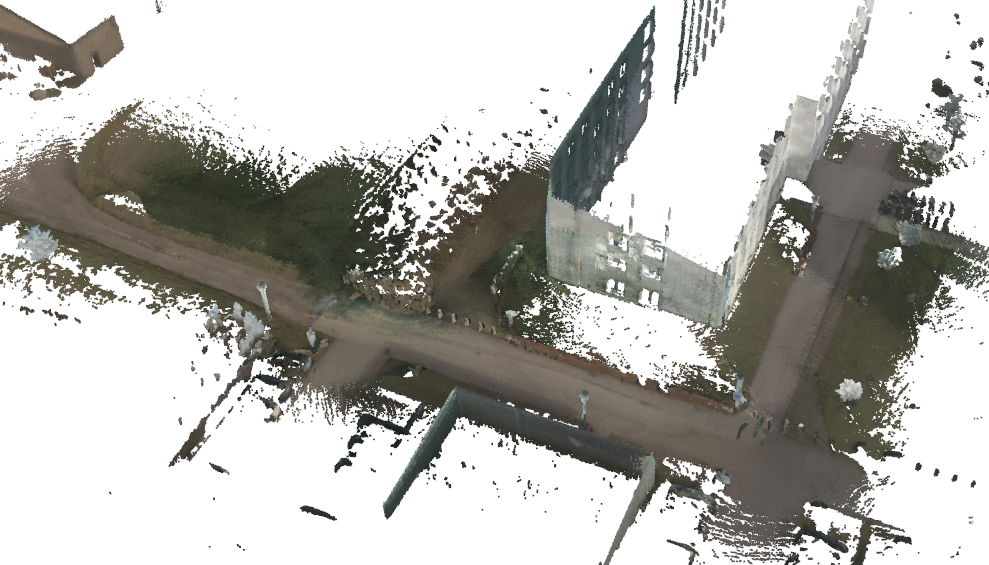}};
  \node[anchor=north west,inner sep=1, draw, xshift=0.2em] (image1) at (image.north east){\includegraphics[trim={0 0 0 0},clip,height=2.4cm]{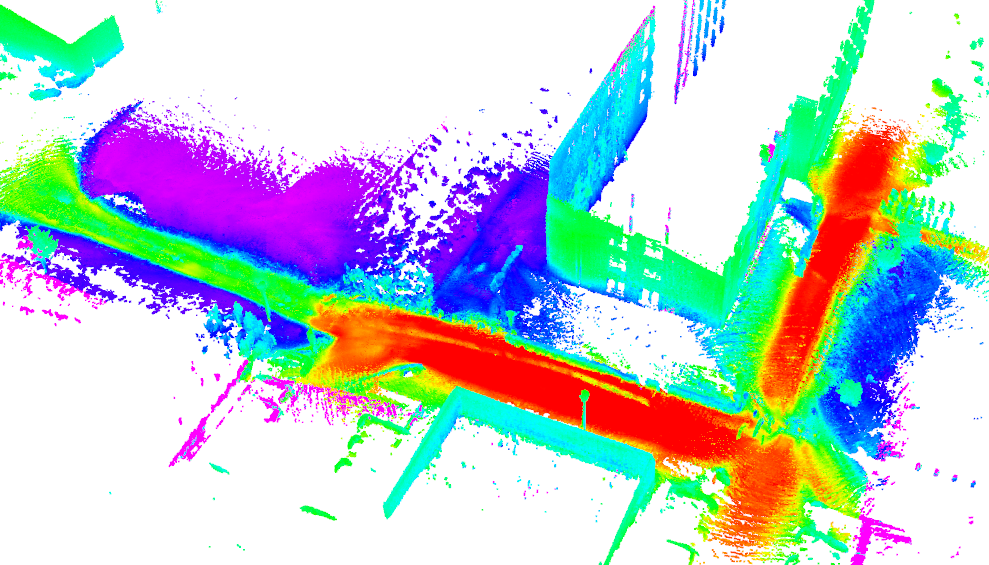}};
  \node[anchor=north west,inner sep=1, draw, xshift=0.2em] (image2) at (image1.north east){\includegraphics[trim={0 0 0 0},clip,height=2.4cm]{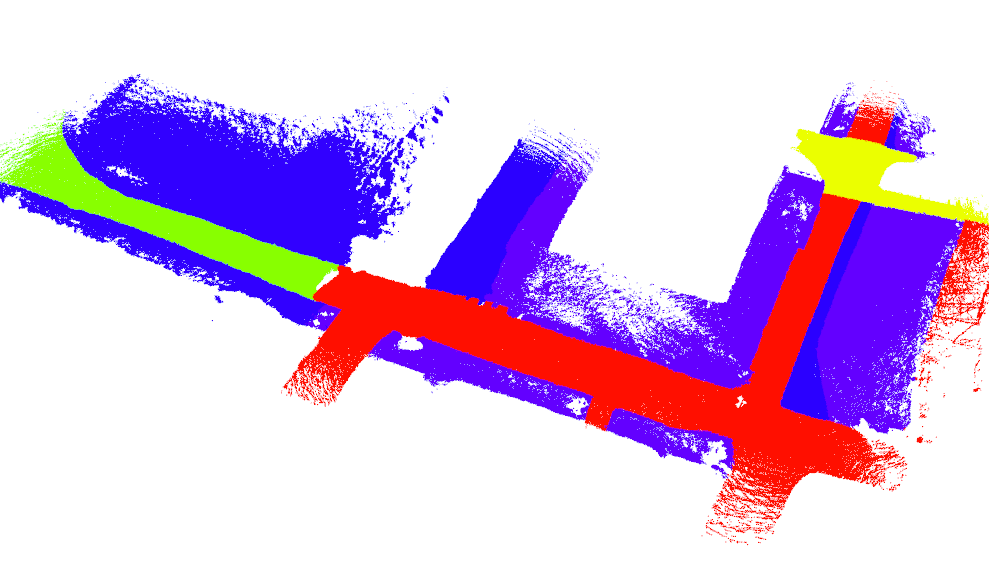}};
  
  \node[anchor=north west,inner sep=1, xshift=0.3em] (imagecb1) at (image2.north east){\includegraphics[trim={0 0 0 0},clip,height=5cm]{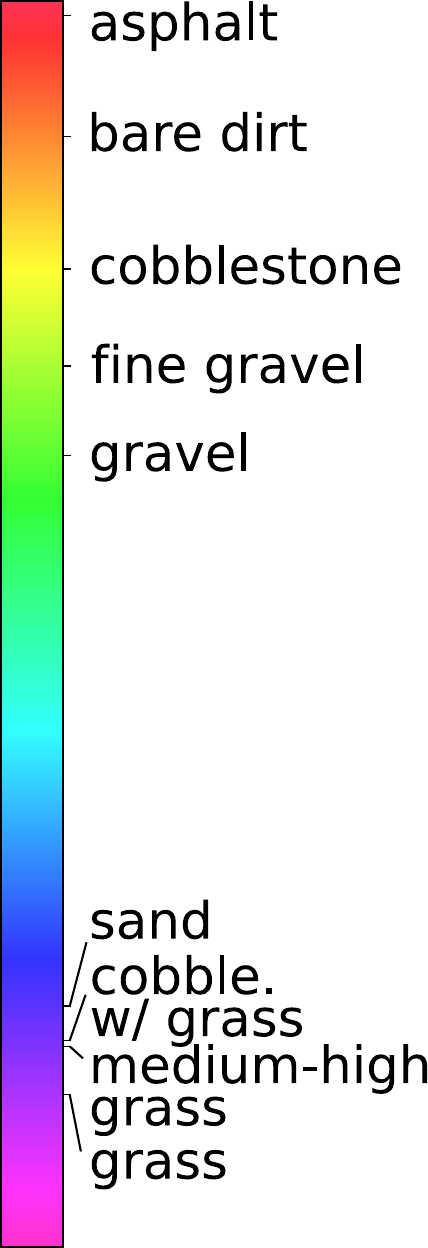}};
  \node[anchor=north west,inner sep=1, xshift=0.3em] (imagecb2) at (imagecb1.north east){\includegraphics[trim={0 0 0 0},clip,height=5cm]{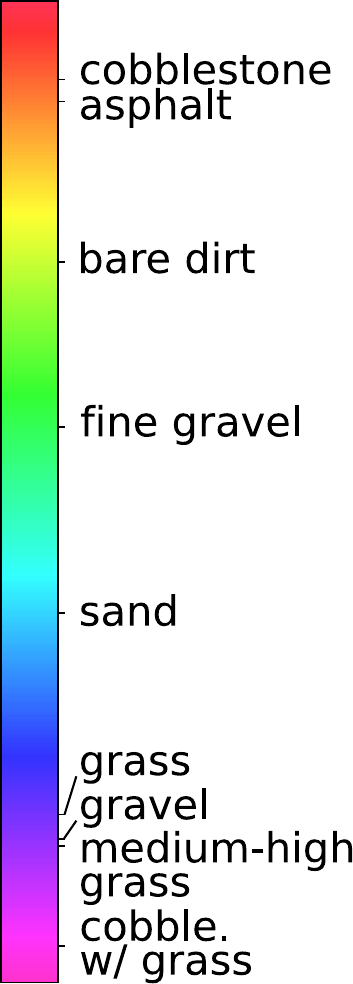}};
  
  \node[anchor=north,inner sep=1, draw, yshift=-0.1cm] (image3) at (image.south){\includegraphics[trim={0 0 0 0},clip,height=2.4cm]{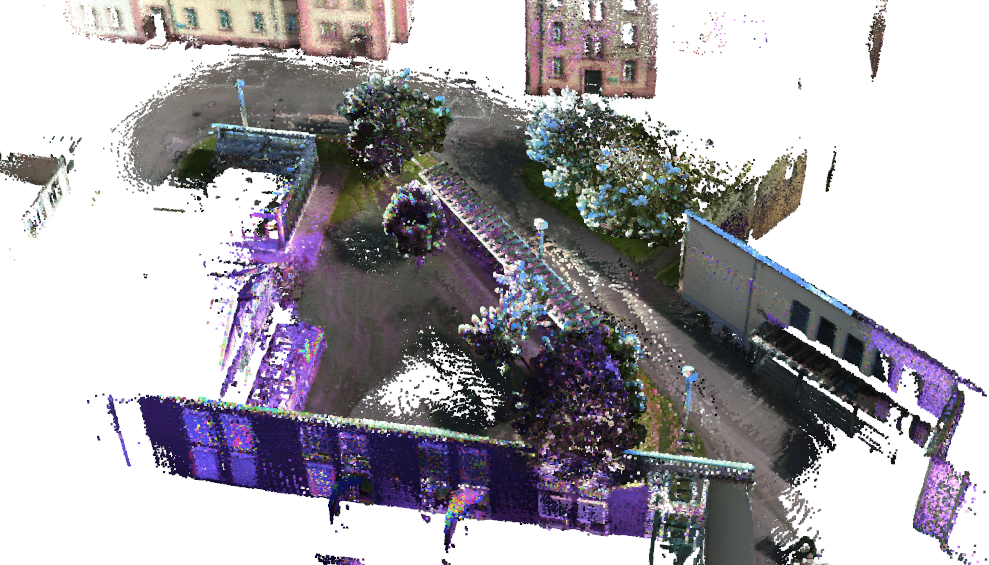}};
  \node[anchor=north west,inner sep=1, draw, xshift=0.2em] (image4) at (image3.north east){\includegraphics[trim={0 0 0 0},clip,height=2.4cm]{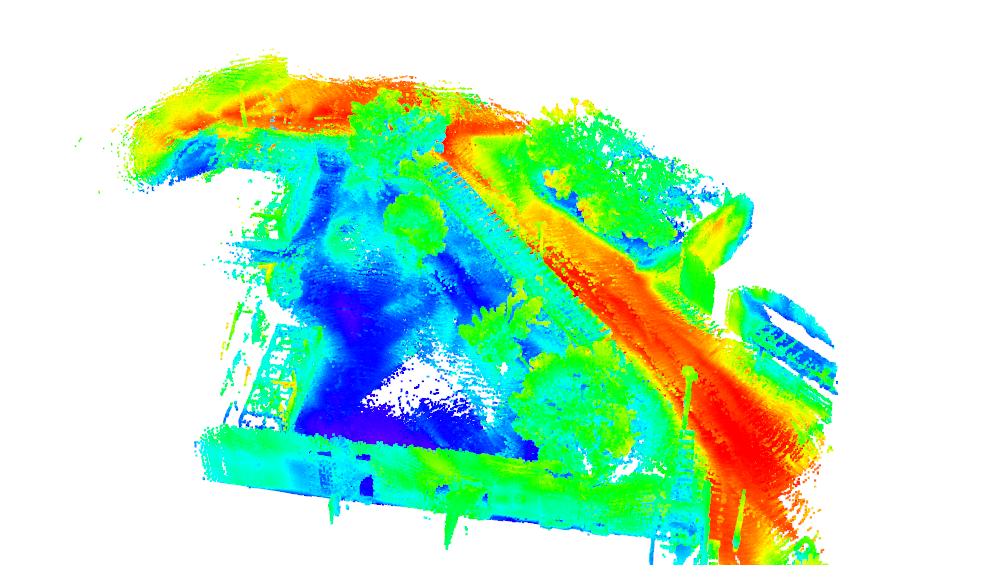}};
  \node[anchor=north west,inner sep=1, draw, xshift=0.2em] (image5) at (image4.north east){\includegraphics[trim={0 0 0 0},clip,height=2.4cm]{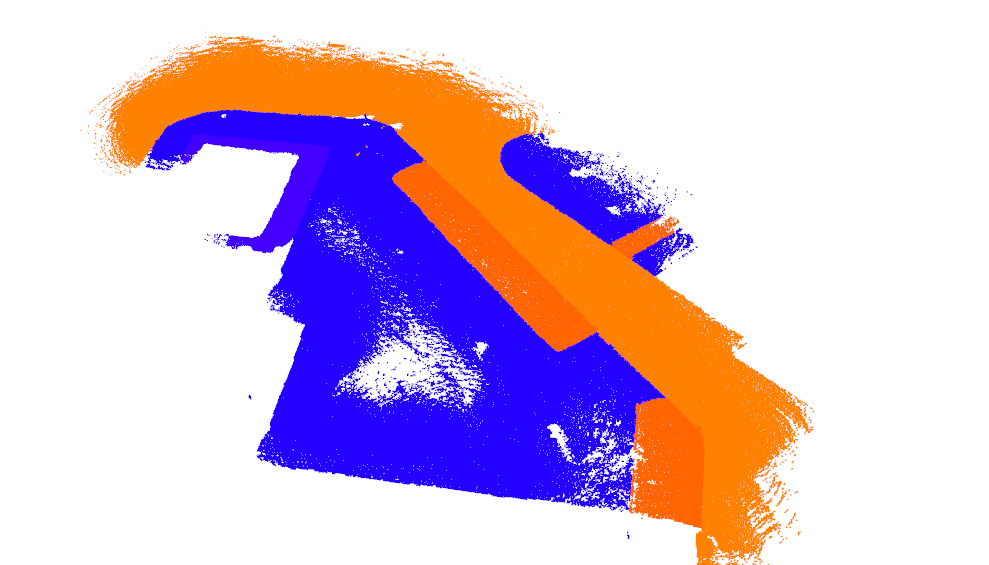}};
  
  \node[inner sep=0.15em,scale=.7, anchor=south west,yshift=0.1em,xshift=0.1em, rectangle, align=left, fill=lightgray, font=\sffamily] (n_0) at (image.south west) {E2, Spot};
  \node[inner sep=0.15em,scale=.7, anchor=south west,yshift=0.1em,xshift=0.1em, rectangle, align=left, fill=lightgray, font=\sffamily] (n_0) at (image3.south west) {E1, Husky};

\node[inner sep=0.,scale=.9, anchor=north,yshift=-0.3em, rectangle, align=left, font=\scriptsize\sffamily] (n_0) at (image3.south) {RGB};
\node[inner sep=0.,scale=.9, anchor=north,yshift=-0.3em, rectangle, align=left, font=\scriptsize\sffamily] (n_1) at (image4.south) {Ours};
\node[inner sep=0.,scale=.9, anchor=north,yshift=-0.3em, rectangle, align=left, font=\scriptsize\sffamily] (n_2) at (image5.south) {Reference};

\node[inner sep=0.,scale=.9, anchor=north west,yshift=-0.3em, rectangle, align=left, font=\scriptsize\sffamily] (n_3) at (imagecb1.south west) {Scores $T_V$ Spot};
\node[inner sep=0.,scale=.9, anchor=north west,yshift=-0.3em, rectangle, align=left, font=\scriptsize\sffamily] (n_4) at (imagecb2.south west) {Scores $T_V$ Husky};

  \end{tikzpicture}
  \caption{Traversability map of E2 (Spot) and E1 (Husky). Our predictions align with the reference, as quantitatively confirmed in \tabref{tab:inc_comparison}. Mean learned traversability score ranges are shown on the right; Spot and Husky largely align, with minor ranking differences. Low-traversability terrains cluster together, while higher-traversability ones are more dispersed.}
  \vspace{-1em}
  \label{fig:maps_3d}
\end{figure*}

\subsubsection{Effect of Traversability Scores and Visual Terrain Masks}
We compare \net~trained with our self-supervised scores to handcrafted scores from prior work in \tabref{tab:T}. Vibration is IMU z-axis PSD over $10.24\,\text{s}$~\cite{howdoesitfeel}, traction follows WVN~\cite{frey2023fast}, and torque uses raw motor current. Our model yields lowest-effort paths, producing smooth, descriptive scores correlated with energy efficiency.
Vibration, Traction, and Raw torque yield unreliable results with lower IoU scores. Motor torque directly relates to energy efficiency but is highly noisy over short intervals, \rebuttal{providing weak supervision.}
Vibration and traction correlate poorly with energy efficiency, \rebuttal{fail to minimize effort, and rank orthogonally to our metrics.}

We further quantitatively evaluate navigation performance of \net~with different terrain masks $T_M$, recording improvements of at least \rebuttal{$2.3$} on effort and $0.1$ on EPL in \tabref{tab:T}. Our \rebuttal{mask generation} outperforms STEGO and FastSAM by at least $ 13.4$pp on both IoU metrics. FastSAM focuses on segmenting small surrounding objects, while STEGO struggles with precise boundary estimation on uneven terrain (e.g., cobblestone with grass), as shown in \figref{fig:TM}.

\begin{table}
\centering
\caption{Influence of self-supervised traversability score $T_S$ and terrain mask $T_M$ on navigation performance (Effort, EPL) and prediction quality (2D segm., 2.5D segm.). mIoU scores are recorded in [\%].}
\label{tab:T}
\setlength\tabcolsep{4pt}
\begin{tabular}
{p{2.2cm}|p{1.4cm}p{1.5cm}p{1cm}p{1cm}}
 \toprule
$T_S$ & Effort $\downarrow$& EPL $\downarrow$ & 2.5D segm.$\uparrow$ & 2D segm. $\uparrow$\\
\midrule
Vibration (PSD)~\cite{howdoesitfeel} & \rebuttal{6.53 $\pm$ 2.03} & \rebuttal{0.167 $\pm$ 0.07} &  \phantom{0}2.16 & \phantom{0}1.01\\ 
Traction(WVN)~\cite{frey2023fast} & \rebuttal{6.18 $\pm$ 1.99} & \rebuttal{0.156 $\pm$ 0.06} & 16.70 & \phantom{0}7.93 \\ 
Torque & \rebuttal{7.98 $\pm$ 1.76} & \rebuttal{0.211 $\pm$ 0.06} & 25.96 & 18.10\\ 
COTRATE (Ours) &  \textbf{\rebuttal{\textbf{3.28} $\pm$ 0.87}} & \textbf{\rebuttal{0.075 $\pm$ 0.01}} & \textbf{39.40} & \textbf{33.23}\\ 
\midrule
$T_M$ & Effort $\downarrow$& EPL $\downarrow$ & 2.5D segm.$\uparrow$ & 2D segm. $\uparrow$\\
\midrule
FastSAM~\cite{zhao2023fast} & \rebuttal{5.69 $\pm$ 1.24} & \rebuttal{0.198 $\pm$ 0.07}  & 10.91 & \phantom{0}9.52\\
STEGO~\cite{stego} & \rebuttal{5.58 $\pm$ 1.43} & \rebuttal{0.178 $\pm$ 0.04} & 19.54 & 19.83\\
COTRATE (Ours) & \textbf{\rebuttal{\textbf{3.28} $\pm$ 0.87}} & \textbf{\rebuttal{0.075 $\pm$ 0.01}}   & \textbf{39.40} & \textbf{33.23}\\
\hline
\end{tabular}
\vspace{-1em}
\end{table}

\subsubsection{Ablation of Visual Model Components}
We ablate the architectural novelties in \tabref{tab:ab_elem}. We attach a DeepLabv3 decoder with class dimension $1$ to DINOv3 \rebuttal{and train it} with an MSE-based loss (row 1). Training a larger decoder for direct regression increases effort by \rebuttal{$88\%$ vs.} $\mathcal{L}_\text{align}$. This indicates better generalization with our alignment loss in higher-dimensional spaces. Selective feature replay improves mIoU by $24.5$pp (2D) and $27.1$pp (2.5D), mitigating catastrophic forgetting. FCM further improves EPL by \rebuttal{$0.08$} and effort by \rebuttal{$2.83$} by injecting replayed features into unannotated regions in this semi-supervised learning setting.

\begin{table}[t]
\centering
\caption{Ablation study on the efficacy of various components of \net. mIoU is reported in [\%]. The first row uses the DeepLabv3 decoder with 1 class dimension. FCM denotes the feature cut-mix strategy (\secref{sec: inc}).}
\label{tab:ab_elem}
\setlength\tabcolsep{3pt}
\begin{tabular}
{p{0.7cm}p{0.7cm}p{0.6cm}|p{1.5cm}p{1.5cm}p{1cm}p{1cm}}
 \toprule
$\mathcal{L}_\text{align}$ & $\mathcal{L}_\text{replay}$ & FCM & Effort $\downarrow$& EPL $\downarrow$ & 2.5D segm.$\uparrow$ & 2D segm. $\uparrow$\\
\midrule
&  & & \rebuttal{15.41 $\pm$ 0.75} & \rebuttal{0.439 $\pm$ 0.03}& \phantom{0}6.29 & \phantom{0}5.46 \\ 
\checkmark & & &  \rebuttal{8.19 $\pm$ 1.40} & \rebuttal{0.235 $\pm$ 0.05} & 10.10 & \phantom{0}7.98 \\ 
\checkmark & \checkmark & & \rebuttal{6.11 $\pm$ 1.42} & \rebuttal{0.159 $\pm$ 0.04} & 37.16 & 32.51 \\ 
\checkmark & \checkmark & \checkmark &  \textbf{\rebuttal{\textbf{3.28} $\pm$ 0.87}} & \textbf{\rebuttal{0.075 $\pm$ 0.01}} & \textbf{39.40} & \textbf{33.23} \\ 
\hline
\end{tabular}
\vspace{-0.5cm}
\end{table}

\subsubsection{\rebuttal{Replay Interval, Update and Buffer Size}} \label{sec:replay}
We ablate settings of our feature replay on 2.5D and 2D mIoU metrics in \tabref{tab:r_size}.
Increasing the replay interval to $10$ iterations lowers 2.5D and 2D mIoU by $21$pp and $17.3$pp, indicating increased forgetting. Nonetheless, replaying every $100$ iterations still improves performance by $5.9$pp (2.5D) over no replay, highlighting the effectiveness of our feature selection strategy.
\rebuttal{We also} vary the feature replay buffer size $B^R$ and observe that $20$ features cannot capture terrain diversity, resulting in a drop of at least $ 4.91$pp in both metrics. While $2000$ features preserve 2D performance, it lowers 2.5D mIoU by $7.2$pp, likely because excess features distract the model from new terrains.
\rebuttal{Finally, $100$ iterations is} the optimal update interval $t_b$ for the replay buffer. While overly frequent updates ($t_b$=10) impose instability in the furthest point sampling ($\downarrow 8.35$pp), few updates ($t_b$=1000) can lead to a deferred inclusion of terrain knowledge for replay, \rebuttal{decreasing 2.5D mIoU by} $20.44$pp.

\begin{table}[t]
\centering
\caption{Influence of replay interval, replay buffer size and replay update frequency on segmentation metrics. mIoU is reported in [\%].}
\label{tab:r_size}
\begin{tabular}
{c|ccc}
 \toprule
Replay \rebuttal{interval} $t_r$ & 2.5D segm. $\uparrow$& 2D segm. $\uparrow$\\
\midrule
0 & 10.10 & 7.98 \\
1 & \textbf{39.40} & \textbf{33.23} \\
10 & 18.37 & 15.97 \\
100 & 16.04 & 12.70 \\
\midrule
Replay buffer size $B_R$ & 2.5D segm. $\uparrow$& 2D segm. $\uparrow$\\
\midrule
20 & 31.44 & 28.32 & \\
200 & \textbf{39.40} & \textbf{33.23} &\\
2000 & 32.19 & 31.01 & \\
\midrule
Replay update freq. $t_b$ & 2.5D segm. $\uparrow$& 2D segm. $\uparrow$\\
\midrule
10 & 31.05 & 32.46 & \\
100 & \textbf{39.40} & \textbf{33.23} &\\
1000 & 18.96 & 19.72 & \\
\hline
\end{tabular}
\vspace{-0cm}
\end{table}

\subsubsection{\rebuttal{Rough Terrain and Platform Transfer}}

\begin{figure}[t]
    \centering

    \begin{tikzpicture}


    \node[anchor=north west,inner sep=0] (r1c1) at (0,0)
    {\includegraphics[trim={0 0 0 0},clip,height=2.4cm]
    {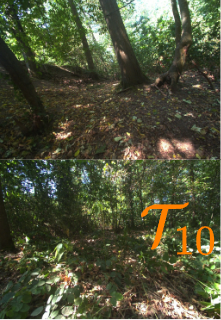}};
    \node[inner sep=0.15em,scale=.7, anchor=south west,yshift=0.em,xshift=0.em, rectangle, align=left, fill=lightgray, font=\sffamily] (n_0) at (r1c1.south west) {Sample Imgs.};

    \node[anchor=north west,inner sep=0] (r1c2) at (r1c1.north east)
    {\includegraphics[trim={0 0 0 0},clip,height=2.4cm]
    {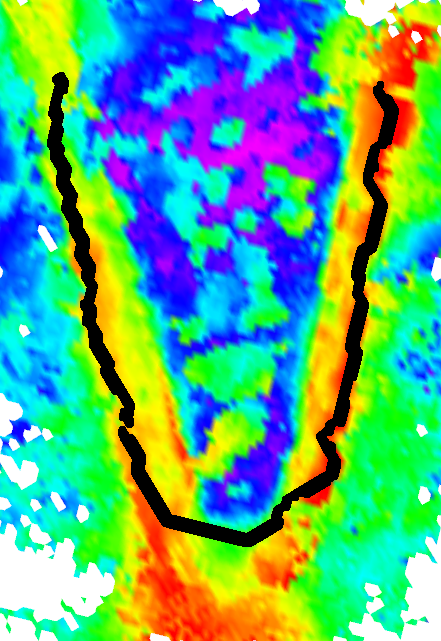}};
    \node[inner sep=0.15em,scale=.7, anchor=south west,yshift=0.em,xshift=0.em, rectangle, align=left, fill=lightgray, font=\sffamily] (n_0) at (r1c2.south west) {Spot};

    \node[anchor=north west,inner sep=0,xshift=0.15em] (r1c3) at (r1c2.north east)
    {\includegraphics[trim={0 0 0 0},clip,height=2.4cm]
    {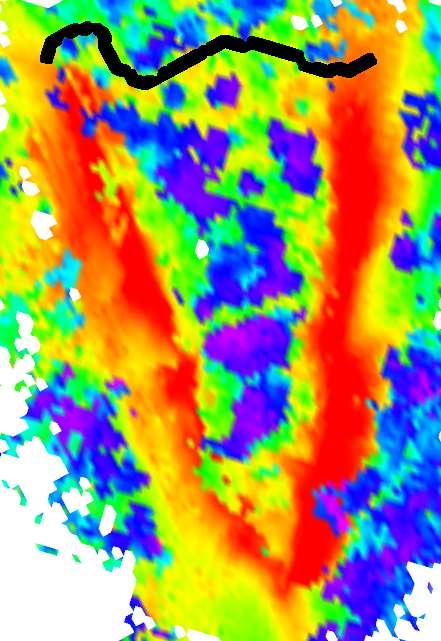}};
    \node[inner sep=0.15em,scale=.7, anchor=south west,yshift=0.em,xshift=0.em, rectangle, align=left, fill=lightgray, font=\sffamily] (n_1) at (r1c3.south west) {Husky};

    \node[anchor=north west,inner sep=0,xshift=0.15em] (r1c4) at (r1c3.north east)
    {\includegraphics[trim={0 0 0 0},clip,height=2.4cm]
    {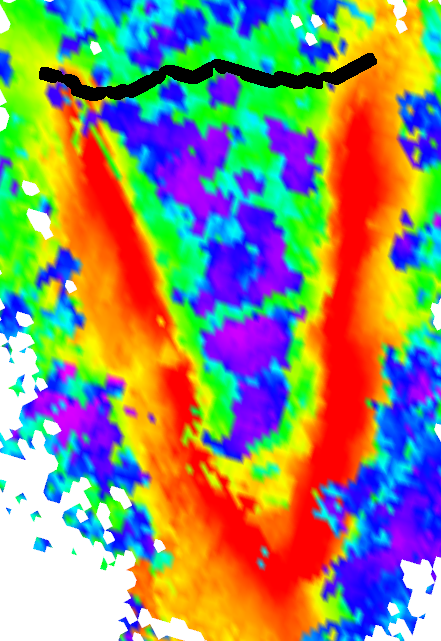}};
    \node[inner sep=0.15em,scale=.7, anchor=south west,yshift=0.em,xshift=0.em, rectangle, align=left, fill=lightgray, font=\sffamily] (n_2) at (r1c4.south west) {Spot $\rightarrow$ Husky};

    \node[anchor=north west,inner sep=0,xshift=0.15em] (r1c5) at (r1c4.north east)
    {\includegraphics[trim={0 0 0 0},clip,height=2.4cm]
    {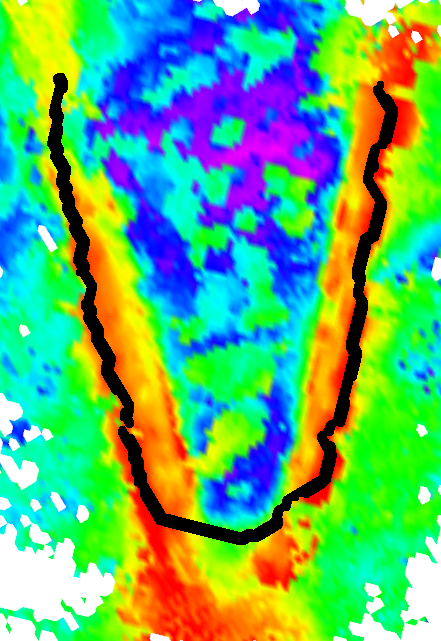}};
    \node[inner sep=0.15em,scale=.7, anchor=south west,yshift=0.em,xshift=0.em, rectangle, align=left, fill=lightgray, font=\sffamily] (n_3) at (r1c5.south west) {Husky $\rightarrow$ Spot};

  \end{tikzpicture}
  \caption{Platform transfer on rough forest terrain: sample images and 2.5D maps with traversability predictions by the resp. robot's original and transferred models.
Forest (center of the map) has high cost for Spot (blue shades) and medium for Husky (green to orange shades), making it prefer the shortcut. Preferences transfer successfully via continued training on forest.
}
  \vspace{-1em}
  \label{fig:maps_forest}
\end{figure}

\rebuttal{To demonstrate application in rough, uneven terrain, we additionally train our models on forest terrain and evaluate on a test environment comprising a U-shaped fine gravel path flanked by forest (\figref{fig:maps_forest}).
Husky drives over low branches without much effort, but Spot's legs get stuck, making traversal extremely hard. This is correctly reflected in the learned scores: medium for Husky, low for Spot. Consequently, Husky prefers a shortcut through sparse forest, while Spot goes around via fine gravel.
We further adapt each robot's model to the other platform via continued training on the corresponding forest data, simulating a setting where one robot takes another's collected experience and corrects differing preferences.
Our replay mechanism prevents catastrophic forgetting while adapting to contradictory signals, as periodically regenerated pseudo-labels gradually shift replay-buffer features toward the new preference.}

\subsubsection{Qualitative Results}
We qualitatively evaluate map traversability predictions for \rebuttal{E2 (Spot) and E1 (Husky)}, in \figref{fig:maps_3d}. \net~robustly differentiates similar terrains, e.g., asphalt vs. fine gravel in E2, \rebuttal{showing} the benefit of our fine-grained traversability score module. For E1, Husky ranks terrains correctly, though separation between gravel and grass is less pronounced. \rebuttal{Comparing paths to baselines in \figref{fig:maps_combined},} \net~selects the most energy-efficient routes, favoring asphalt or cobblestone while WVN and LangSAM navigate longer on high-effort gravel (E1),  cobble w. grass (E2) or sand (E3). \rebuttal{Further, \net's scores are smoother for grass than LangSAM's.} Videos and maps are in the multimedia material.


\section{Conclusion}
We present \net, a novel self-supervised, robot-agnostic, online visual traversability estimation method. Our approach first learns robust terrain separation from proprioceptive and inertial sensor data using non-contrastive representation learning in a Variational Auto-Encoder. Traversability scores are computed by comparing encoded sensor data to smooth terrain features. 
We generate conclusive terrain masks from DINOv3 visual features and align them with traversability scores using our novel alignment loss. To mitigate forgetting in online learning scenarios, we introduce selective visual feature replay.
Experiments on three environments across two robot platforms show that our method reduces path effort versus existing self-supervised and open-vocabulary baselines.
To the best of our knowledge, this is among the first works to address self-supervised, continual traversability estimation in off-road environments.\looseness=-1


{\footnotesize
\bibliographystyle{IEEEtran}
\bibliography{references}
}

\ifdefined\SUP
\clearpage
\renewcommand{\baselinestretch}{1}

\begin{strip}
\begin{center}
\vspace{-5ex}

\textbf{\Large \bf
Self-Supervised Online Robot-Agnostic Traversability Estimation for Open-World Environments} \\
\vspace{3ex}

\Large{\bf- Supplementary Material -}\\
\vspace{0.4cm}
\normalsize{Julia Hindel, Simon Bultmann, Houman Masnavi, Daniele Cattaneo and Abhinav Valada}
\end{center}
\end{strip}

In this supplementary material, we present additional analysis on our score creation process
in \secref{s1}. Further, we show additional quantitative and qualitative results in \secref{quan1} and \secref{qual1}, respectively.

\section{Score Creation}\label{s1}

\subsection{Hyperparameters}
We detail additional hyperparameters in \tabref{tab:hyperparams}. For Spot, all sensors are operating at least at $10\,\text{Hz}$, so a window size of $100$ (i.e. $1\,\text{s}$) is sufficient to capture terrain-specific sensor information.
For Husky, which has sensors operating at different rates down to $1\,\text{Hz}$, we use a window size of $W = 1000$ (i.e., $10\,\text{s}$) to capture sufficient temporal correlations. After we compute the traversability scores per timestamp, we set the terrain score to be the minimum within a $2.5\,\text{s}$ window for spot and $10\,\text{s}$ time window for husky to obtain robust scores.
We use the proposed hyperparameter settings for $\mathcal{L}_{VIC}$. Further loss hyperparameters, temporal settings, and other training details are tuned using Optuna on Husky.
The hyperparameter optimization objective is to minimize

\begin{equation}
\mathcal{L}_{\text{hyper}} =
\,\mathcal{O}_{\text{pairwise}}
\;+\;
\,\mathcal{R}_{\text{location}}
\;+\;
\,(1 - \mathcal{S})
\;+\;
\,(1 - \rho_{\text{energy}}),
\end{equation}
where $\mathcal{O}_{\text{pairwise}}$ denotes the pairwise overlap between terrain score distributions,
$\mathcal{R}_{\text{terrain}}$ is the mean score range per location,
$\mathcal{S}$ represents the stability metric,
and $\rho_{\text{energy}}$ is the Pearson correlation with the terrain-specific effort. We apply equivalent hyperparameters for loss and training details for Spot, but set a platform-specific learning rate due to the varied number of sensor inputs. 

\begin{table}[t]
\centering
\caption{Hyperparameters for score creation VAE on each platform. Learning rates and input dimensions differ per platform. Otherwise, identical parameters are used.}
\label{tab:hyperparams}
\begin{tabular}{lcc}
\hline
\textbf{Parameter} & \textbf{Spot} & \textbf{Husky} \\
\hline
\multicolumn{3}{c}{\textit{Input Dimensions}} \\
\hline
$W$ & 100 & 1000 \\
$S$ & 129 & 33 \\
\hline

\multicolumn{3}{c}{\textit{Architecture}} \\
\hline
Kernel sizes      & (11, 7) & (11, 7) \\
Channels          & (64, 64) & (64, 64) \\
Latent Dimension  & 16 & 16 \\
\hline

\multicolumn{3}{c}{\textit{Loss hyperparameters}} \\
\hline
$\alpha$ & 16  & 16   \\
$\beta$  & 16 & 16 \\
$\gamma$ & 3 & 3 \\
\hline

\multicolumn{3}{c}{\textit{$\mathcal{L}_{VIC}$ settings}} \\
\hline
$H$ [s]  & 1.6  & 1.6 \\
$\lambda$ & 25 & 25 \\
$\mu$    & 25 & 25 \\
$\nu$    & 1 & 1 \\
\hline

\multicolumn{3}{c}{\textit{Training}} \\
\hline
Batch size    & 128 & 128  \\
Base Epochs        & 13   & 13   \\
Base Learning rate & 0.00006 & 0.0016 \\
Online Epochs & 1 & 1 \\
Online Learning rate & $6\times 10{^-7}$  & $1.6\times 10{^-5}$ \\
\hline

\end{tabular}
\end{table}

\subsection{Qualitative Analysis of VAE Feature Space}
We visualize the t-SNE projection of the $16$-dimensional feature space in \figref{fig:data}.
We observe that different terrains are consistently separated.
Although not all terrains form tight clusters, their relative positions in the 2D projection reflect similarities to asphalt, our reference terrain.
In particular, paved terrains appear closer to asphalt than grass, consistent with their higher cosine similarity in the original 16-dimensional space.

\begin{figure}[t]
    \centering
    \includegraphics[width=0.37\textwidth]{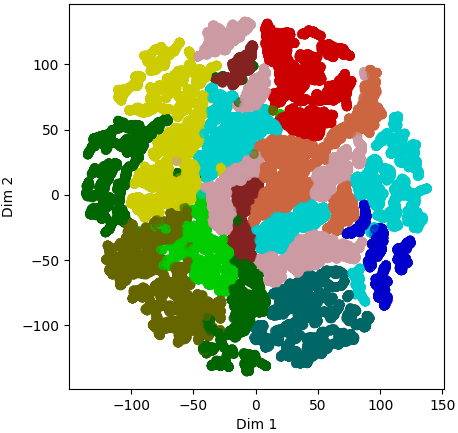} 
    \includegraphics[width=0.08\textwidth]{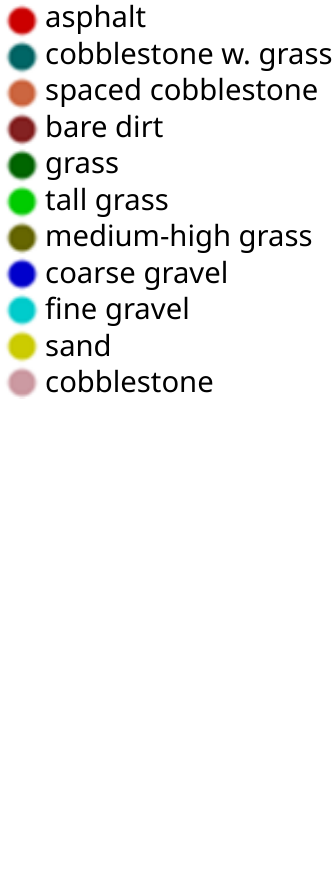}
    \caption{t-SNE visualization of the feature space learned by our VAE. Relative cluster positions reflect similarity to anchor terrain \textit{asphalt}.}
    \label{fig:data}
    \vspace{-0.3cm}
\end{figure}

\subsection{Effect of Each Sensor}

Next, we show the impact of removing selected sensors from the score creation, as shown in \tabref{tab:ab_sensors}. We report the best trade-off when all sensors are included. Further, we observe that removing the raw IMU signals has little impact on the metrics. On the other hand, removing feet or torque significantly reduces the correlation with the terrain-specific effort. 
Moreover, stability is partially affected, indicating that the feature space structure becomes less consistent when foot inputs are missing. Consequently, we argue that proprioceptive signals should be combined to obtain robust effort-correlated traversability scores. 

\begin{table}
\centering
\caption{Ablation study on the efficacy of removing various sensor information of \net~from score clustering for the Spot robot. We record metrics for pairwise overlap across different terrains, average score ranges per location, feature-space stability, and Pearson correlation with energy efficiency. Vel: Velocity from Odometry, Cmd Vel: Command velocity.}
\label{tab:ab_sensors}
\begin{tabular}
{p{0.3cm}p{0.3cm}p{0.4cm}p{0.4cm}p{0.4cm}p{0.5cm}|p{0.62cm}p{0.62cm}p{0.62cm}p{0.62cm}}
 \toprule
IMU&Joint&Feet&Tor que&Cmd Vel & Vel & Pairwise overlap $\downarrow$& Average Range $\downarrow$& Stability $\uparrow$& Cor- relation $\uparrow$\\
\midrule
& \checkmark & \checkmark & \checkmark &\checkmark & \checkmark & 0.21 & 0.13 & 0.87 & 0.90 \\
\checkmark & & \checkmark & \checkmark &\checkmark & \checkmark & 0.22 & 0.14 & 0.83 & 0.90 \\
\checkmark & \checkmark & & \checkmark &\checkmark & \checkmark & 0.47 & \textbf{0.05} & 0.73 & 0.17 \\
\checkmark & \checkmark & \checkmark & &\checkmark & \checkmark & 0.34 & 0.23 & \textbf{0.95} & 0.62 \\
\checkmark & \checkmark & \checkmark & \checkmark & & \checkmark & \textbf{0.18} & 0.12 & 0.81 & 0.90 \\
\checkmark & \checkmark & \checkmark & \checkmark & \checkmark & & 0.22 & 0.15 & 0.81 & 0.90 \\ 
\checkmark & \checkmark & \checkmark & \checkmark & \checkmark & \checkmark & 0.21 & 0.13 & 0.87 & \textbf{0.90}\\
\bottomrule
\end{tabular}
\vspace{-0.3cm}
\end{table}

\subsection{Comparison with Other Methods}

We compare different score creation methods in \figref{fig:score_comp}, showing their respective value ranges on the training terrains. Quantitative results of path planning and segmentation metrics are shown in the main paper. Our score creation module produces scores that follow the trend of the reference terrain-specific effort metric. However, it does not produce the correctly correlated scores for medium-high grass and spaced cobblestone w. grass on Spot and tall grass, coarse gravel and paved cobblestone w. grass on Husky. We posit that our score creation module is self-supervised, and soft or diverse terrains attenuate various sensor measurements, making it challenging to discern these differences. For Husky, vibration in the z-axis is very dominant for coarse gravel, which steers the feature space towards a vibration-dependent division. Further, we note that raw torque is very noisy, which can be seen from the large, overlapping ranges produced for each terrain. 
 Next, the traction method as introduced in WVN~\citeS{frey2023fast} doesn't produce terrain-aware scores. 
 Despite only subtle terrain differences for vibration (PSD), we observe comparatively large discrepancies in performance for tall grass on Spot and fine gravel on Husky.
 Further, vibration does not correlate with our effort-based reference metric for navigation energy efficiency.

\begin{figure*}
  \centering
  \footnotesize
  \setlength{\tabcolsep}{0.05cm}
  {\renewcommand{\arraystretch}{1.1}
    \begin{tabular}{p{1.1cm}p{8.5cm}p{8.5cm}}  

      & \multicolumn{1}{c}{Spot} & \multicolumn{1}{c}{Husky} \\

      \textbf{Reference Effort} &
      \includegraphics[width=\linewidth]{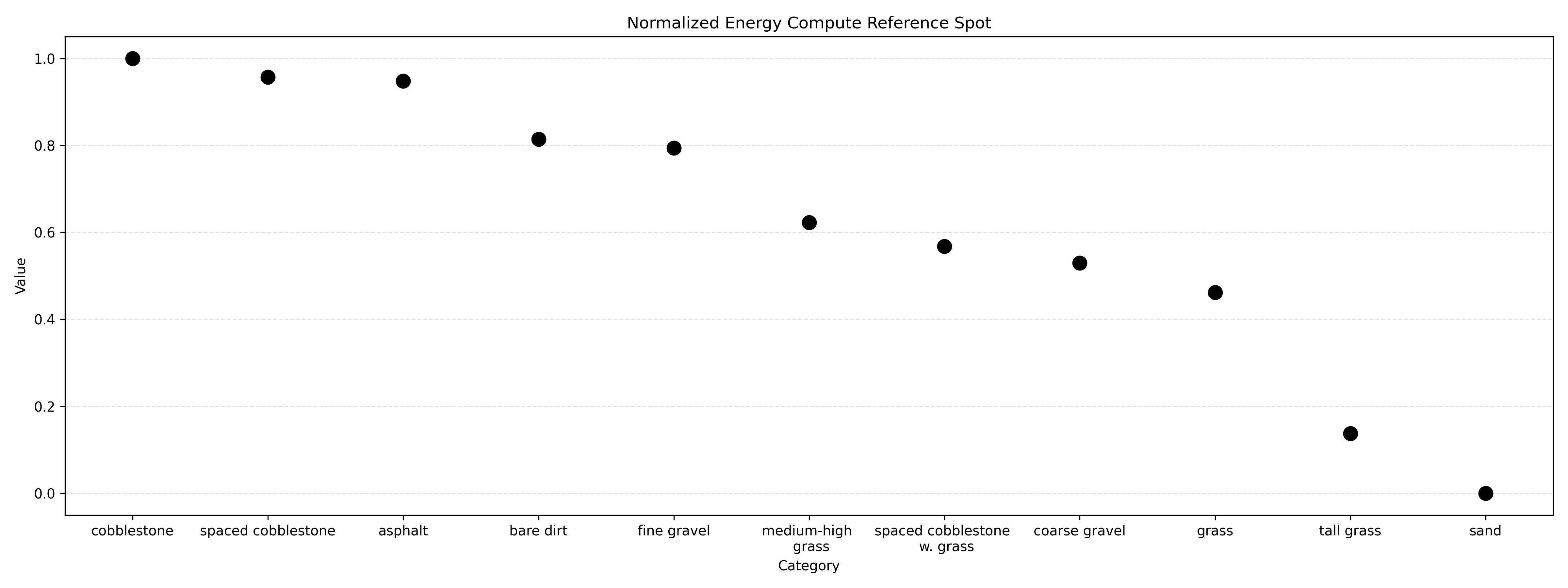} &
      \includegraphics[width=\linewidth]{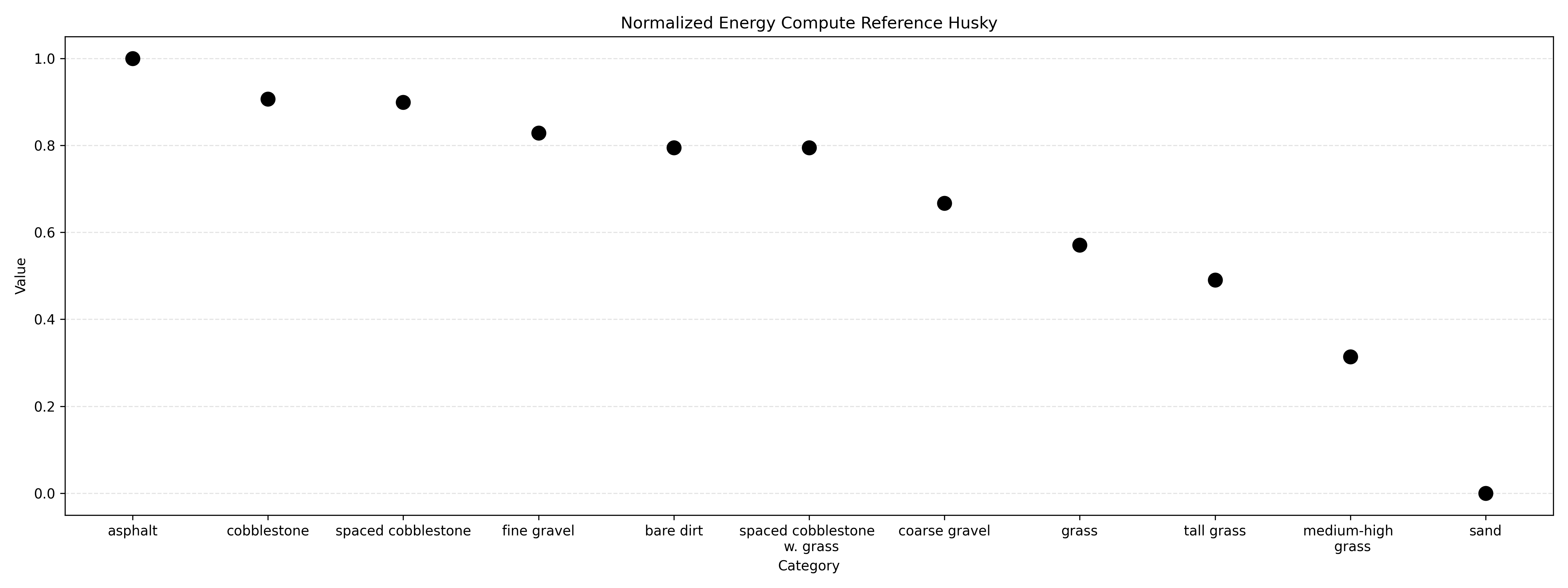} \\

      \textbf{Ours} &
      \includegraphics[width=\linewidth]{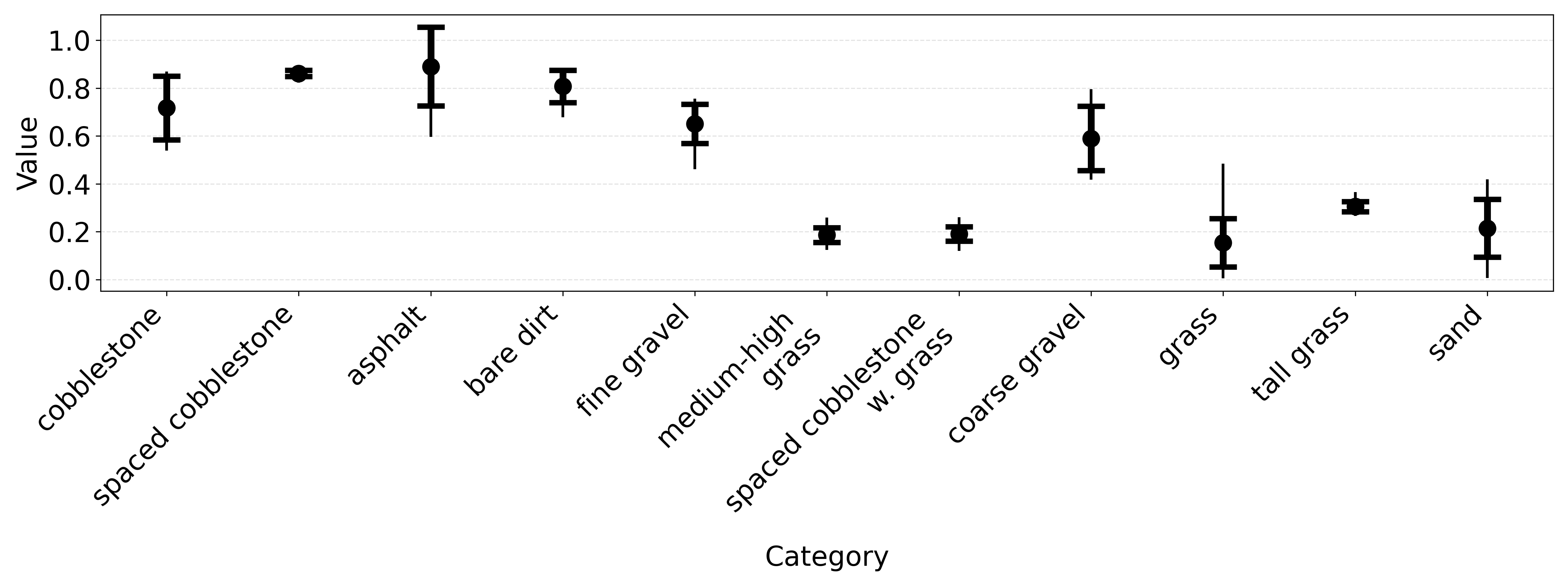} &
      \includegraphics[width=\linewidth]{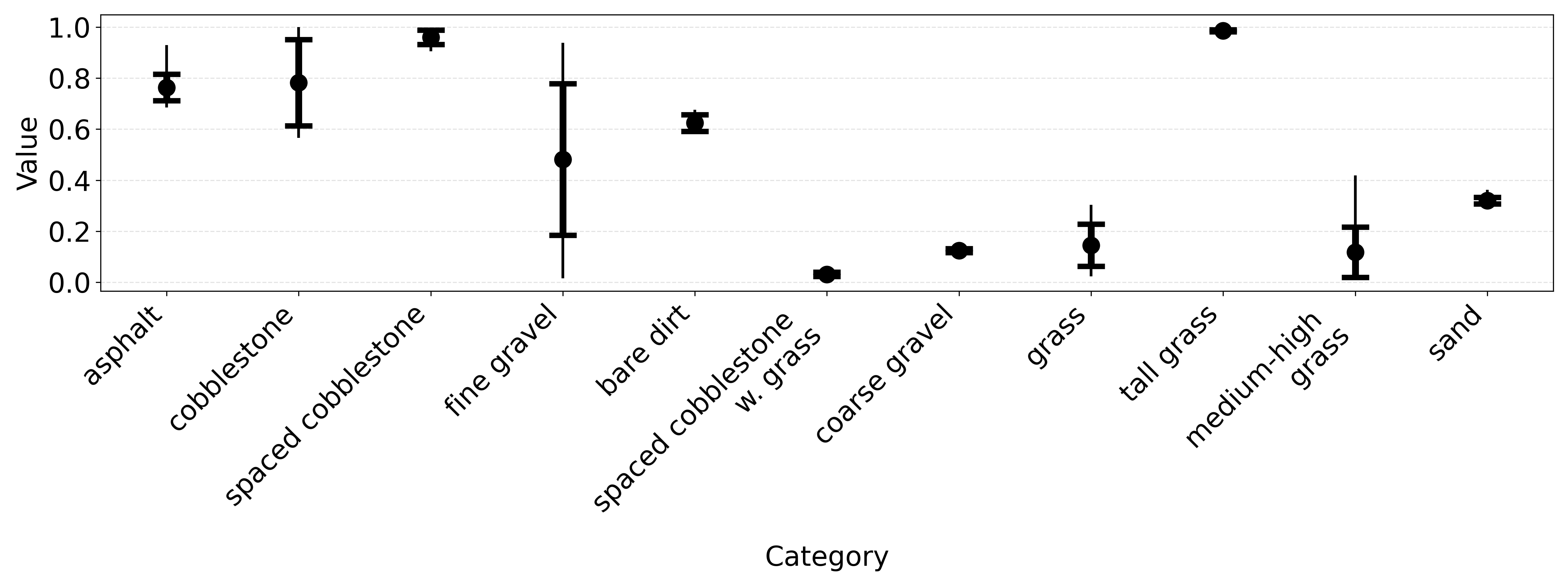} \\

      \textbf{Raw Torque} &
      \includegraphics[width=\linewidth]{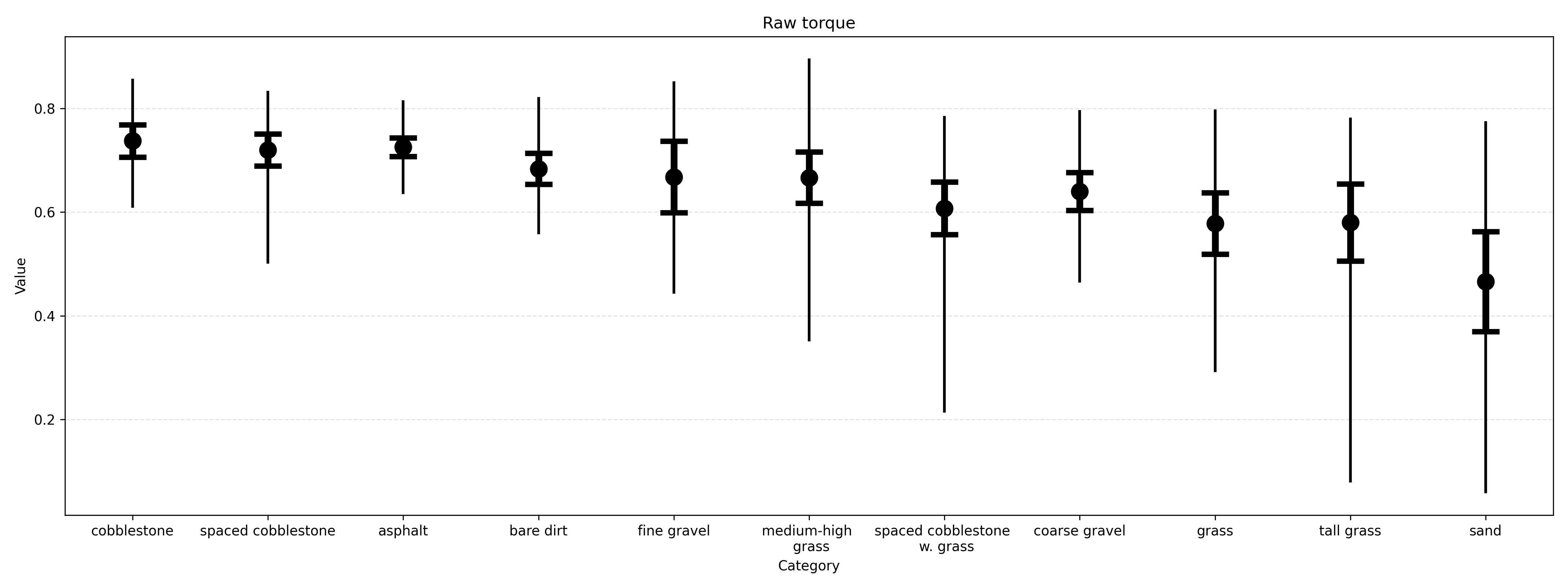} &
      \includegraphics[width=\linewidth]{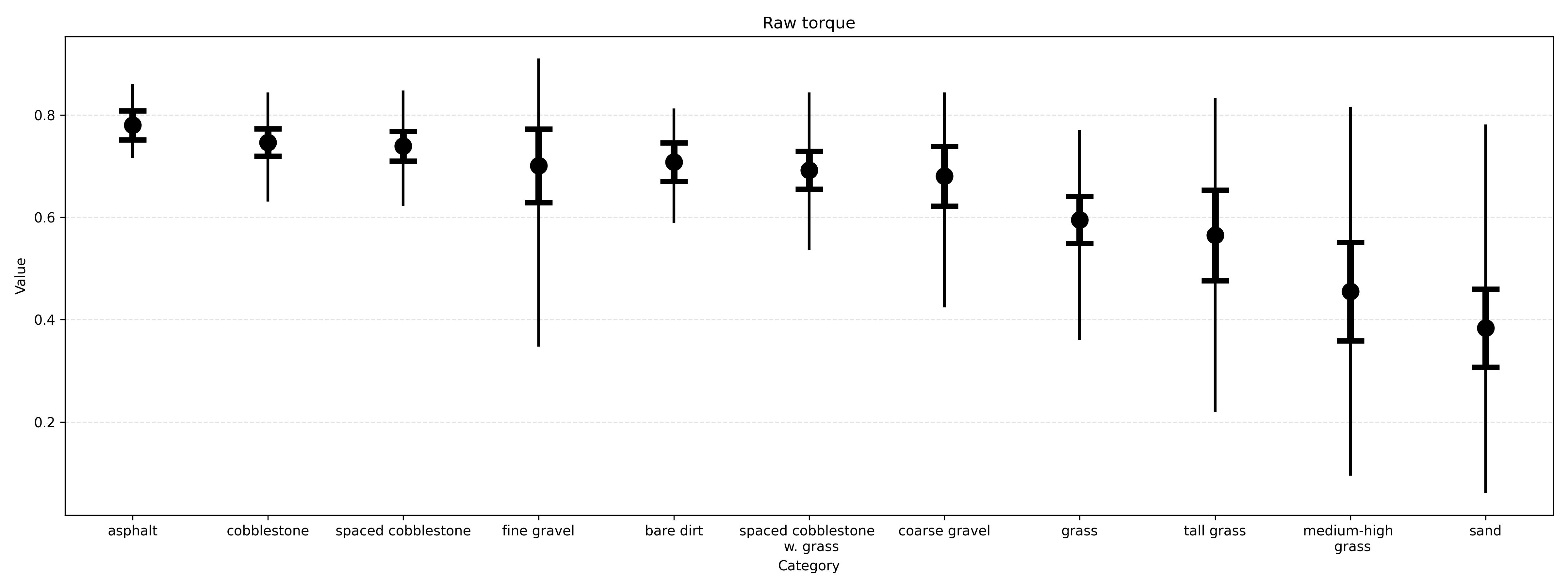} \\

      \textbf{Traction} &
      \includegraphics[width=\linewidth]{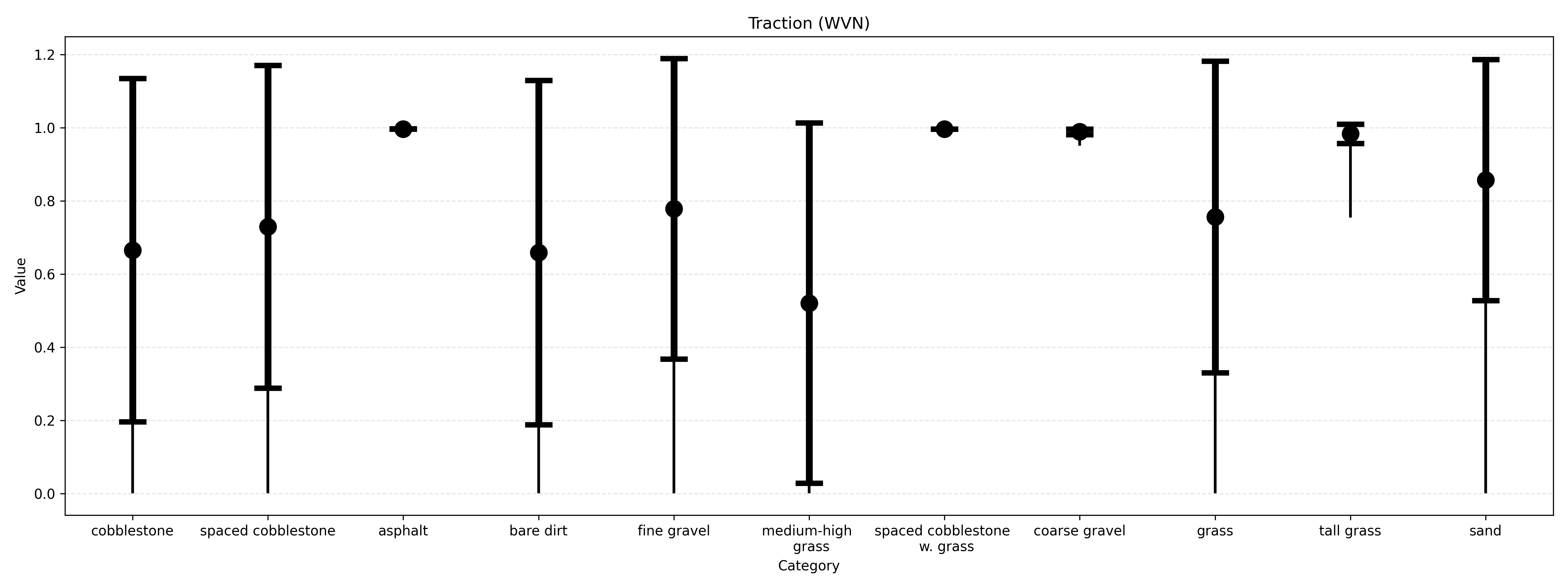} &
      \includegraphics[width=\linewidth]{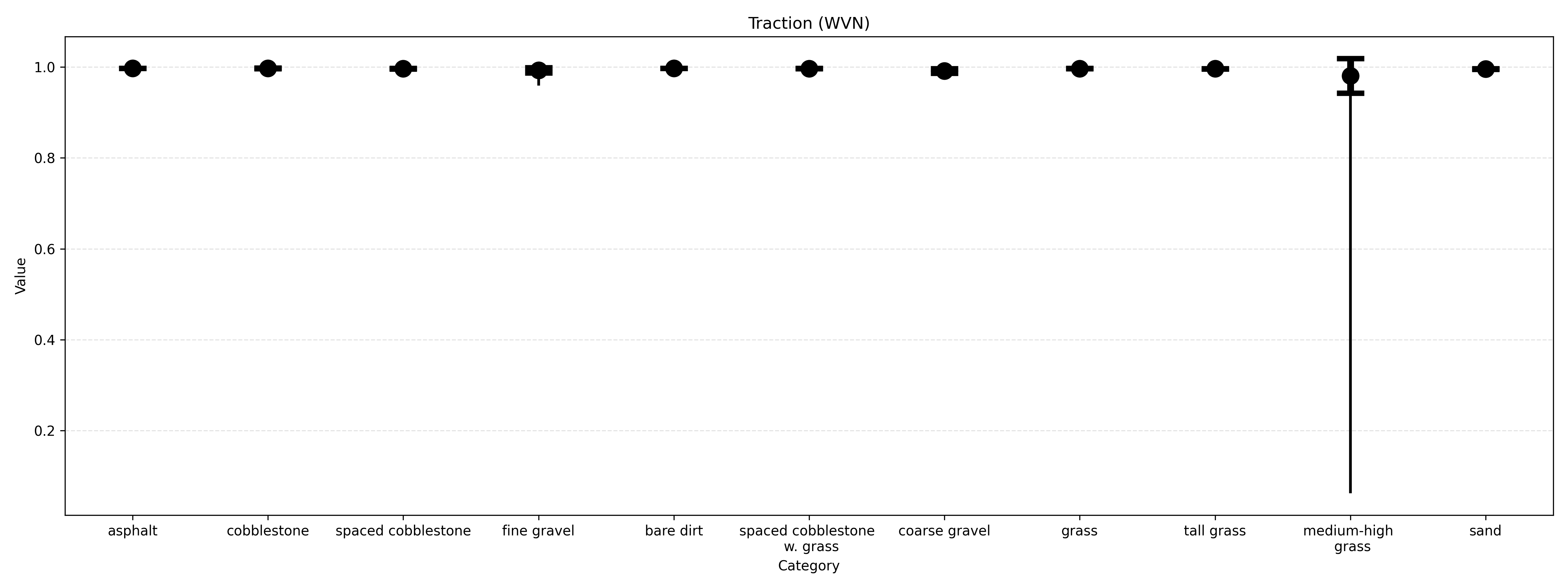} \\

      \textbf{Vibration (PSD)} &
      \includegraphics[width=\linewidth]{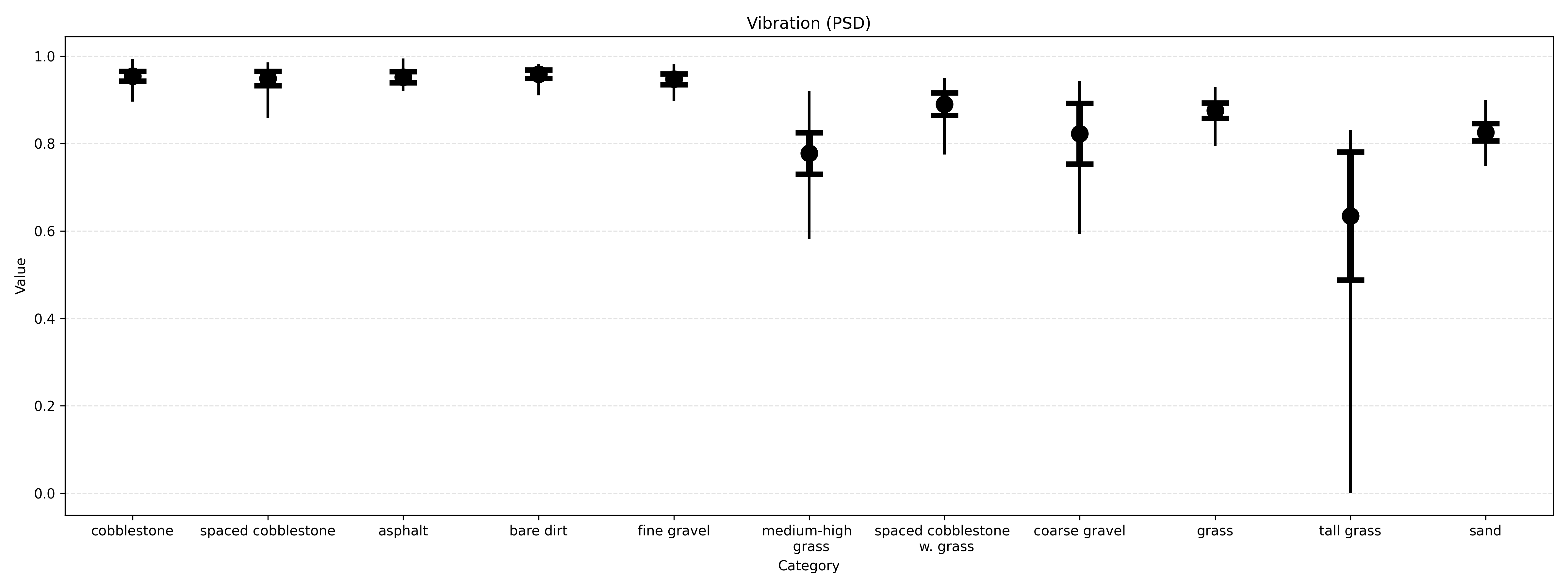} &
      \includegraphics[width=\linewidth]{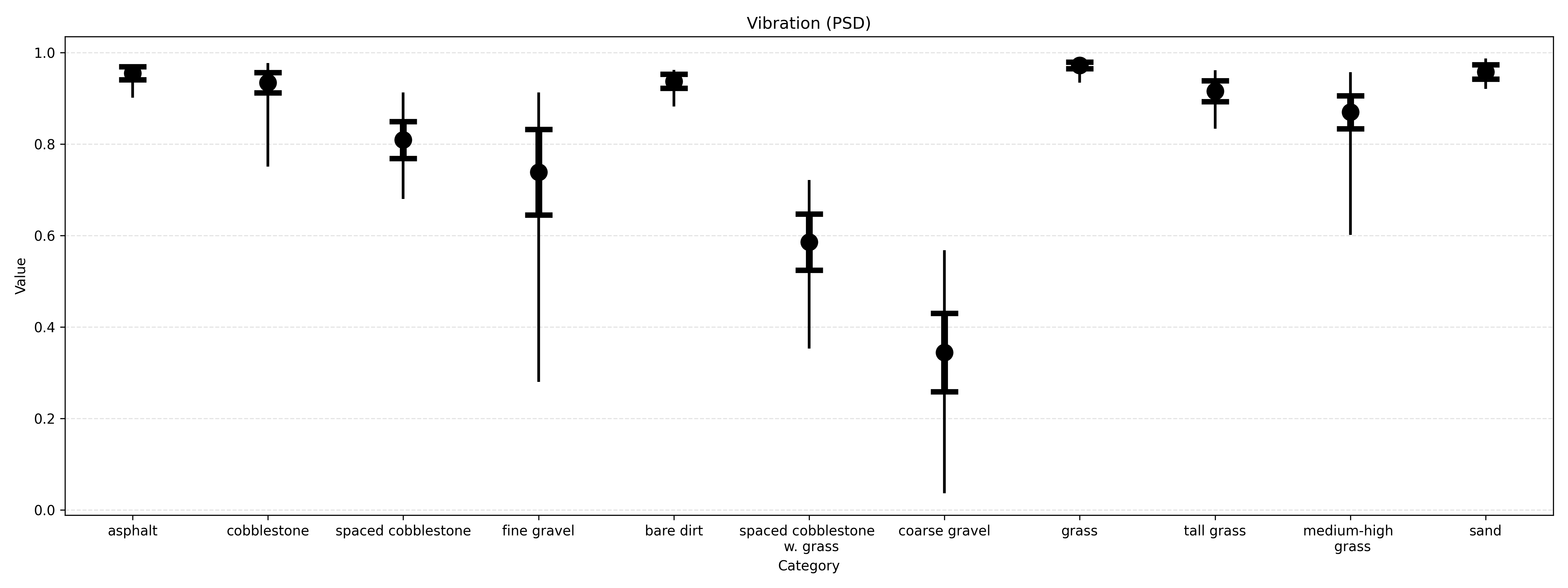} \\

    \end{tabular}
  }
  \caption{Comparison of score creation modules for Spot and Husky.}
  \label{fig:score_comp}
  \vspace{-0.1cm}
\end{figure*}

\section{Additional Quantitative Results}\label{quan1}

\net~better preserves terrain knowledge, outperforming the baselines on 7 out of 11 terrains in terms of 2D segmentation scores on Spot in \tabref{tab:iou_comparison}. We further observe that zero-shot deployment of this Spot-trained model on Husky outperforms all baselines across grass terrains.
All learning-based baselines perform poorly on bare dirt for Spot.
We attribute this to the large visual domain gap: the bare dirt in the Spot test environment, taken from a playground, differs from the pathway used in training, and appears visually similar to sand, as shown in \figref{fig:error}iii–iv, causing frequent confusion. Further, tall grass is not included in the test domain for Husky, whereas spaced cobblestone is not included in the test domains for either Husky or Spot.
Additionally, all baselines show low scores for coarse gravel on Husky, which we attribute to strong lighting conditions that hinder detection even under joint training, as illustrated in \figref{fig:error}v.
We observe the same issue for cobblestone on Husky, shown in \figref{fig:error}vi.

\begin{table*}
\centering
\caption{Comparison of 2D segmentation performance for each terrain in  mIoU scores, results are recorded in [\%]}
\begin{tabular}{p{0.5cm}p{1.5cm}|p{1cm}p{1cm}p{1cm}p{1cm}p{1cm}p{1cm}p{1cm}p{1cm}p{1cm}p{1cm}p{1cm}p{1cm}}
 \toprule
\textbf{Robot} & \textbf{Method} & asphalt & cobblestone w. grass & grass & coarse gravel & bare dirt & cobblestone & tall grass & medium-high grass & spaced cobblestone & sand & fine gravel\\
\midrule

\multirow{11}{*}{\rotatebox{90}{Spot}} & Joint & 74.41 & 52.96 & 76.62 & 16.82 & 49.45 & 25.06 & 3.59 & 62.66 & 0.88 & 11.52 & 25.83 \\
 \cmidrule{2-13}
 \cmidrule{2-13}
& LangSAM & 83.62 & 0.00 & 0.00 & 4.93 & \textbf{73.01} & 1.89 & 0.00 & 44.73 & 0.00 & 25.73 & 10.63 \\
& Image replay & 85.98 & 0.49 & 63.57 & \textbf{16.18} & 0.30 & 30.50 & \textbf{14.49} & 69.77 & 0.00 & 29.50 & \textbf{27.19} \\
& VAE-based replay\citeS{lee25cte}& 77.41 & 0.00 & 3.59 & 5.98 & 0.00 & 17.75 & 10.18 & 0.00 & 0.00 & 11.81 & 18.18 \\
& I-MOST \citeS{ma2024imost} & 70.29 & 0.00 & 0.00 & 5.99 & 0.04 & 8.01 & 0.00 & 0.00 & 0.00 & 29.35 & 6.48 \\
 \cmidrule{2-13}
& COTRATE (Ours) & \textbf{88.01} & \textbf{23.35} & \textbf{73.15} & 9.65 & 0.92 & \textbf{35.39} & 6.24 & \textbf{71.96} & \textbf{1.57} & \textbf{30.94} & 24.33 \\
& COTRATE (Ours, Zero-shot) & 55.53 & 7.89 & 58.99 & 5.97 & 0.13 & 23.36 & 5.96 & 57.90 & 0.00 & 15.84 & 1.10 \\

\midrule
\midrule 
\multirow{11}{*}{\rotatebox{90}{Husky}} & Joint & 60.47 & 0.00 & 65.59 & 0.02 & 14.88 & 11.06 & 0.00 & 43.81 & 0.00 & 3.89 & 8.66 \\
 \cmidrule{2-13}
 \cmidrule{2-13}
 \cmidrule{2-13}
& LangSAM & \textbf{81.65} & 0.00 & 0.00 & \textbf{6.85} & \textbf{69.52} & 0.64 & 0.00 & 37.16 & 0.00 & 16.39 & 8.02 \\ 
& Image replay & 55.14 & 0.00 & 54.21 & 0.00 & 0.31 & 15.45 & 0.00 & 33.29 & 0.00 & 21.64 & 8.07 \\
& VAE-based replay\citeS{lee25cte}& 61.33 & 0.00 & 20.38 & 0.00 & 0.34 & 14.88 & 0.00 & 46.26 & 0.00 & \textbf{22.62} & 6.44 \\
& I-MOST \citeS{ma2024imost} & 61.62 & 0.00 & 1.54 & 0.00 & 0.26 & 12.00 & 0.00 & 28.25 & 0.00 & 12.79 & 5.43 \\
 \cmidrule{2-13}
& COTRATE (Ours) & 72.88 & 0.00 & 44.61 & 0.00 & 0.16 & \textbf{17.49} & 0.00 & 37.13 & 0.00 & 10.33 & 8.09 \\
& COTRATE (Ours, Zero-shot) & 75.23 & 0.00 & \textbf{61.90} & 0.00 & 12.66 & 15.15 & 0.00 & \textbf{48.45} & 0.00 & 17.25 & \textbf{9.15} \\
\bottomrule 
\end{tabular}
\label{tab:iou_comparison}
\end{table*}

\begin{figure*}
\centering
\footnotesize
\setlength{\tabcolsep}{0.05cm}
{
\renewcommand{\arraystretch}{0.2}
\newcolumntype{M}[1]{>{\centering\arraybackslash}m{#1}}
\begin{tabular}{M{2.2cm}M{2.3cm}M{2.7cm}M{2.5cm}M{2.5cm}M{2.27cm}}
\includegraphics[width=\linewidth, frame]{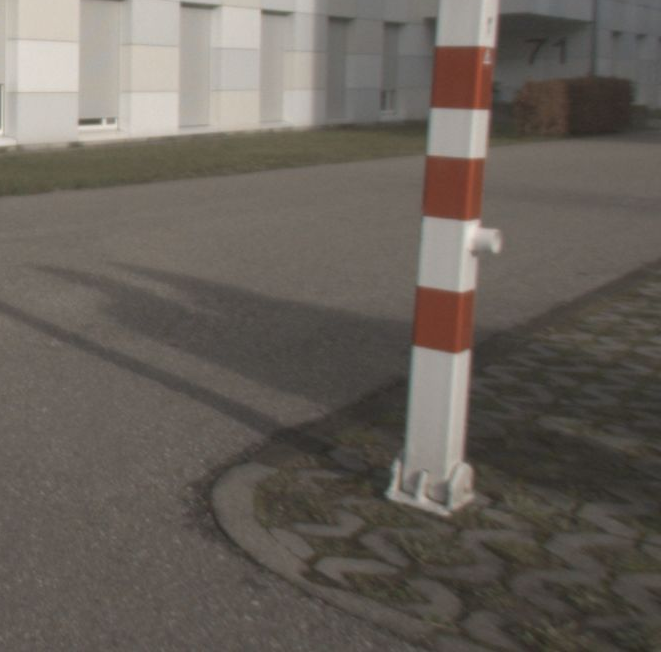} & 
\includegraphics[width=\linewidth, frame]{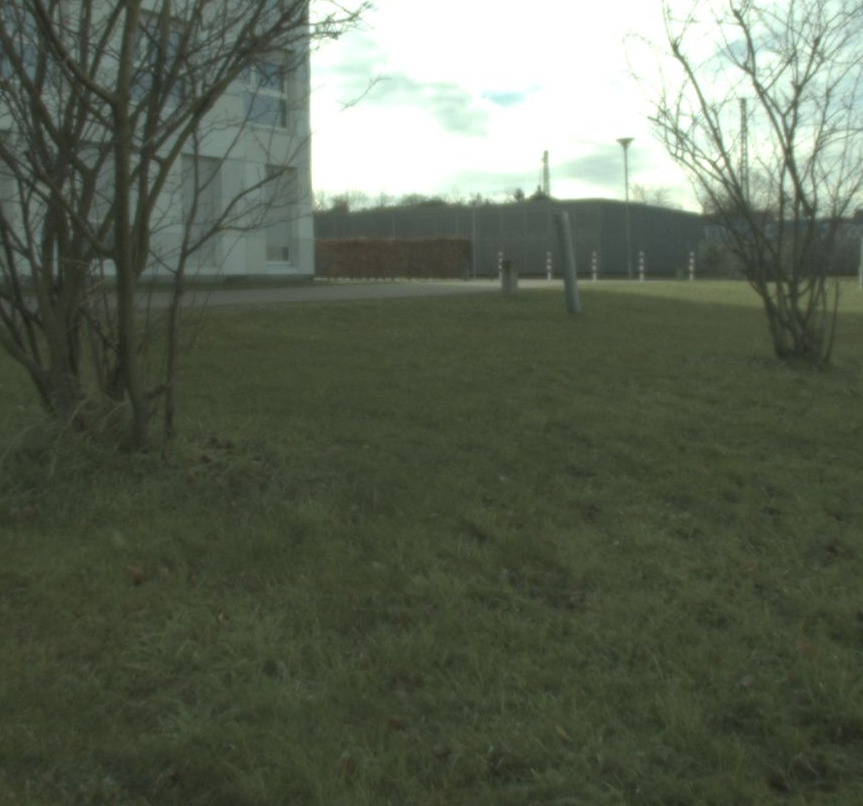} & 
\includegraphics[width=\linewidth, frame]{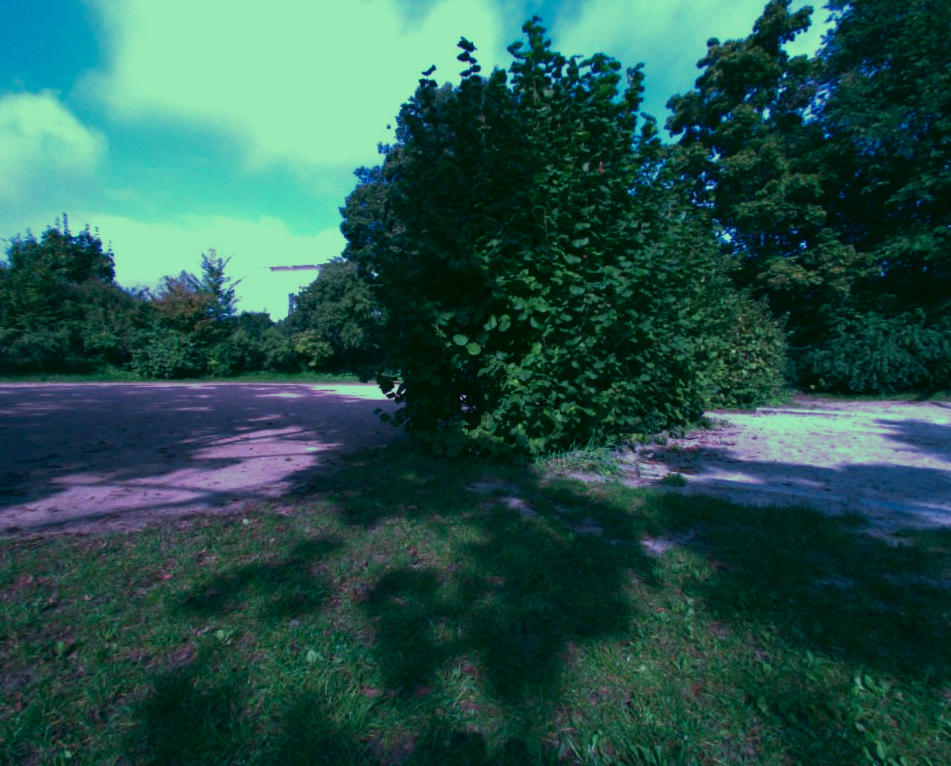} &
\includegraphics[width=\linewidth, frame]{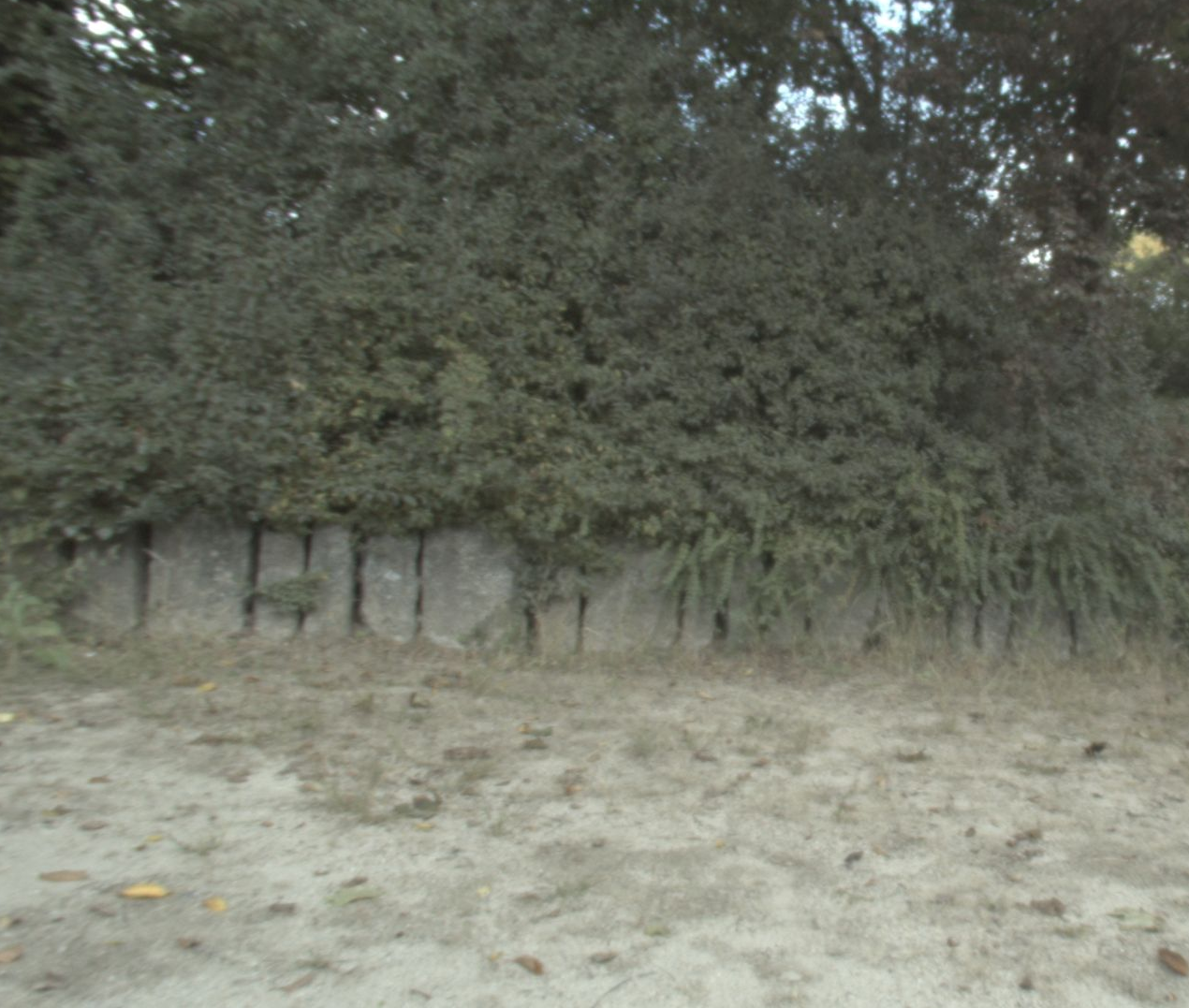} & 
\includegraphics[width=\linewidth, frame]{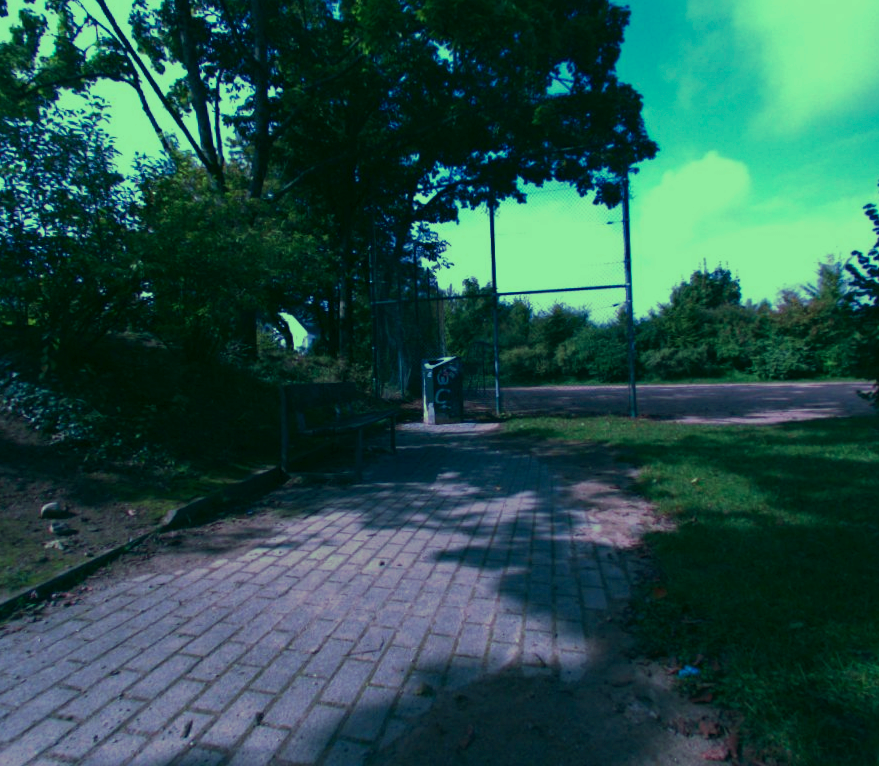} & \includegraphics[width=\linewidth, frame]{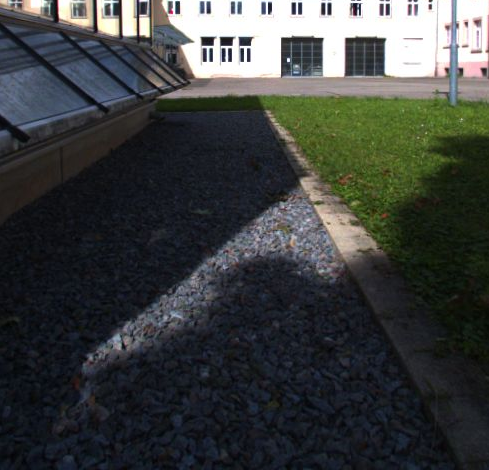} \\ 
(i) Differentiation paved-terrain LangSAM & (ii) Differentiation grass types LangSAM & (iii) Differentiation sand and bare dirt Husky & (iv) Differentiation sand and bare dirt Spot & (v) Lighting cobblestone Husky & (vi) Lighting coarse gravel Husky \\
\end{tabular}
}
\caption{Sample images of failure cases.}
\label{fig:error}
\vspace{-0.3cm}
\end{figure*}

\section{Additional Qualitative Results}\label{qual1}
In this section, we show the created paths of all baselines and our method. For Spot, the three environments E1, E2, E3, are detailed in \figref{fig:map_spot_E1}, \figref{fig:map_spot_E2}, and \figref{fig:map_spot_E3} while for Husky we show them in \figref{fig:map_husky_E1}, \figref{fig:map_husky_E2}, and \figref{fig:map_husky_E3}, respectively. We observe that \net~avoids traversing the high-cost terrains of all grass types in all selected environments, resulting in the most energy-efficient navigation paths. Additionally, we note that the zero-shot model (shown in grey) identifies effective solutions by largely following the baseline trajectories while applying small modifications that improve energy efficiency. Furthermore, we note that \net{} produces traversability maps with uniform scores across terrain segments, unlike LangSAM, which often mixes areas at the edges, as shown in \figref{fig:error}i-ii. Our model also enables more fine-grained terrain differentiation, e.g., between different types of cobblestone and asphalt.
However, we observe that our model commonly confuses bare dirt and sand for both robots in \figref{fig:map_spot_E3} and \figref{fig:map_husky_E3}. We attribute this to the large visual domain gap: the sand in the test environment resembles bare dirt, which is highly traversable, as discussed in \secref{quan1}.

\begin{figure*}[t]
  \centering
  \begin{tikzpicture}
  \node[anchor=north west,inner sep=0] (image) at (0,0){\includegraphics[trim={0 0 0 0},clip,height=4.1cm]{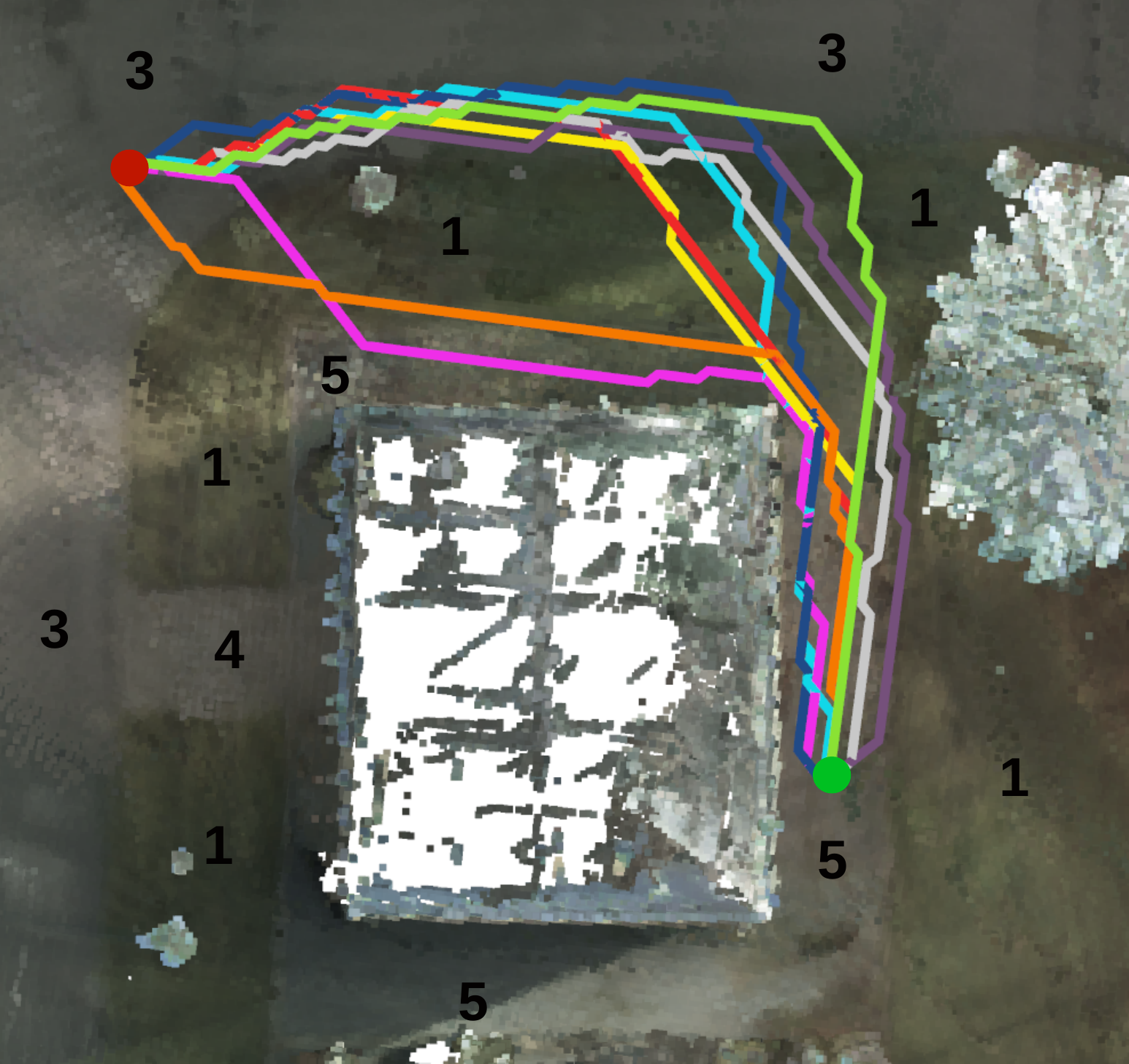}};
  \node[anchor=north west,inner sep=0, xshift=0.2em] (image1) at (image.north east){\includegraphics[trim={0 0 0 0},clip,height=4.1cm]{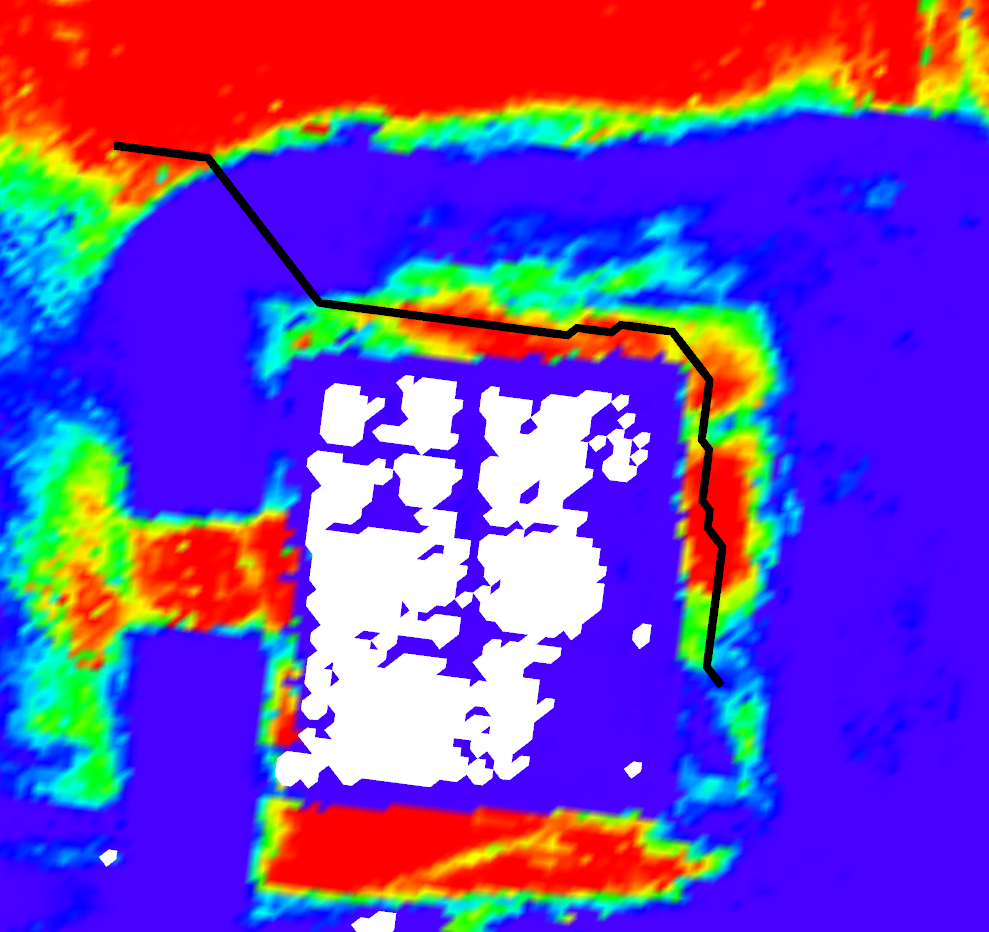}};
  \node[anchor=north west,inner sep=0, xshift=0.2em] (image2) at (image1.north east){\includegraphics[trim={0 0 0 0},clip,height=4.1cm]{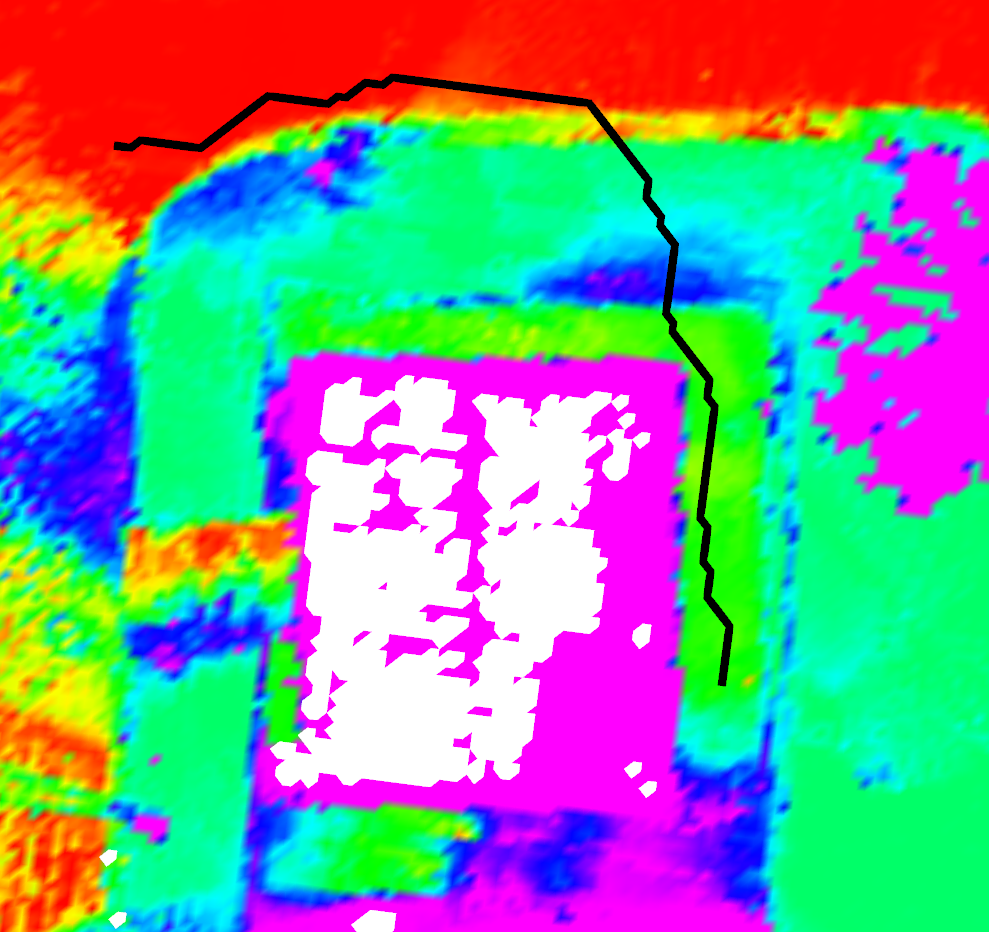}};
  \node[anchor=north west,inner sep=0, xshift=0.2em] (image3) at (image2.north east){\includegraphics[trim={0 0 0 0},clip,height=4.1cm]{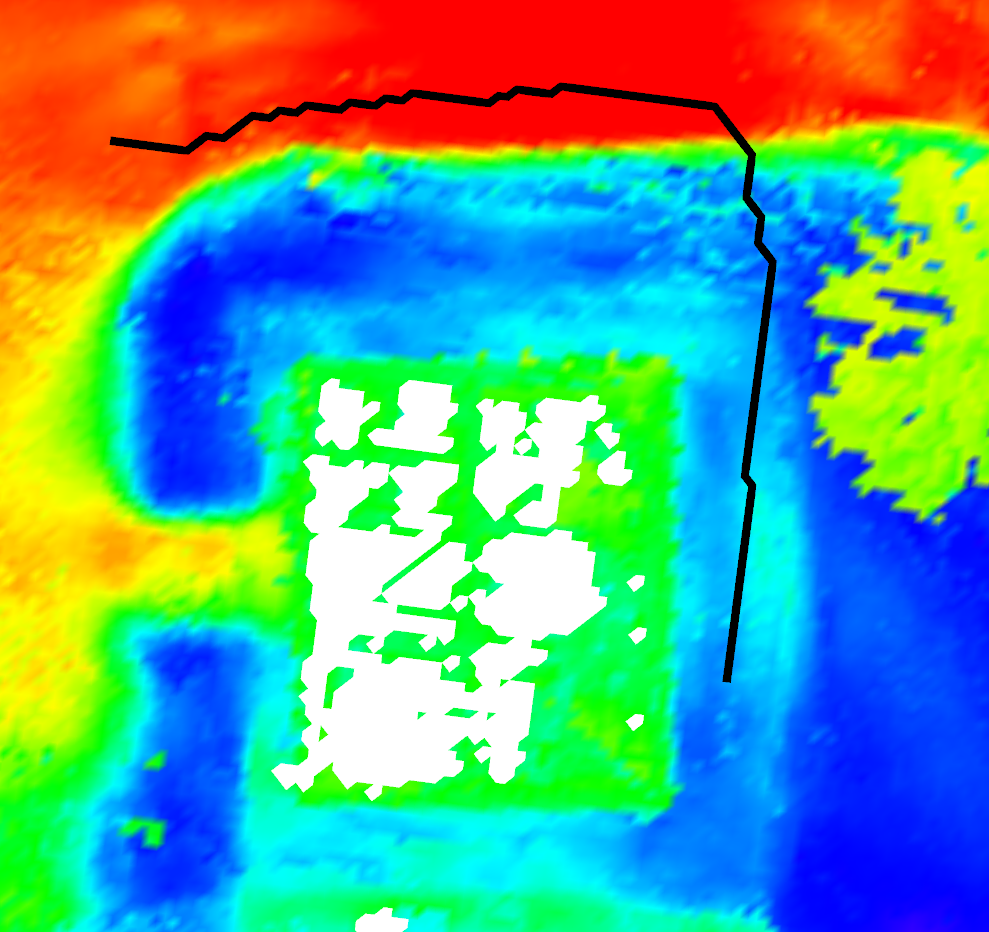}};

  \node[inner sep=0.,scale=.9, anchor=north,yshift=-0.3em, rectangle, align=left, font=\scriptsize\sffamily] (n_0) at (image.south) {RGB};
  \node[inner sep=0.,scale=.9, anchor=north,yshift=-0.3em, rectangle, align=left, font=\scriptsize\sffamily] (n_0) at (image1.south) {WVN};
  \node[inner sep=0.,scale=.9, anchor=north,yshift=-0.3em, rectangle, align=left, font=\scriptsize\sffamily] (n_0) at (image2.south) {LangSAM};
  \node[inner sep=0.,scale=.9, anchor=north,yshift=-0.3em, rectangle, align=left, font=\scriptsize\sffamily] (n_0) at (image3.south) {Ours};

  \end{tikzpicture}
  \caption{Paths E1 for Spot. Euclidean (Orange), WVN (Pink), LangSAM (Light Blue), \net~(Light Green), VAE-based replay (Yellow), I-MOST (Purple), Image-replay (Red), Joint (Dark Blue), \net--Zero-Shot (Grey). Terrains: 1-grass, 3-asphalt, 4-cobblestone, 5-coarse gravel. WVN scores high-effort gravel close to low-effort asphalt and cobblestone, resulting in a suboptimal path. LangSAM predicts higher traversability for grass compared to ours and WVN, resulting in longer traversal on grass and higher path effort. \net~minimizes traversal on grass, resulting in the lowest path effort.}
  \label{fig:map_spot_E1}
\end{figure*}

\begin{figure*}[t]
  \centering
  \begin{tikzpicture}
  \node[anchor=north west,inner sep=0] (image) at (0,0){\includegraphics[trim={0 0 0 0},clip,height=4.8cm]{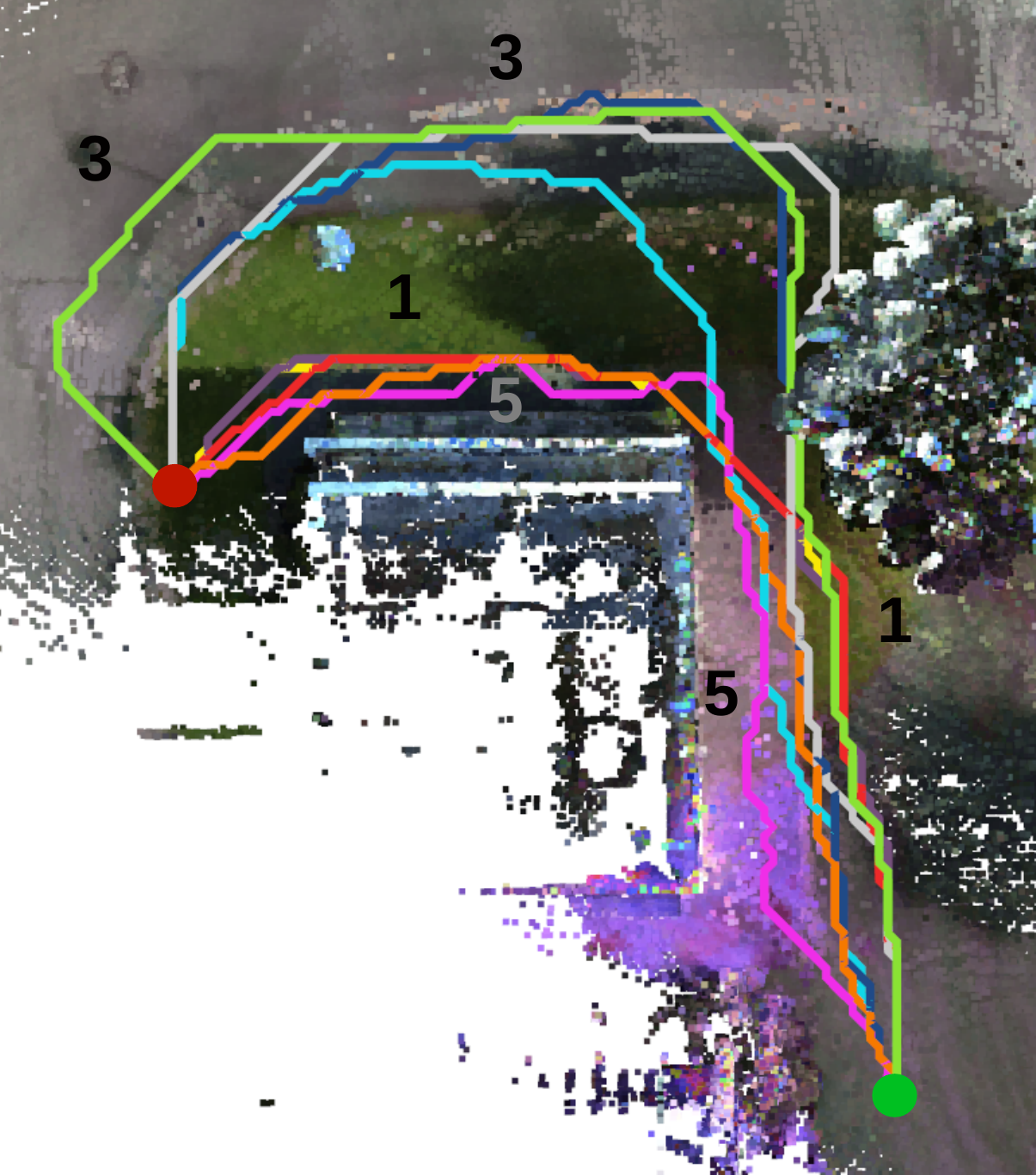}};
  \node[anchor=north west,inner sep=0, xshift=0.2em] (image1) at (image.north east){\includegraphics[trim={0 0 0 0},clip,height=4.8cm]{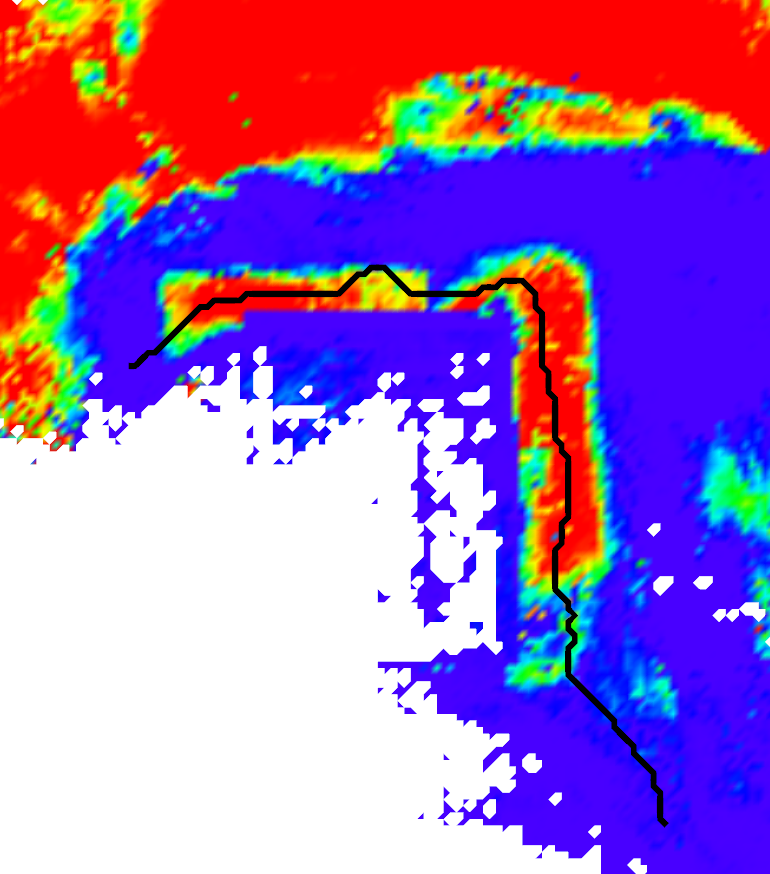}};
  \node[anchor=north west,inner sep=0, xshift=0.2em] (image2) at (image1.north east){\includegraphics[trim={0 0 0 0},clip,height=4.8cm]{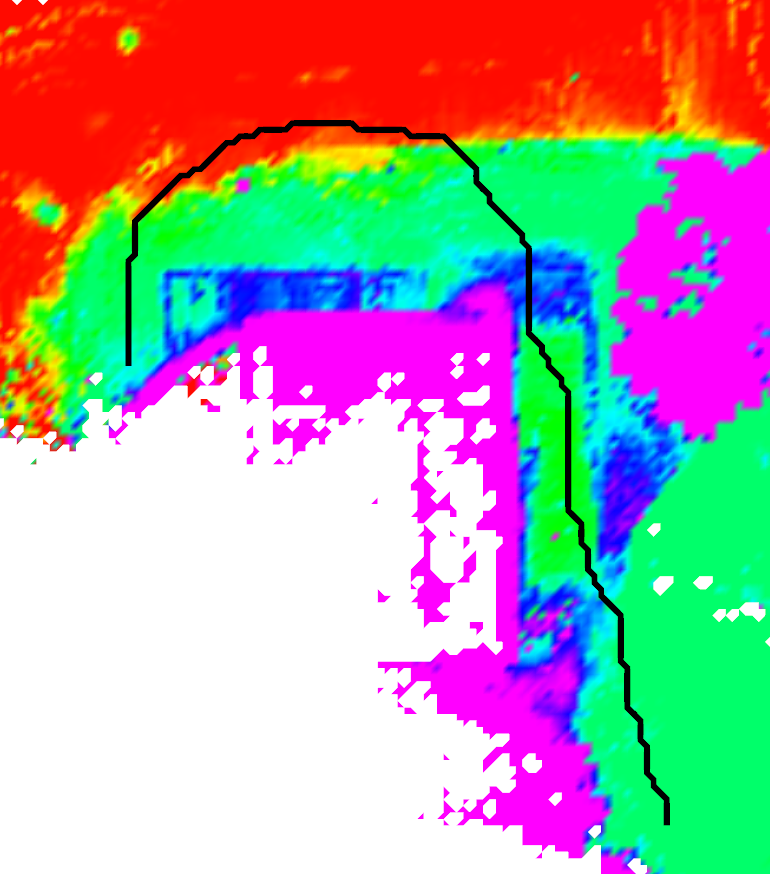}};
  \node[anchor=north west,inner sep=0, xshift=0.2em] (image3) at (image2.north east){\includegraphics[trim={0 0 0 0},clip,height=4.8cm]{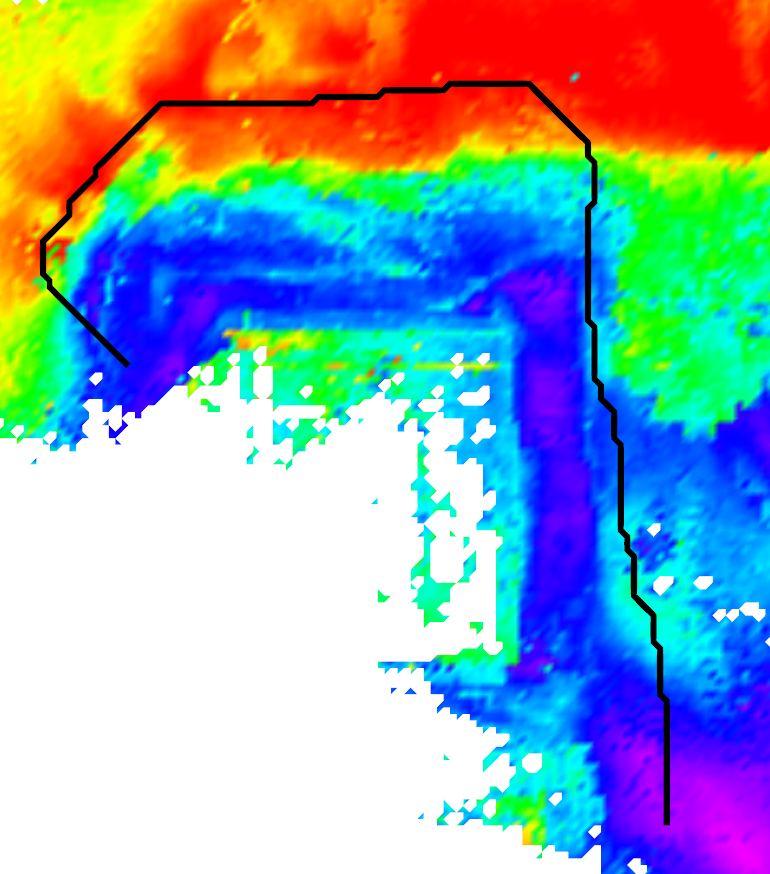}};

  \node[inner sep=0.,scale=.9, anchor=north,yshift=-0.3em, rectangle, align=left, font=\scriptsize\sffamily] (n_0) at (image.south) {RGB};
  \node[inner sep=0.,scale=.9, anchor=north,yshift=-0.3em, rectangle, align=left, font=\scriptsize\sffamily] (n_0) at (image1.south) {WVN};
  \node[inner sep=0.,scale=.9, anchor=north,yshift=-0.3em, rectangle, align=left, font=\scriptsize\sffamily] (n_0) at (image2.south) {LangSAM};
  \node[inner sep=0.,scale=.9, anchor=north,yshift=-0.3em, rectangle, align=left, font=\scriptsize\sffamily] (n_0) at (image3.south) {Ours};

  \end{tikzpicture}
  \caption{Paths E1 for Husky. Euclidean (Orange), WVN (Pink), LangSAM (Light Blue), \net~(Light Green), VAE-based replay (Yellow), I-MOST (Purple), Image-replay (Red), Joint (Dark Blue), \net--Zero-Shot (Grey). Terrains: 1-grass, 3-asphalt, 5-coarse gravel. WVN scores high-effort gravel close to low-effort asphalt and cobblestone, resulting in a suboptimal path. LangSAM predicts higher traversability for grass and gravel compared to ours, resulting in higher path effort. \net~minimizes traversal on high-effort gravel, resulting in the lowest path effort.}
  \label{fig:map_husky_E1}
\end{figure*}

\begin{figure*}[t]
  \centering
  \begin{tikzpicture}
   \node[anchor=north west,inner sep=0] (image) at (0,0){\includegraphics[trim={0 0 0 0},clip,height=9.cm]{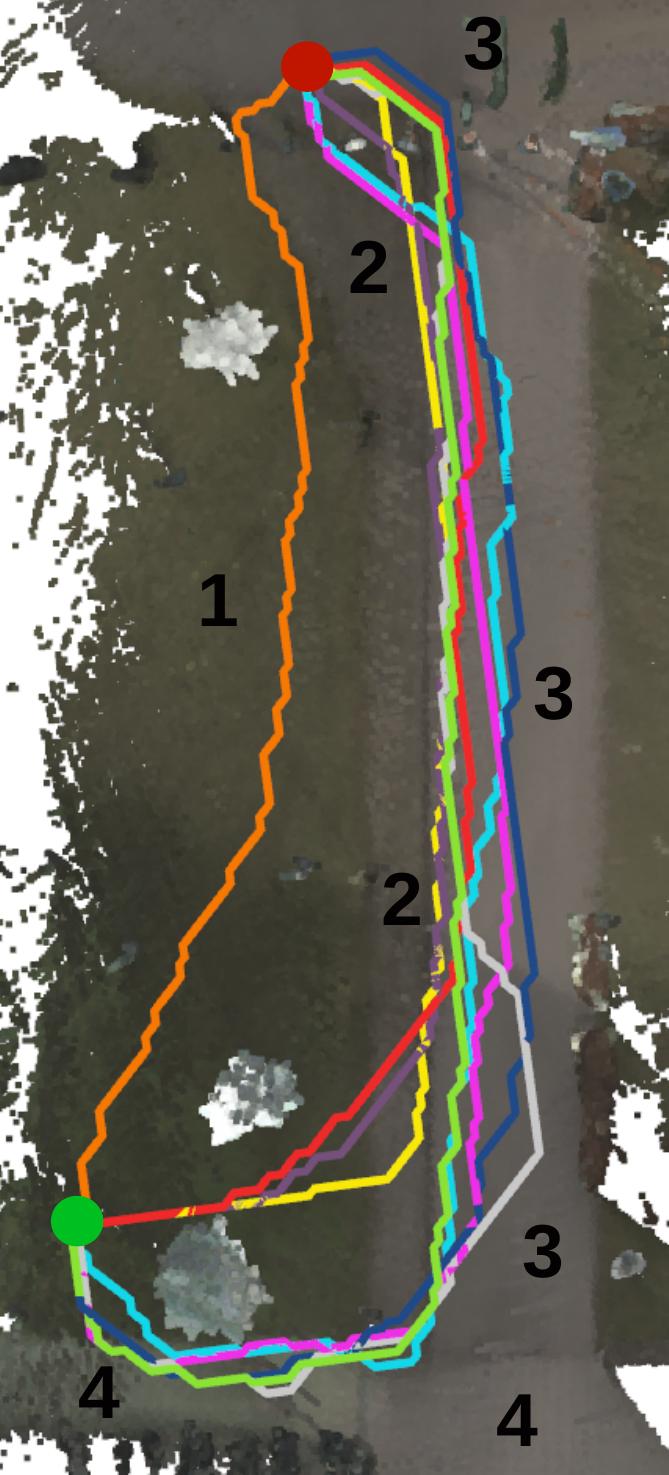}};
  \node[anchor=north west,inner sep=0, xshift=0.2em] (image1) at (image.north east){\includegraphics[trim={0 0 0 0},clip,height=9.cm]{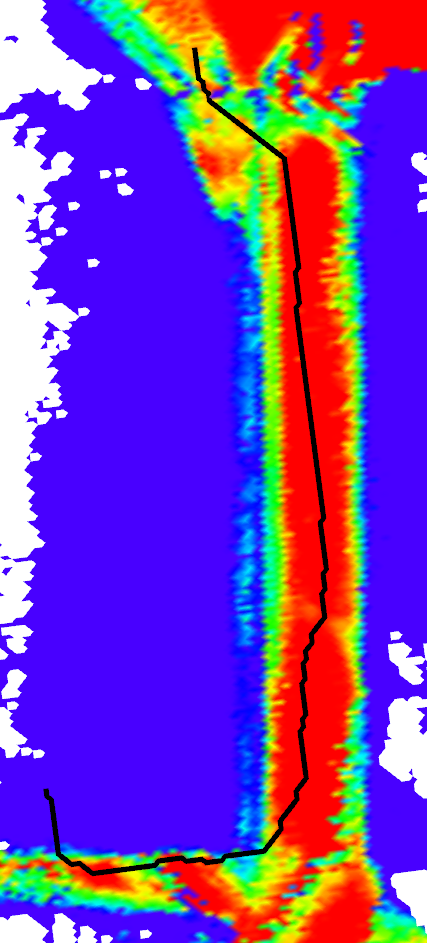}};
  \node[anchor=north west,inner sep=0, xshift=0.2em] (image2) at (image1.north east){\includegraphics[trim={0 0 0 0},clip,height=9.cm]{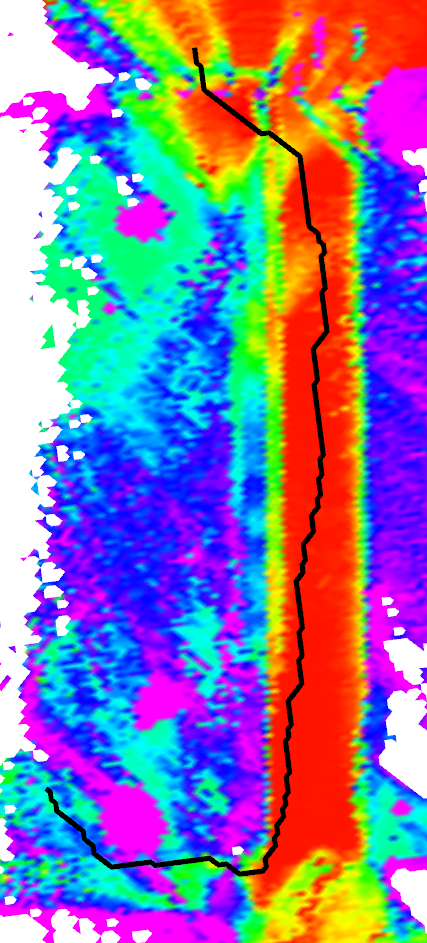}};
  \node[anchor=north west,inner sep=0, xshift=0.2em] (image3) at (image2.north east){\includegraphics[trim={0 0 0 0},clip,height=9.cm]{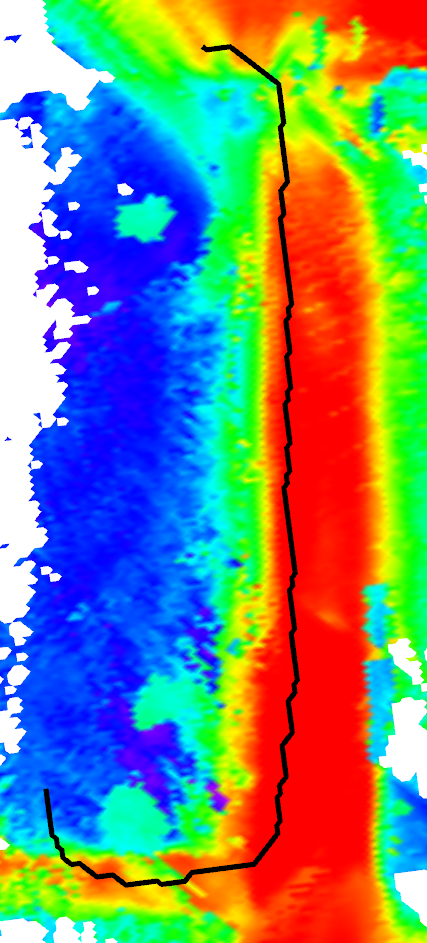}};

  \node[inner sep=0.,scale=.9, anchor=north,yshift=-0.3em, rectangle, align=left, font=\scriptsize\sffamily] (n_0) at (image.south) {RGB};
  \node[inner sep=0.,scale=.9, anchor=north,yshift=-0.3em, rectangle, align=left, font=\scriptsize\sffamily] (n_0) at (image1.south) {WVN};
  \node[inner sep=0.,scale=.9, anchor=north,yshift=-0.3em, rectangle, align=left, font=\scriptsize\sffamily] (n_0) at (image2.south) {LangSAM};
  \node[inner sep=0.,scale=.9, anchor=north,yshift=-0.3em, rectangle, align=left, font=\scriptsize\sffamily] (n_0) at (image3.south) {Ours};

  \end{tikzpicture}
  \caption{Paths E2 for Spot. Euclidean (Orange), WVN (Pink), LangSAM (Light Blue), \net~(Light Green), VAE-based replay (Yellow), I-MOST (Purple), Image-replay (Red), Joint (Dark Blue), \net--Zero-Shot (Grey). Terrains: 1-grass, 2-cobble w/ grass, 3-asphalt, 4-cobblestone. LangSAM and WVN wrongly score cobble w/ grass (2) close to asphalt near the start (red dot); LangSAM further has inconsistent grass predictions (1). \net{} consistently predicts different score ranges for grass (1), cobble w/ grass (2), and asphalt (3), resulting in the only navigation path that avoids (2) and minimizes distance on grass (1).}
  \label{fig:map_spot_E2}
\end{figure*}

\begin{figure*}[t]
  \centering
  \begin{tikzpicture}
  \node[anchor=north west,inner sep=0] (image) at (0,0){\includegraphics[trim={0 0 0 0},clip,height=4.8cm]{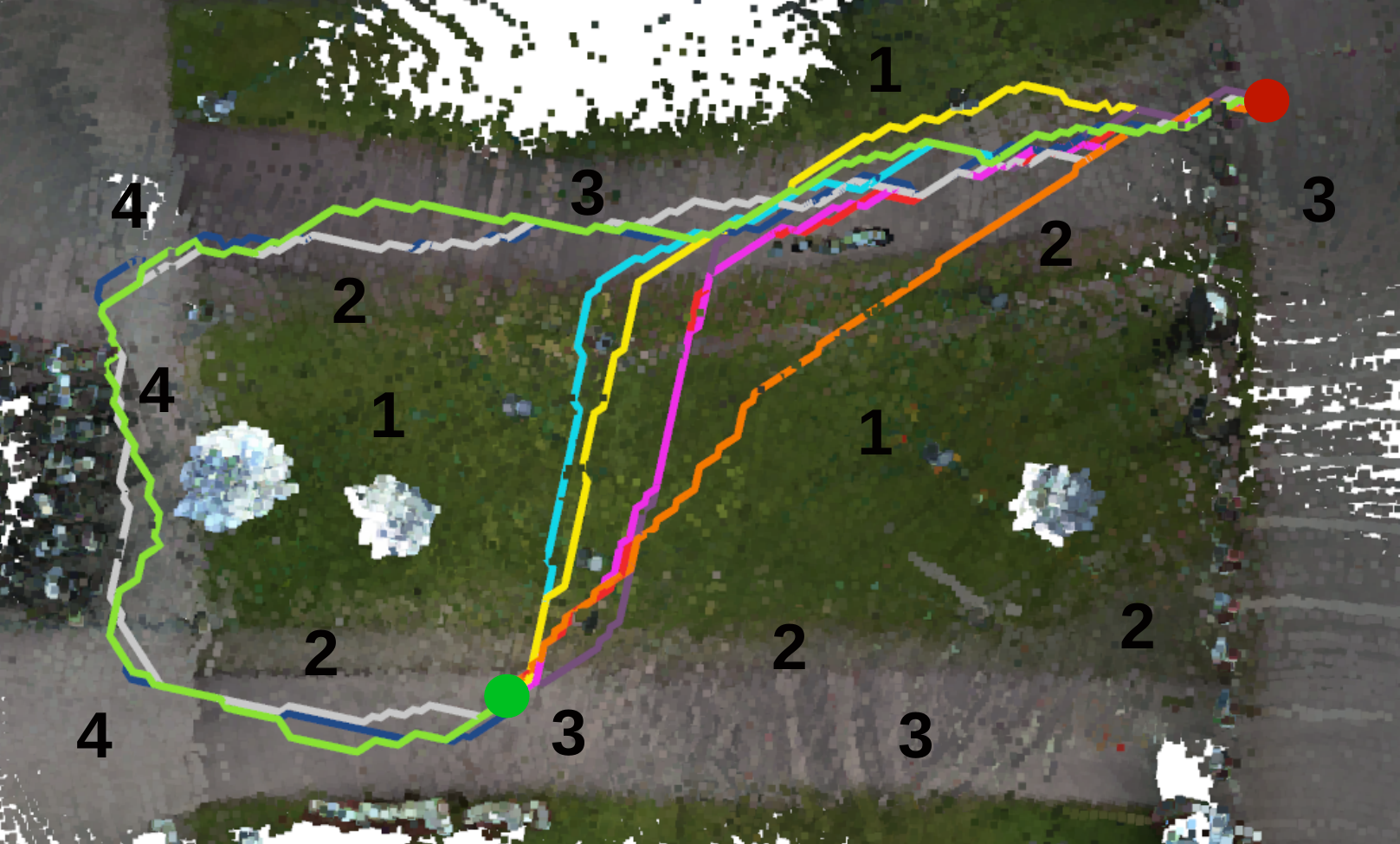}};
  \node[anchor=north west,inner sep=0, xshift=0.2em] (image1) at (image.north east){\includegraphics[trim={0 0 0 0},clip,height=4.8cm]{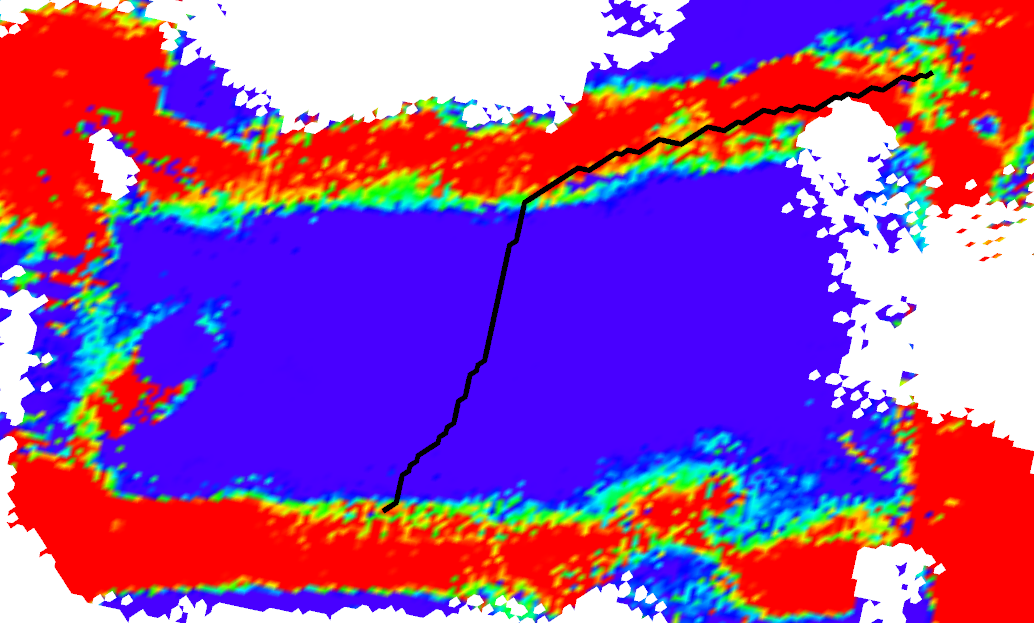}};
  \node[anchor=north west,inner sep=0, yshift=-1.2em] (image2) at (image.south west){\includegraphics[trim={0 0 0 0},clip,height=4.8cm]{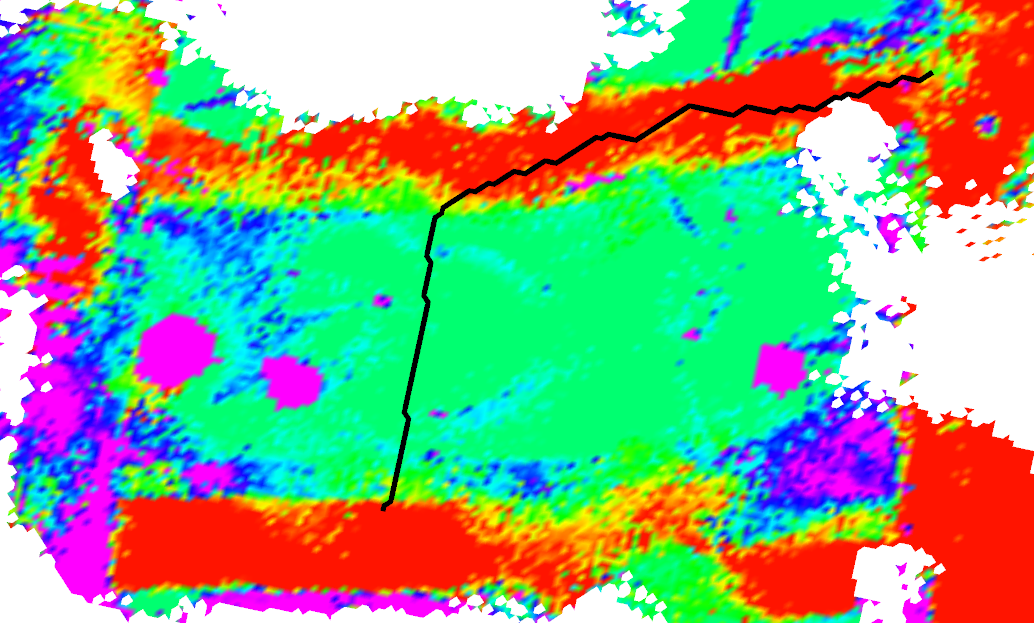}};
  \node[anchor=north west,inner sep=0, xshift=0.2em] (image3) at (image2.north east){\includegraphics[trim={0 0 0 0},clip,height=4.8cm]{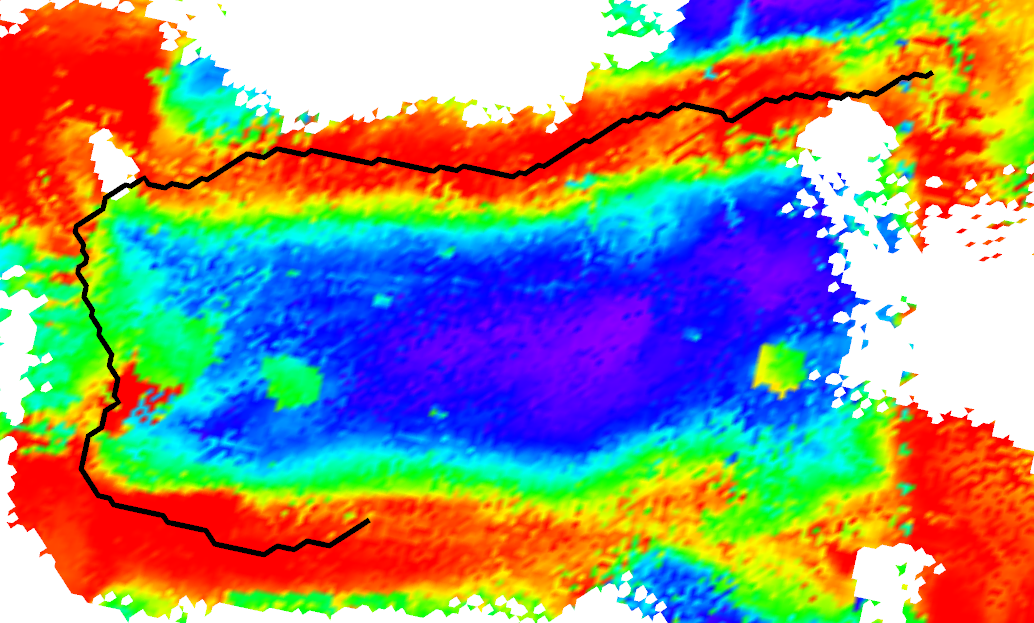}};

  \node[inner sep=0.,scale=.9, anchor=north,yshift=-0.3em, rectangle, align=left, font=\scriptsize\sffamily] (n_0) at (image.south) {RGB};
  \node[inner sep=0.,scale=.9, anchor=north,yshift=-0.3em, rectangle, align=left, font=\scriptsize\sffamily] (n_0) at (image1.south) {WVN};
  \node[inner sep=0.,scale=.9, anchor=north,yshift=-0.3em, rectangle, align=left, font=\scriptsize\sffamily] (n_0) at (image2.south) {LangSAM};
  \node[inner sep=0.,scale=.9, anchor=north,yshift=-0.3em, rectangle, align=left, font=\scriptsize\sffamily] (n_0) at (image3.south) {Ours};

  \end{tikzpicture}
  \caption{Paths E2 for Husky. Euclidean (Orange), WVN (Pink), LangSAM (Light Blue), \net~(Light Green), VAE-based replay (Yellow), I-MOST (Purple), Image-replay (Red), Joint (Dark Blue), \net--Zero-Shot (Grey). Terrains: 1-grass, 2-cobble w/ grass, 3-asphalt, 4-cobblestone. \net~consistently seperats the terrains (1)-(4). WVN and LangSAM wrongly predict low traversability for parts of the cobblestone (4), resulting in paths crossing the high-effort grass. Although taking a longer detour, \net{} produces the only paths consistently avoiding grass (1) and staying on the most efficient asphalt (3) and cobblestone (4), for both default setting and zero-shot transfer.}
  \label{fig:map_husky_E2}
\end{figure*}

\begin{figure*}[t]
  \centering
  \begin{tikzpicture}
  \node[anchor=north west,inner sep=0] (image) at (0,0){\includegraphics[trim={0 0 0 0},clip,height=4.8cm]{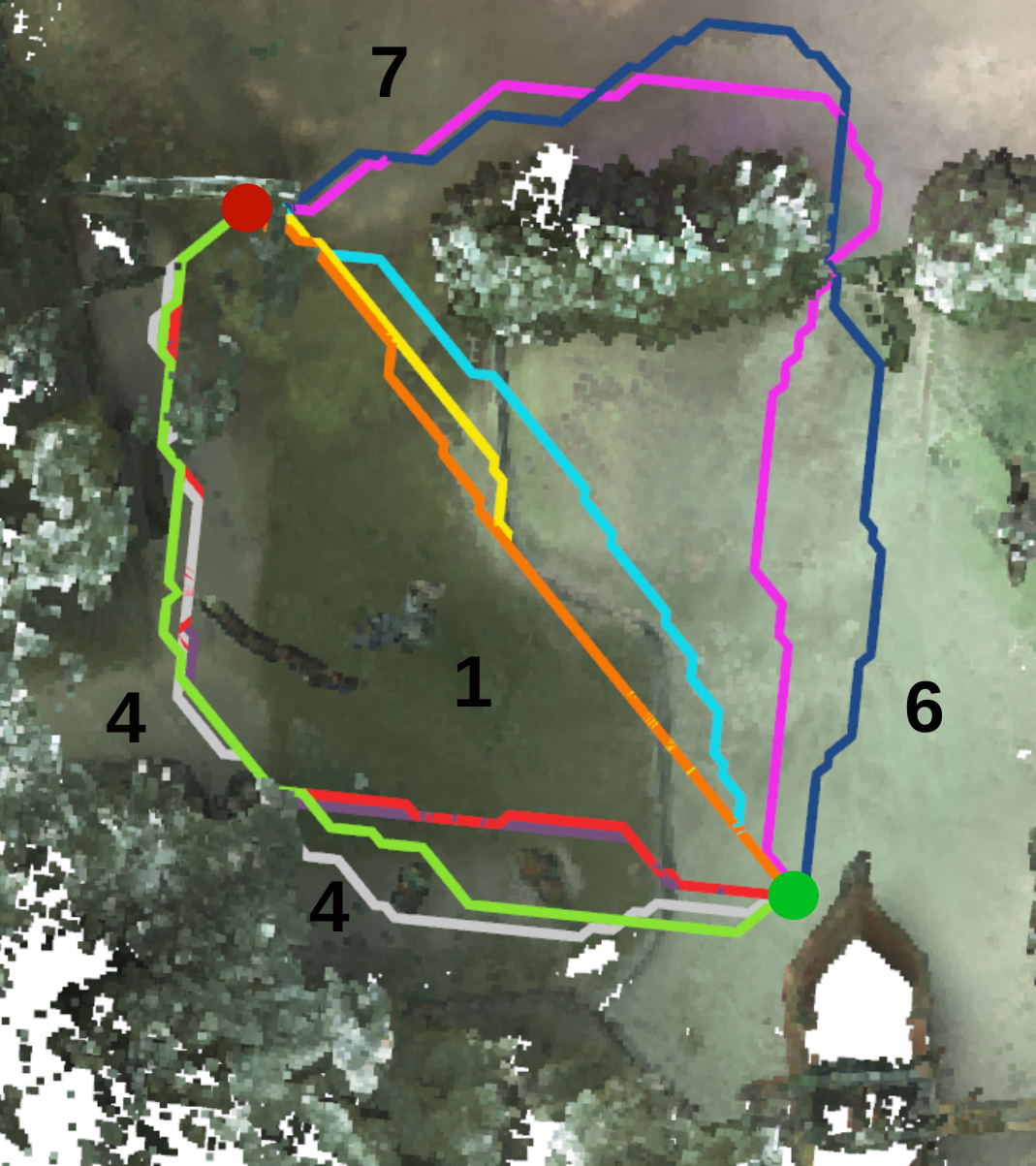}};
  \node[anchor=north west,inner sep=0, xshift=0.2em] (image1) at (image.north east){\includegraphics[trim={0 0 0 0},clip,height=4.8cm]{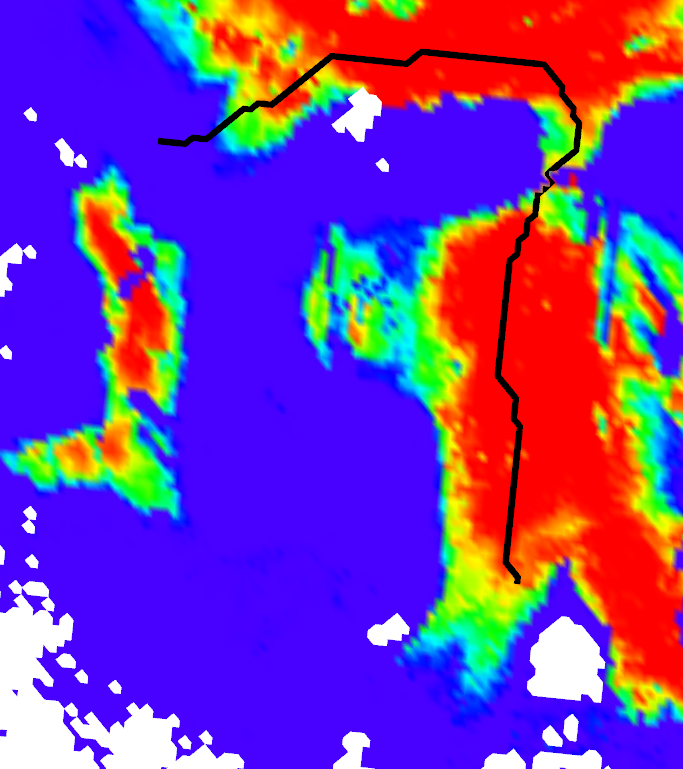}};
  \node[anchor=north west,inner sep=0, yshift=0.2em] (image2) at (image1.north east){\includegraphics[trim={0 0 0 0},clip,height=4.8cm]{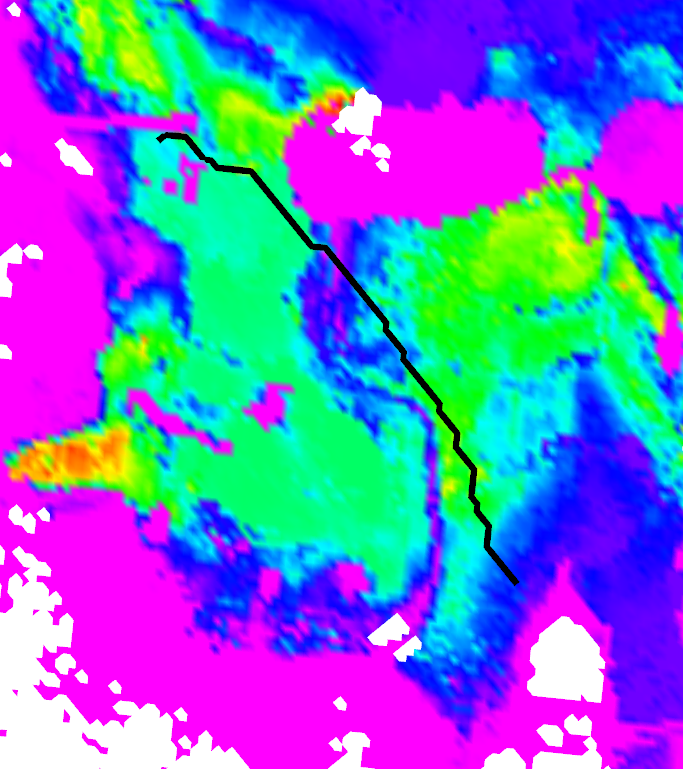}};
  \node[anchor=north west,inner sep=0, xshift=0.2em] (image3) at (image2.north east){\includegraphics[trim={0 0 0 0},clip,height=4.8cm]{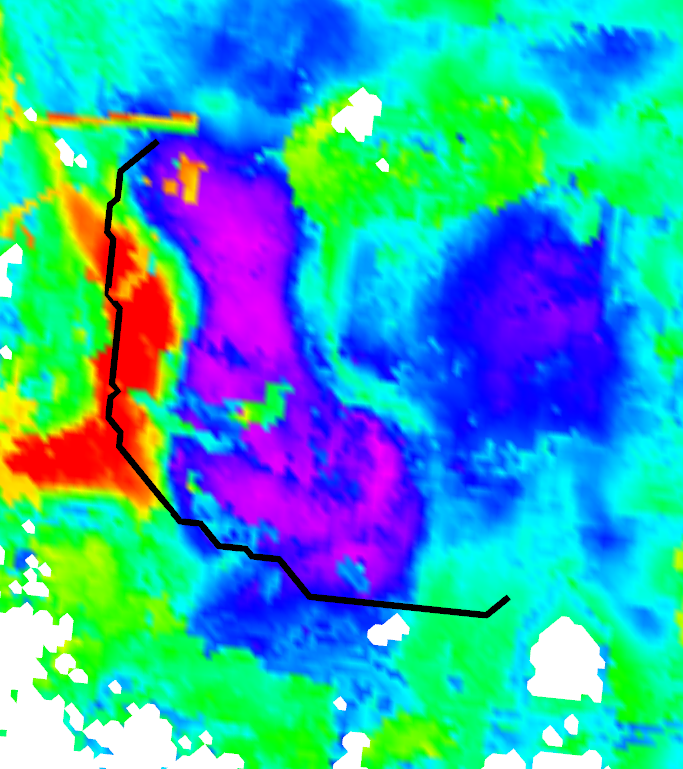}};

  \node[inner sep=0.,scale=.9, anchor=north,yshift=-0.3em, rectangle, align=left, font=\scriptsize\sffamily] (n_0) at (image.south) {RGB};
  \node[inner sep=0.,scale=.9, anchor=north,yshift=-0.3em, rectangle, align=left, font=\scriptsize\sffamily] (n_0) at (image1.south) {WVN};
  \node[inner sep=0.,scale=.9, anchor=north,yshift=-0.3em, rectangle, align=left, font=\scriptsize\sffamily] (n_0) at (image2.south) {LangSAM};
  \node[inner sep=0.,scale=.9, anchor=north,yshift=-0.3em, rectangle, align=left, font=\scriptsize\sffamily] (n_0) at (image3.south) {Ours};

  \end{tikzpicture}
  \caption{Paths E3 for Spot. Euclidean (Orange), WVN (Pink), LangSAM (Light Blue), \net~(Light Green), VAE-based replay (Yellow), I-MOST (Purple), Image-replay (Red), Joint (Dark Blue), \net--Zero-Shot (Grey). Terrains: 1-grass, 4-cobblestone, 6-sand, 7-bare dirt. WVN wrongly scores higher effort sand (6) close to cobblestone (4) and dirt (7), resulting in a high-effort path. LangSAM cannot distinguish the low-effort cobblestone (4) well. \net~superiorly distinguishes cobblestone (4), correctly avoids grass (1), and minimizes distance traveled on sand (6) in comparison to the baselines. Joint training prioritizes traversal on well-traversable bare dirt, sacrificing performance on sand.}
  \label{fig:map_spot_E3}
\end{figure*}

\begin{figure*}[t]
  \centering
  \begin{tikzpicture}
  \node[anchor=north west,inner sep=0] (image) at (0,0){\includegraphics[trim={0 0 0 0},clip,height=4.3cm]{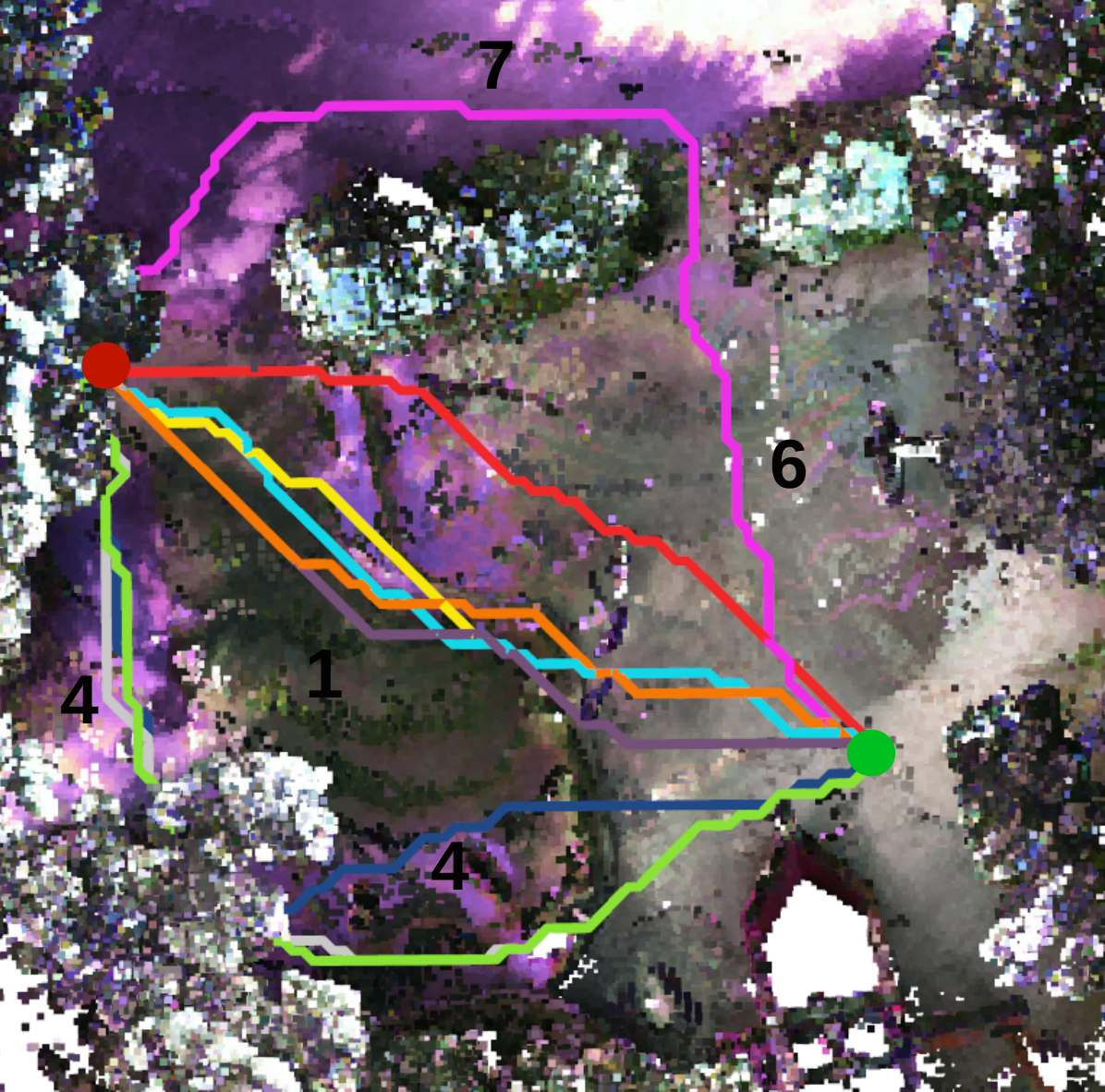}};
  \node[anchor=north west,inner sep=0, xshift=0.2em] (image1) at (image.north east){\includegraphics[trim={0 0 0 0},clip,height=4.3cm]{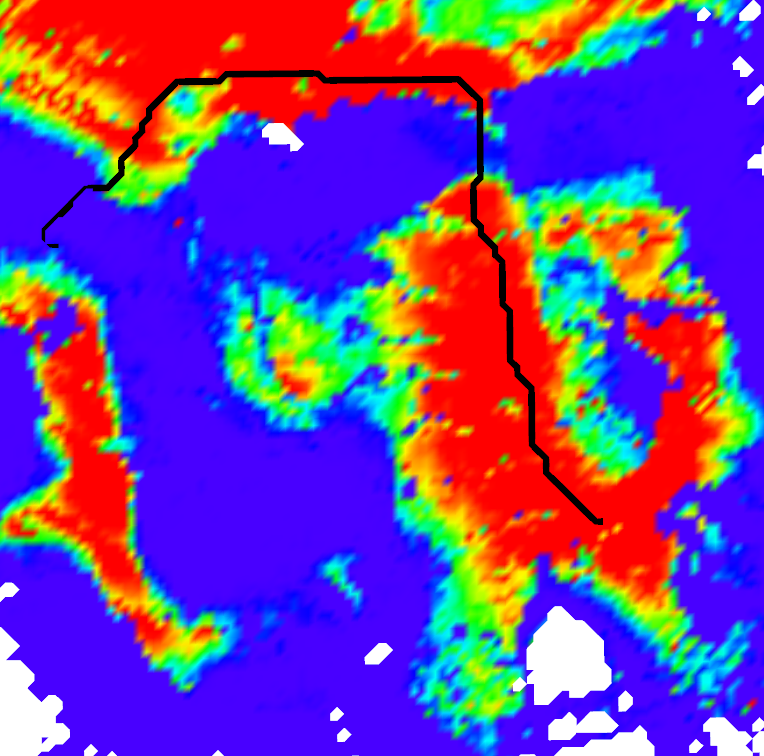}};
  \node[anchor=north west,inner sep=0, xshift=0.2em] (image2) at (image1.north east){\includegraphics[trim={0 0 0 0},clip,height=4.3cm]{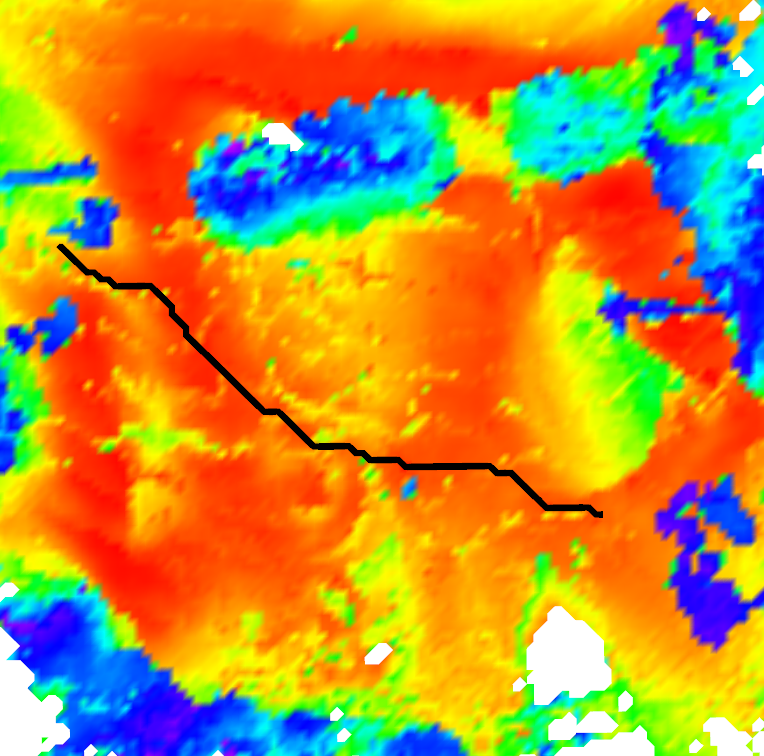}};
  \node[anchor=north west,inner sep=0, xshift=0.2em] (image3) at (image2.north east){\includegraphics[trim={0 0 0 0},clip,height=4.3cm]{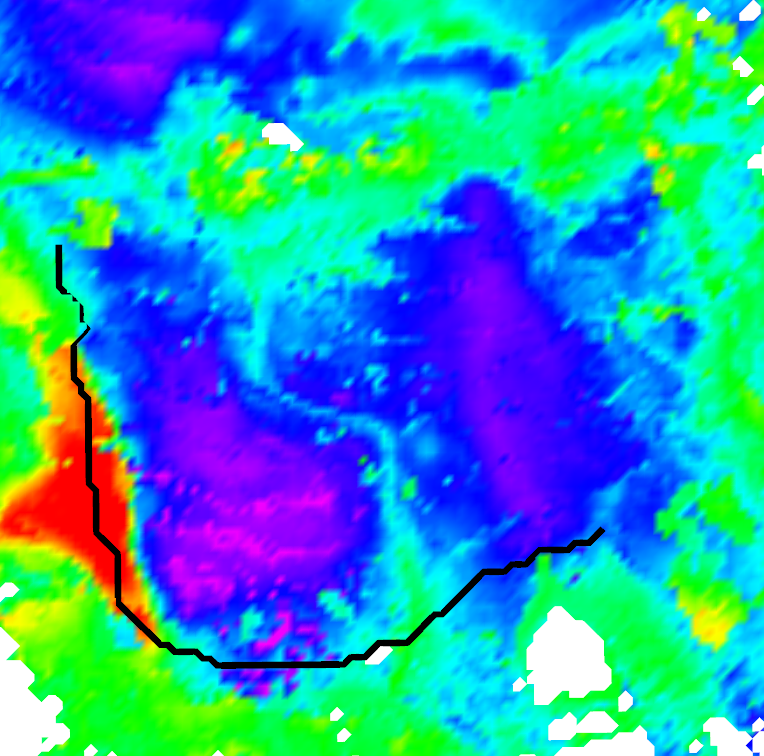}};

  \node[inner sep=0.,scale=.9, anchor=north,yshift=-0.3em, rectangle, align=left, font=\scriptsize\sffamily] (n_0) at (image.south) {RGB};
  \node[inner sep=0.,scale=.9, anchor=north,yshift=-0.3em, rectangle, align=left, font=\scriptsize\sffamily] (n_0) at (image1.south) {WVN};
  \node[inner sep=0.,scale=.9, anchor=north,yshift=-0.3em, rectangle, align=left, font=\scriptsize\sffamily] (n_0) at (image2.south) {LangSAM};
  \node[inner sep=0.,scale=.9, anchor=north,yshift=-0.3em, rectangle, align=left, font=\scriptsize\sffamily] (n_0) at (image3.south) {Ours};

  \end{tikzpicture}
  \caption{Paths E3 for Husky. Euclidean (Orange), WVN (Pink), LangSAM (Light Blue), \net~(Light Green), VAE-based replay (Yellow), I-MOST (Purple), Image-replay (Red), Joint (Dark Blue), \net--Zero-Shot (Grey). Terrains: 1-grass, 4-cobblestone, 6-sand, 7-bare dirt. WVN wrongly scores higher effort sand (6) close to cobblestone (4) and dirt (7), resulting in a high-effort path. LangSAM doesn't achieve terrain separation in this challenging environment, \net~correctly predicts the range for cobblestone (4), enabling it to produce the most efficient navigation path.}
  \label{fig:map_husky_E3}
\end{figure*}

\bibliographystyleS{IEEEtran}
\bibliographyS{references}

    \newpage
\fi

\newpage
{\footnotesize
\bibliographystyleS{IEEEtran}
\bibliographyS{references}
}

\end{document}